\documentclass{article}

 \usepackage{amsmath}
 \usepackage{amsthm}
 \usepackage{amssymb}
\usepackage[eandd, final]{neurips_2026}

\usepackage[utf8]{inputenc} 
\usepackage[T1]{fontenc}    
\usepackage{hyperref} 
\usepackage{url}      
\usepackage{booktabs} 
\usepackage{amsfonts} 
\usepackage{nicefrac} 
\usepackage{microtype}      
\usepackage{xcolor}   
\usepackage{graphicx}
\usepackage{multirow}
\usepackage{array}
\usepackage{graphicx}
\usepackage{overpic} 
\usepackage{tikz}
\usepackage{subcaption}
\usepackage{wrapfig}
\usepackage{enumitem}
\usepackage{adjustbox}

\title{B-XAIC Dataset: Benchmarking Explainable AI for Graph Neural Networks Using Chemical Data}

\author{%
Magdalena Proszewska$^{1,2}$\\
\texttt{m.proszewska@ed.ac.uk}
\And
Tomasz Danel$^{2,3}$\\
\texttt{tomasz.danel@uj.edu.pl}
\And
Antoni Antoszek$^{2}$ \\
\texttt{ant.antoszek@student.uj.edu.pl}
\And
Dawid Rymarczyk$^{2,4}$ \\
\texttt{dawid.rymarczyk@uj.edu.pl}
\thanks{$^{1}$School of Informatics, University of Edinburgh, $^{2}$Faculty of Mathematics and Computer Science, Jagiellonian University; $^{3}$ Faculty of Chemistry, Jagiellonian University; $^{4}$ Ardigen SA}
}
\usepackage{xspace}

\newcommand{\ours}{B-XAIC\xspace}

\newtheorem{proposition}{Proposition}

\begin{document}

\maketitle

\begin{abstract}
Understanding the reasoning behind deep learning model predictions is crucial in cheminformatics and drug discovery, where molecular design determines their properties. However, current evaluation frameworks for Explainable AI (XAI) in this domain often rely on artificial datasets or simplified tasks, employing data-derived metrics that fail to capture the complexity of real-world scenarios and lack a direct link to explanation faithfulness. To address this, we introduce B-XAIC, a novel benchmark constructed from real-world molecular data and diverse tasks with known ground-truth rationales for assigned labels. Through a comprehensive evaluation using B-XAIC, we reveal limitations of existing XAI methods for Graph Neural Networks (GNNs) in the molecular domain. This benchmark provides a valuable resource for gaining deeper insights into the faithfulness of XAI, facilitating the development of more reliable and interpretable models.
\end{abstract}

\section{Introduction}

Graph Neural Networks (GNNs) have become the standard for predictive modeling of small molecules~\citep{wieder2020compact}, achieving exceptional performance across property prediction, virtual screening, and related pharmaceutical tasks. While their predictive capabilities are well-established, a growing emphasis is now placed on understanding their reasoning~\citep{jimenez2020drug}. In scientific applications of deep learning to small molecules, transparent explanation mechanisms are not merely desirable but crucial. They build researcher trust, ensure model reliability, and potentially uncover novel insights that can accelerate drug discovery~\citep{wu2023black}.

\begin{figure}[t]
    \centering
    \includegraphics[width=0.76\linewidth]{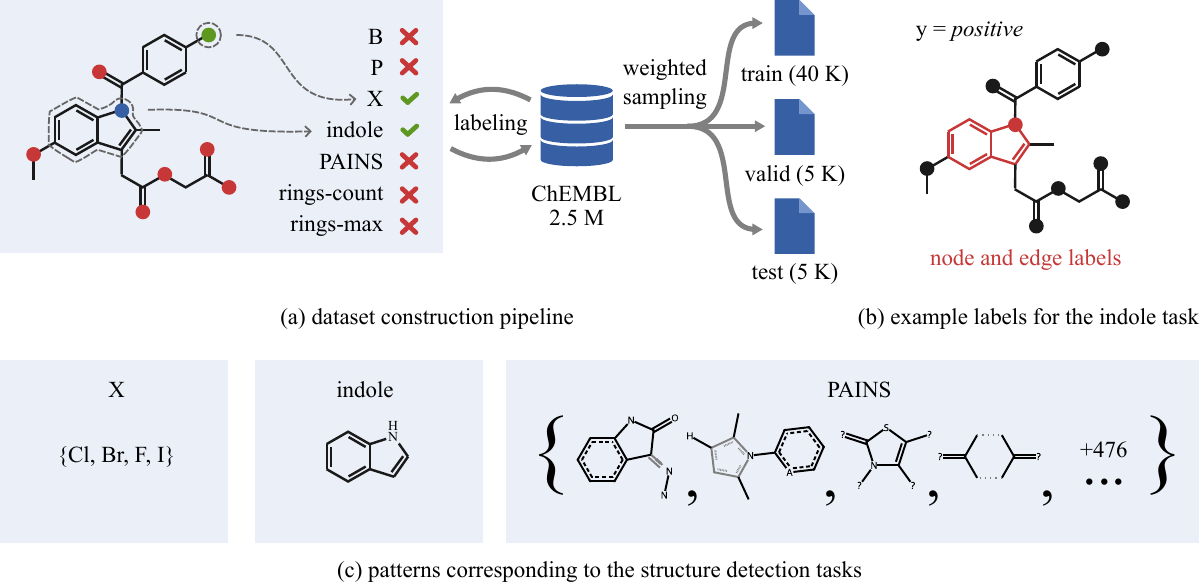}
    \caption{Schematic of our B-XAIC dataset and benchmark; (a) the dataset preparation steps include compound labeling, filtering, and sampling to the training, validation, and testing subsets; (b) for each positive example, atom and bond labels are provided to assess model explanations; (c) the patterns for the halogen and indole tasks are presented, as well as four example PAINS patterns.}
    \label{fig:dataset}
    \vspace{-\baselineskip}
\end{figure}

To address this gap, a range of Explainable AI (XAI) techniques have been adapted or specifically designed for GNNs, aiming to reveal the mechanisms behind their predictions~\citep{jimenez2020drug,kakkad2023surveyexplainabilitygraphneural}. These approaches generally fall into two categories: counterfactual methods~\citep{chen2022greasegeneratefactualcounterfactual,lucic2022cfgnnexplainercounterfactualexplanationsgraph,Tan_2022}, which seek to identify minimal input changes that alter a model's prediction, and factual methods~\citep{luo2020parameterizedexplainergraphneural,schlichtkrull2022interpretinggraphneuralnetworks,ying2019gnnexplainergeneratingexplanationsgraph}, which aim to highlight important substructures within the input graph. Factual methods further diverge into post-hoc explainers \citep{ying2019gnnexplainergeneratingexplanationsgraph}, which analyze a trained black-box model, and inherently interpretable architectures  \cite{feng2022kergnnsinterpretablegraphneural,velickovic2018graph,zhang2021protgnnselfexplaininggraphneural}, which aim for transparency through their design. However, recent findings indicate a critical challenge: regardless of the specific XAI method used, the resulting explanations can be unreliable, or even misleading, potentially interfering with scientific understanding~\cite{faber2021comparing}. Despite the strong performance of these models on established benchmarks and metrics for GNNs and small molecules, such unreliable explanations are still observed.

In response to the limitations of current XAI evaluation for GNNs, the community has developed synthetic datasets~\citep{agarwal2023evaluating,azzolin2023globalexplainabilitygnnslogic,luo2020parameterizedexplainergraphneural,wu2022discoveringinvariantrationalesgraph,ying2019gnnexplainergeneratingexplanationsgraph}. However, these often lack real-world complexity, while real-world ground-truth explanations are hard or impossible to obtain. Existing real-world datasets like MUTAG are small and task-limited. Furthermore, many evaluation methods rely on thresholding importance maps or selecting top-k elements, which can be problematic for tasks dependent on the presence/absence of substructures, where no single element is inherently more important. This arbitrary selection can yield inaccurate explanations and misleading metrics. AUROC~\citep{bajaj2022robustcounterfactualexplanationsgraph,zhang2020relexmodelagnosticrelationalmodel} avoids thresholding but is undefined for empty ground-truth explanations.

To address these limitations, we introduce B-XAIC (Benchmark for eXplainable Artificial Intelligence in Chemistry), a novel benchmark comprising 50K small molecules and 7 diverse tasks, accompanied by both ground-truth labels and corresponding explanations, making accuracy-based metrics a directly applicable and reliable evaluation method. B-XAIC is intentionally designed as a \emph{unit test} for XAI on molecular graphs: simple, well-defined tasks with unambiguous, structure-derived ground truth, in the same spirit as FunnyBirds~\citep{hesse2023funnybirds} in vision XAI. If explainers cannot recover rules demonstrably learned by a model, evaluating them on more complex tasks would be premature. B-XAIC tackles the challenges associated with thresholding explanations or selecting the top-k most important elements in two scenarios: (1) cases where a specific part of the input graph constitutes the explanation, which can be effectively evaluated using AUROC, and (2) cases where the entire graph is equally important for the prediction, in which the evaluation focuses on ensuring the explanation does not contain irrelevant outliers. Ultimately, B-XAIC enables a direct and fair comparison of various XAI approaches, both post-hoc explainers and inherently self-explainable models.

\section{Related Work}

\paragraph{Explainability in GNNs}
Recent research in GNNs has increasingly focused on developing methods to interpret and explain the decisions made by these models. XAI techniques for GNNs can be broadly categorized based on the type of explanation they provide \citep{kakkad2023surveyexplainabilitygraphneural}. These methods may involve identifying key substructures within the input graph that influence the model’s predictions, offering factual explanations by highlighting relevant parts of the input \citep{dai2021selfexplainablegraphneuralnetwork,luo2020parameterizedexplainergraphneural,ying2019gnnexplainergeneratingexplanationsgraph,yuan2021explainabilitygraphneuralnetworks}, or generating counterfactual examples where perturbed inputs lead to a different prediction \citep{chen2022greasegeneratefactualcounterfactual,lucic2022cfgnnexplainercounterfactualexplanationsgraph,Tan_2022}. 

Furthermore, factual methods for explaining GNN predictions can be broadly classified into post-hoc and self-interpretable approaches. Post-hoc methods aim to explain the predictions of a pre-trained GNN by identifying important nodes, edges, or features that influence the model’s decision \citep{ying2019gnnexplainergeneratingexplanationsgraph,luo2020parameterizedexplainergraphneural,schlichtkrull2022interpretinggraphneuralnetworks}. 
Other post-hoc approaches include GraphLIME~\citep{huang2020graphlimelocalinterpretablemodel}, which uses local feature selection to explain node predictions, and surrogate-based methods such
as DnX~\citep{pereira2024distillnexplainexplaining}, which distills the original GNN into a simpler model to generate explanations.
In contrast, self-interpretable methods design the GNN architecture to inherently incorporate explainability using information constraints, such as attention blocks \citep{miao2022interpretablegeneralizablegraphlearning,velickovic2018graph} or bottlenecks \citep{wu2020graphinformationbottleneck}, or integrating structural constraints like prototypes \citep{rymarczyk2023progrest,zhang2021protgnnselfexplaininggraphneural} or graph kernels \citep{Cosmo_2025,feng2022kergnnsinterpretablegraphneural,proszewska2026aiming}, to ensure that the model is more interpretable.


\begin{table}[tb]
\centering
\caption{Comparison of XAI evaluation benchmarks for GNNs in chemistry. \textbf{Size}: number of graphs; \textbf{Tasks}: number of distinct prediction tasks; \textbf{Real mol.}: built from real molecules; \textbf{NE eval}: supports null-explanation (empty ground truth) evaluation; \textbf{Multi-pat.}: tasks with multiple admissible patterns; \textbf{Edge+Node}: provides both node- and edge-level ground truth. \ours is the largest and most diverse, and the only one to support null-explanation evaluation on molecular data.}\label{tab:comp}
\small
\begin{tabular}{@{}lcccccccc@{}}
\toprule
\textbf{Feature} & \textbf{MUTAG} & \textbf{Benzene} & \textbf{Flu.-Carb.} & \textbf{Alk.-Carb.} & \textbf{BA-2Mot.} & \textbf{ShapeGGen} & \textbf{B-XAIC} \\ \midrule
Size       & 0.2K     & 12K  & 8.7K    & 1.1K    & 1K    & config.       & \textbf{50K}    \\
Tasks      & 1  & 1    & 1       & 1       & 1     & 1   & \textbf{7}      \\
Real mol.  & Yes      & Yes  & Yes     & Yes     & No    & No  & \textbf{Yes}    \\
NE eval    & No       & No   & No      & No      & No    & No  & \textbf{Yes}    \\
Multi-pat. & No       & No   & No      & No      & No    & No  & \textbf{Yes}    \\
Edge+Node  & No       & Yes  & Yes     & Yes     & Varies      & Varies    & \textbf{Yes}    \\ \bottomrule
\end{tabular}
\vspace{-2\baselineskip}
\end{table}

\paragraph{Explainability Benchmarks}
Several benchmarks with ground-truth explanations have been proposed to evaluate GNN explainers. Synthetic ones use planted motifs as the ground truth: BA-Shapes, BA-Community, Tree Cycle, and Tree Grids \citep{ying2019gnnexplainergeneratingexplanationsgraph} target node classification, where the model predicts whether a node belongs to a motif such as a cycle, house, or grid, while BA-2Motifs \citep{luo2020parameterizedexplainergraphneural}, BAMultiShapes \citep{azzolin2023globalexplainabilitygnnslogic}, and Spurious Motifs \citep{wu2022discoveringinvariantrationalesgraph} target graph classification, where the model detects given motifs in the whole graph. The more recent ShapeGGen \citep{agarwal2023evaluating} generates diverse synthetic datasets but, like other synthetic benchmarks, misses the complexity and noise of real-world data.

Among real-world datasets, molecular ones are the most common, because ground-truth explanations can be derived from known chemistry. MUTAG \citep{doi:10.1021/jm00106a046}, Benzene, Fluoride-Carbonyl, and Alkane-Carbonyl \citep{doi:10.1073/pnas.1820657116,NEURIPS2020_417fbbf2} (0.2K, 12K, 8.7K, and 1.1K graphs, respectively) are graph classification tasks whose explanations rely on the presence or absence of simple chemical structures. They are widely used, but experimental labels such as those in MUTAG may depend on subtler factors besides known patterns like the nitro group, so the ground truth cannot be assigned with certainty.

Table~\ref{tab:comp} compares \ours with these benchmarks. \ours is the only one that (i) is built from \emph{real} molecules, (ii) supports \emph{null-explanation} evaluation, i.e.\ instances whose correct explanation is empty, (iii) contains \emph{multiple patterns} per task (e.g.\ several PAINS alerts or halogen alternatives), (iv) provides \emph{both} node- and edge-level ground truth, and (v) spans \emph{seven} tasks of graded difficulty. Together, these properties let \ours measure spurious attribution (via null explanations) and pattern localization (via subgraph explanations) separately on chemically grounded data.

\ours combines mechanically defined tasks, such as ring counting and atom detection, with chemically meaningful ones: detecting PAINS, based on the 480 structural alerts for assay interference introduced by Baell and Holloway~\citep{baell2010new}, and detecting the indole scaffold. Following the diagnostic rationale of FunnyBirds~\citep{hesse2023funnybirds}, \ours tests whether explainers recover structural rules, not whether they provide chemically insightful explanations; the former is a prerequisite for evaluating explanations on complex molecular endpoints. We define faithfulness with respect to the label-generating rule: an explanation is faithful if it recovers the substructures used by the model to make its prediction, rather than merely providing a plausible or chemically meaningful rationale \citep{jacovi-goldberg-2020-towards,Nauta_2023}. Since our tasks follow explicit rules, plausibility and faithfulness should coincide once the correct rule is recovered.

Many GNN explainers are also evaluated on more complex datasets without ground-truth explanations, such as NCI1 \citep{4053093}, BBBP \citep{doi:10.1021/ci300124c}, Tox21 \citep{10.3389/fenvs.2015.00080}, or Proteins \citep{10.1093/bioinformatics/bti1007}, and on benchmarks from other domains. MNIST-75sp \citep{knyazev2019understandingattentiongeneralizationgraph} turns images into graphs with superpixels as nodes; Graph SST2, Graph SST5, and Graph Twitter \citep{yuan2022explainabilitygraphneuralnetworks} convert sentiment analysis datasets into graphs; and social graphs such as German Credit, Recidivism, and Credit Defaulter are used to evaluate fairness on sensitive data \citep{agarwal2021unifiedframeworkfairstable}. Because ground-truth explanations are unavailable or impossible to define in these settings, evaluation moves from accuracy to other properties of XAI systems \citep{Nauta_2023}: fidelity \citep{amara2024graphframexsystematicevaluationexplainability,Longa_2025,zheng2024robustfidelityevaluatingexplainability}, sparsity \citep{lucic2022cfgnnexplainercounterfactualexplanationsgraph,yu2022improvingsubgraphrecognitionvariational}, sufficiency and necessity \citep{chen2022greasegeneratefactualcounterfactual,Tan_2022}, robustness \citep{bajaj2022robustcounterfactualexplanationsgraph}, or their combinations~\citep{agarwal2023evaluating,proszewska2026aiming}. These metrics typically measure how the explanation or the predicted class changes when the input graph is perturbed.

\ours fills this gap with 50\,K molecules and 7 tasks, each paired with ground-truth explanations. Evaluation proceeds in two stages: we first check whether a method correctly identifies instances with no relevant nodes, and then how accurately it highlights the relevant subgraph. The methodology could in principle be extended to materials or proteins, but we focus on organic compounds because other domains may require different modeling techniques, which would complicate benchmarking.

\section{B-XAIC Benchmark}

In this section, we first introduce the B-XAIC dataset that is the core of our benchmark, and then provide details on how to evaluate explanations using B-XAIC. The benchmark dataset and code are publicly available at \url{https://huggingface.co/datasets/mproszewska/B-XAIC} and \url{https://github.com/mproszewska/B-XAIC}, respectively.

\subsection{Dataset Construction}

Our benchmark dataset was sourced from ChEMBL 35~\citep{gaulton2012chembl}, which is a public database of 2.5\,M molecules with drug-like properties shared under the CC BY-SA 3.0 license. The molecules were pre-filtered by removing invalid or duplicated SMILES strings. The solvent molecules and counterions were removed to keep only one molecular graph per example.

The benchmark tasks are based on the presence of chemical substructures (see Figure~\ref{fig:dataset}), with increasing difficulty of chemical patterns:

\begin{enumerate}[leftmargin=*]
    \item Detection of organoboron and organophosphorus compounds. The goal of these two tasks is to predict if a compound contains a boron (B) or phosphorus (P) atom, respectively.
    \item Detection of halogens (X). The prediction should be positive if any of the halogen atoms (bromine, chlorine, fluorine, and iodine) are present in the molecule. This task verifies whether GNNs can find one of multiple alternative patterns in the graph.
    \item Detection of indoles, a bicyclic structure with a benzene ring fused to a pyrrole ring. Compounds containing this structure are widely distributed in nature. In this task, a GNN should effectively pass messages between nodes to detect a larger pattern in the graph.
    \item Detection of pan-assay interference compounds (PAINS), which cause false positives in high-throughput screens. We use the list of such patterns proposed by Baell and Holloway~\citep{baell2010new}. This task aims to test whether a GNN is able to learn multiple more complex patterns.
    \item Counting rings. The model should predict if a molecule contains more than four rings. This task involves both detecting a pattern and counting its occurrences.  Spiro, fused, and bridged rings are considered distinct rings. For example, spiro[5.2]octane will be identified as having two rings.
    \item Detecting large rings with more than six atoms. This task involves counting nodes within a substructure. Spiro, fused, and bridged rings are considered distinct rings.
\end{enumerate}

\begin{wrapfigure}[16]{r}{0.53\textwidth}
    \vspace{-1.6\baselineskip}
    \centering
    \includegraphics[width=\linewidth]{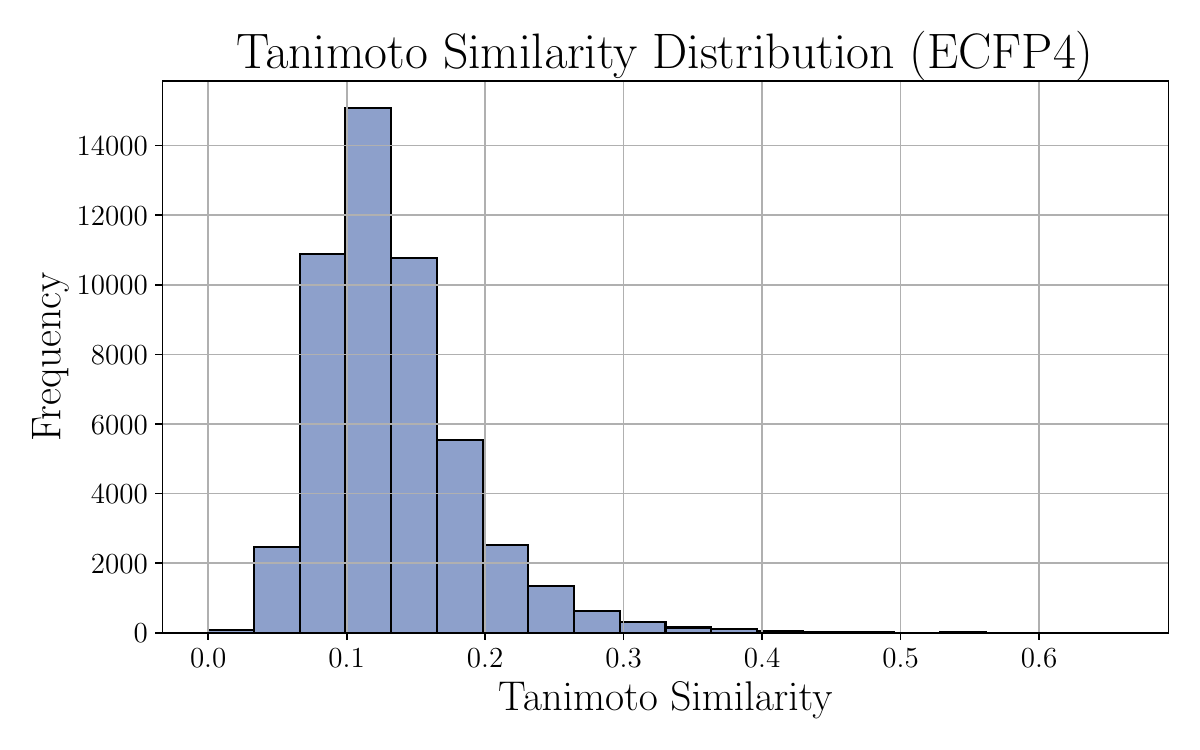}
    \vspace{-1\baselineskip}
    \caption{Histogram of pairwise Tanimoto similarities between molecules in B-XAIC; lower values indicate greater diversity.}
    \label{fig:div}
\end{wrapfigure}

The final dataset is produced by sampling 50\,K molecules from the pre-filtered ChEMBL dataset. We stop at 50\,K because performance saturates there for most tasks (Appendix~\ref{app:dataset-size}) and costly explainers stay affordable, though the pipeline scales to all 2.5\,M molecules. Weighted sampling is used to avoid huge class imbalance. The weights are defined as the product of the ratios between the majority and minority classes for all tasks. The average size of a graph in the resulting dataset is 34.56.  The data is split randomly into training, validation, and testing sets using the 8-1-1 ratio. The dataset contains a set of binary task labels for each compound and two sets of explanation labels, one for the atoms and one for the edges involved in each detected pattern.
Figure~\ref{fig:div} shows the distribution of pairwise Tanimoto similarities between ECFP4 fingerprints (extended-connectivity fingerprints of radius 2); Tanimoto similarity is the ratio of bits shared by two binary fingerprints to bits set in either, ranging from 0 (no shared substructures) to 1 (identical).

\subsection{Evaluation Metrics}

For each task and each graph, the ground-truth explanation is defined as a subset of nodes and edges that are relevant to the task.
\newpage
Explanations are of two types:
\begin{itemize}[leftmargin=*]
    \item \textbf{Null explanations (NE)} -- no nodes or edges are more important than others. For example, in task B, if atom B is not present in the graph, then no specific substructure is considered relevant.
    \item \textbf{Subgraph explanations (SE)} -- only a part of the graph is relevant to the task. For instance, in task B, if atom B is present, that node constitutes the explanation.
\end{itemize}

We evaluate these two groups separately, for both node and edge explanations (Table~\ref{tab:stats}):

\begin{wraptable}[12]{r}{0.55\textwidth}
\vspace{-1.21\baselineskip}
\centering
\scriptsize
\caption{Summary of B-XAIC statistics. SE denotes relevant subgraphs in positive instances, while NE denotes negative instances without explanations.} \label{tab:stats}
\begin{tabular}{l@{\hspace{-2pt}}cccc}
\toprule
\multirow{2}{*}{\shortstack{task}}  & \multirow{2}{*}{\shortstack{\% of graphs \\ with pos. label}} &  \multirow{2}{*}{\shortstack{\% of graphs \\ with NE}} & \multirow{2}{*}{\shortstack{\% of nodes \\ in SE}} & \multirow{2}{*}{\shortstack{\% of edges \\ in SE}} \\
\\
\midrule
B     & 2.23 & 97.77 & 4.1 ± 2.1 & --- \\ 
P     & 13.22 & 86.78 & 4.3 ± 2.4 & --- \\ 
X     &  55.87 & 44.13 & 6.0 ± 4.0 & ---  \\ 
indole  & 36.81 & 63.19 & 31.3 ± 12.1 & 31.4 ± 12.1 \\ 
PAINS & 32.89 & 67.11 & 34.1 ± 14.3 & 31.4 ± 14.7 \\ 
rings-count & 30.12 & 1.34 & 64.0 ± 16.0 & 61.8 ± 14.8 \\ 
rings-max   & 5.68 & 94.32 & 27.1 ± 9.8 & 25.1 ± 9.5 \\ 
\bottomrule
\end{tabular}
\end{wraptable}

\begin{itemize}[leftmargin=*]
    \item \textbf{Null explanations}: A reliable explainer should treat all nodes and edges equally when no pattern is present, so we cast NE evaluation as outlier detection and use the classical interquartile range (IQR) rule with the standard $1.5\times$ multiplier~\citep{Tukey1977}, assigning a score of $1$ if no outliers are detected and $0$ otherwise. The rule is nonparametric, scale-free, and invariant to affine rescaling of attributions (proof in Appendix~\ref{app:iqr-affine}), which makes it robust to per-explainer normalization choices. The same rule correctly handles ``negative reasoning'': uniform attention to recognize a pattern's absence yields near-uniform attributions, not outliers, and is therefore scored well.
    \item \textbf{Subgraph explanations}: Because it may be difficult to find the optimal threshold for explanation methods to extract all relevant nodes or edges, we rely on the AUROC metric to test if the most significant nodes and edges are prioritized over the remaining graph structure.
\end{itemize}

This distinction is important because null instances make up most examples in tasks B and P. An explainer may perform well on SE while producing spurious attributions when the pattern is absent. 
Because B-XAIC labels are generated by explicit structural rules, we control exactly which instances carry a null explanation and which carry a subgraph explanation, and can measure each regime separately instead of inferring it from a single aggregate score. This is not possible on experimental endpoints, where it is unknown whether any substructure is responsible for a label.
We also distinguish thresholding used to \emph{construct} an explanation from that used to \emph{evaluate} one: the IQR threshold only detects attribution outliers and does not define the explanation. 
The complementary role of NE and Gini in diagnosing spurious attributions is discussed in Appendix~\ref{app:ne_motiv}.

\section{Results}

\subsection{Experimental setup}

\paragraph{Explainers.}
We evaluate explanation methods for GNNs grouped into four families.

\textbf{Gradient-based (GB)} methods estimate the influence of input features on a prediction using gradients or modified backpropagation. IntegratedGradients~\citep{sundararajan2017axiomaticattributiondeepnetworks}, Saliency~\citep{simonyan2014deepinsideconvolutionalnetworks}, and Input$\times$Gradient~\citep{shrikumar2017justblackboxlearning} assign importance scores to input features, while Deconvolution~\citep{10.1007/978-3-319-46466-4_8,shrikumar2017justblackboxlearning} and GuidedBackprop~\citep{springenberg2015strivingsimplicityconvolutionalnet} modify gradient propagation to highlight features contributing to the prediction. Because sensitivity is nonzero wherever information can reach the output, we expect dense attributions that rank present patterns highly (good SE) but place peaks on graphs where the pattern is absent (low NE).

\textbf{Graph-specific (GS)} methods identify relevant graph structures and relationships underlying a prediction. GNNExplainer~\citep{ying2019gnnexplainergeneratingexplanationsgraph} and PGExplainer~\citep{luo2020parameterizedexplainergraphneural} learn masks over graph components to identify important substructures, while GraphMask~\citep{schlichtkrull2022interpretinggraphneuralnetworks} learns sparse masks over message-passing computations. PGMExplainer~\citep{vu2020pgmexplainerprobabilisticgraphicalmodel} models dependencies between input features and predictions, FlowX~\citep{gui2023flowxexplainablegraphneural} traces contributions through message-flow pathways, SubgraphX~\citep{yuan2021explainabilitygraphneuralnetworks} searches for influential subgraphs using Shapley values, and GraphSHAP-IQ~\citep{muschalik2025exact} estimates feature interactions in message-passing GNNs. Mask-based objectives reward preserving the prediction rather than localizing the pattern, so we expect sparsity and entropy terms to push masks toward near-binary or near-uniform solutions, with NE and SE diverging strongly across methods.

\textbf{Perturbation-based (PB)} methods estimate the contribution of input components by examining their effects on model predictions. ShapleyValueSampling~\citep{trumbelj2010AnEE} approximates Shapley values to quantify the contributions of nodes or edges, while EdgeSHAPer~\citep{mastropietro2022edgeshaper} specifically estimates the contributions of edges. These methods reveal which components contribute most to a prediction through their marginal effects. They should work well on single-atom patterns (B, P, X), but masking atoms produces off-distribution molecules and sampling is noisy, so we expect spurious attributions on negative graphs and weaker results on patterns that require several atoms jointly (indole, rings).

\textbf{Surrogate (SU)} methods explain predictions using a simpler model trained to approximate the original GNN. DnX and FastDnX~\citep{pereira2024distillnexplainexplaining} distill the original GNN into a surrogate GNN and extract node- or edge-level explanations through a convex optimization problem. FastDnX exploits the linear decomposition of the surrogate to accelerate explanation generation. These methods aim to identify relevant graph components while reducing the computational cost of explaining the original model. Their explanations are faithful to the surrogate rather than to the GNN, so quality may drop where the surrogate cannot mimic the model (ring-based tasks), and a linear decomposition cannot express absence, which may yield low NE.

\paragraph{Models.} 
We apply explainers to popular GNN architectures: GCN~\citep{kipf2017semisupervisedclassificationgraphconvolutional}, GAT~\citep{velickovic2018graph}, GIN~\citep{xu2019powerfulgraphneuralnetworks}, and ProtGNN~\citep{zhang2021protgnnselfexplaininggraphneural}, a prototype-based, interpretable GNN, instantiated with GCN, GAT, and GIN backbones. Results for GIN are reported here; explanation scores for GCN, GAT, and ProtGNN variants are given in Tables~\ref{tab:node-details} and~\ref{tab:edge-details} in the Appendix.

\subsection{Benchmarking}

Our benchmark evaluation incorporates multiple GNN architectures, with classification metrics for the designated tasks presented in Table~\ref{tab:f1}. While all evaluated methods demonstrate strong F1 scores across most tasks, GIN consistently outperforms alternative architectures.

\begin{table}[t]
\centering
\caption{Weighted-F1 and macro-F1 (\%) across tasks for GCN, GAT, GIN,
and ProtGNN with GIN backbone. Results are reported as mean $\pm$ standard
deviation; the highest scores are highlighted in bold along with the numbers
that are not significantly lower according to the one-sided Wilcoxon test.}
\label{tab:f1}
\small
\setlength{\tabcolsep}{4pt}
\renewcommand{\arraystretch}{1.15}
\begin{adjustbox}{width=\linewidth}
\begin{tabular}{lcc cc cc cc}
\toprule
& \multicolumn{2}{c}{\textbf{GCN}}
& \multicolumn{2}{c}{\textbf{GAT}}
& \multicolumn{2}{c}{\textbf{GIN}}
& \multicolumn{2}{c}{\textbf{ProtGNN+GIN}} \\
\cmidrule(lr){2-3}
\cmidrule(lr){4-5}
\cmidrule(lr){6-7}
\cmidrule(lr){8-9}
\textbf{Task}
& Weighted-F1 & Macro-F1
& Weighted-F1 & Macro-F1
& Weighted-F1 & Macro-F1
& Weighted-F1 & Macro-F1 \\
\midrule
B
& \textbf{99.94} $\pm$ \textbf{0.05} & \textbf{99.82} $\pm$ \textbf{0.19}
& 99.11 $\pm$ 0.19 & \textbf{99.77} $\pm$ \textbf{0.15}
& \textbf{99.96} $\pm$ \textbf{0.05} & 99.50 $\pm$ 0.04
& 96.98 $\pm$ 0.69 & 82.10 $\pm$ 10.45 \\

P
& \textbf{99.98} $\pm$ \textbf{0.02} & \textbf{99.92} $\pm$ \textbf{0.03}
& \textbf{99.97} $\pm$ \textbf{0.01} & \textbf{99.93} $\pm$ \textbf{0.08}
& \textbf{99.98} $\pm$ \textbf{0.03} & \textbf{99.94} $\pm$ \textbf{0.07}
& 99.77 $\pm$ 0.14 & 99.10 $\pm$ 0.23 \\

X
& \textbf{99.84} $\pm$ \textbf{0.05} & \textbf{99.99} $\pm$ \textbf{0.02}
& 99.18 $\pm$ 0.23 & 99.85 $\pm$ 0.10
& \textbf{99.94} $\pm$ \textbf{0.02} & \textbf{99.94} $\pm$ \textbf{0.02}
& 98.74 $\pm$ 0.56 & 98.55 $\pm$ 1.11 \\

indole
& 88.33 $\pm$ 1.74 & 83.76 $\pm$ 0.95
& 67.30 $\pm$ 2.21 & 67.18 $\pm$ 1.79
& \textbf{98.32} $\pm$ \textbf{0.36} & \textbf{98.19} $\pm$ \textbf{0.38}
& 95.76 $\pm$ 0.89 & 95.55 $\pm$ 0.96 \\

PAINS
& 79.89 $\pm$ 0.47 & 76.10 $\pm$ 0.42
& 65.02 $\pm$ 1.65 & 60.67 $\pm$ 0.97
& \textbf{92.90} $\pm$ \textbf{0.54} & \textbf{91.94} $\pm$ \textbf{0.63}
& 85.88 $\pm$ 4.26 & 62.63 $\pm$ 19.85 \\

rings-count
& 87.27 $\pm$ 1.61 & 82.75 $\pm$ 0.70
& 82.88 $\pm$ 0.59 & 79.92 $\pm$ 0.95
& \textbf{99.62} $\pm$ \textbf{0.21} & \textbf{99.54} $\pm$ \textbf{0.25}
& 83.48 $\pm$ 0.69 & 78.65 $\pm$ 0.69 \\

rings-max
& 91.25 $\pm$ 1.13 & 64.45 $\pm$ 2.16
& 91.03 $\pm$ 0.79 & 63.55 $\pm$ 1.19
& \textbf{92.98} $\pm$ \textbf{0.84} & \textbf{70.01} $\pm$ \textbf{2.22}
& 91.63 $\pm$ 0.19 & 57.55 $\pm$ 1.25 \\
\bottomrule
\end{tabular}
\end{adjustbox}
\vspace{-1.5\baselineskip}
\end{table}

As anticipated, the detection of PAINS patterns emerges as the most challenging, requiring the identification of various alert substructures. Several architectures also exhibit limitations in recognizing indole rings, suggesting insufficient capacity to capture extensive substructures within molecular graphs. The ring-counting task similarly presents difficulties for most models, particularly ProtGNN, which is not capable of highlighting disconnected molecular fragments~\cite{elhadri2025looks}. The weighted- vs. macro-F1 gap on rings-count and rings-max reflects MPNNs' limitations in cycle counting~\cite{chen2020graphneuralnetworkscount}.

Given our benchmark's primary focus on comparing XAI methods, subsequent analysis will emphasize results from the GIN architecture, which achieves near-perfect performance across our synthetic tasks. This exceptional performance suggests that GIN formulates predictions based on appropriate chemical principles, making it an ideal candidate for our explainability evaluations.

\paragraph{Node explanations.} First, we will focus on node explanations. Table~\ref{tab:exp-scores} shows the results of various XAI methods applied to GIN, our best-performing model. The reported evaluation metrics are averaged across all tasks, and all best results that are not statistically significantly worse than the highest number (according to a one-sided Wilcoxon test) are highlighted in bold. Gradient-based methods are, on average, better at localizing important patterns than other methods. However, they tend to highlight molecular fragments even when the pattern is absent, resulting in low NE scores.
We report SE and NE as the primary results, with the average score included only as a summary statistic, since a high average can mask weak performance in one of the two settings.

\begin{table}[tb]
\centering
\caption{Evaluation of the node and edge explanations for the GIN model. Explainers are grouped as gradient-based (GB), graph-specific (GS), perturbation-based (PB), and surrogate (SU). Bold marks the best score and those not significantly worse (one-sided Wilcoxon test). Avg is the mean of NE and SE; uncertainty is reported for NE and SE.}
\label{tab:exp-scores}
\setlength{\tabcolsep}{10.5pt}
\scriptsize
\begin{tabular}{llcccccc}
\toprule
 & & \multicolumn{3}{c}{nodes} & \multicolumn{3}{c}{edges} \\
 \cmidrule(r){3-5} \cmidrule(l){6-8}
Class & Explainer      & NE  & SE  & avg  & NE & SE & avg   \\
\midrule
GB & Saliency       & 0.51$\pm$0.12 & 0.85$\pm$0.12 & 0.68 & 0.31$\pm$0.15 & 0.67$\pm$0.06 & 0.49 \\
   & Deconvolution  & 0.79$\pm$0.10 & 0.81$\pm$0.09 & \textbf{0.80} & 0.38$\pm$0.20 & 0.69$\pm$0.04 & 0.53 \\
   & InputXGradient & 0.54$\pm$0.14 & 0.83$\pm$0.14 & 0.68 & 0.37$\pm$0.14 & 0.65$\pm$0.05 & 0.51 \\
   & GuidedBackprop & 0.41$\pm$0.12 & \textbf{0.87$\pm$0.09} & 0.64 & 0.30$\pm$0.08 & 0.70$\pm$0.10 & 0.50 \\
   & IntegratedGradients  & 0.39$\pm$0.26 & 0.85$\pm$0.13 & 0.62 & 0.32$\pm$0.18 & 0.70$\pm$0.06 & 0.51 \\
\midrule
GS & GNNExplainer   & 0.68$\pm$0.12 & 0.67$\pm$0.08 & 0.68 & \textbf{0.56$\pm$0.12} & 0.59$\pm$0.03 & 0.58 \\
   & GraphMaskExplainer   & 0.66$\pm$0.03 & 0.66$\pm$0.07 & 0.66 & 0.35$\pm$0.04 & 0.54$\pm$0.01 & 0.45 \\
   & PGExplainer    &    --- &    --- &     ---      & 0.02$\pm$0.02 & \textbf{0.72$\pm$0.07} & 0.37 \\
   & PGMExplainer   & \textbf{0.98$\pm$0.02} & 0.76$\pm$0.18 & \textbf{0.87} &    --- &      ---     &     ---      \\
   & FlowX    & 0.70$\pm$0.13 & 0.77$\pm$0.15 & 0.73 & \textbf{0.54$\pm$0.07} & 0.64$\pm$0.03 & \textbf{0.59} \\
    & SubgraphX & \textbf{0.94$\pm$0.05} & 0.74$\pm$0.10 & \textbf{0.84} &   --- & --- & --- \\
    & GraphSHAP-IQ & 0.64$\pm$0.09 & 0.72$\pm$0.25 & 0.68 & --- & --- & --- \\
\midrule
PB & ShapleyValueSampling & 0.48$\pm$0.23 & 0.83$\pm$0.15 & 0.65 & 0.18$\pm$0.11 & 0.64$\pm$0.03 & 0.41 \\
& EdgeSHAPer & --- & --- & --- & 0.20$\pm$0.11 & 0.58$\pm$0.09 & 0.39 \\
\midrule
SU & DnX & 0.22$\pm$0.24 & 0.60$\pm$0.16 & 0.41 & --- & --- & --- \\
& FastDnX & 0.12$\pm$0.15  & 0.69$\pm$0.28 & 0.41 & --- & --- & --- \\
\bottomrule
\end{tabular}
\vspace{-\baselineskip}
\end{table}

The detailed results for each task are shown in Figure~\ref{fig:node} and further detailed in the Appendix. Two groups of explanation methods are formed. The first one achieves good NE scores and obtains lower SE scores, and the other group exhibits the opposite behavior. This suggests that some methods provide more contrastive and precise explanations, while other methods return more uniform attributions. The tasks of finding boron, phosphorus, and halogen atoms are the easiest for all the methods. The most difficult patterns to find are those related to rings, either counting them or measuring their size. Class balance differs sharply across tasks: null instances make up 97.8\% of graphs for B but only 1.3\% for rings-count (Table~\ref{tab:stats}). We therefore report NE and SE separately for each task. An explainer that returns uniform attributions gains on NE but loses on SE.

\begin{figure}[tb]
    \centering
    \scriptsize
    \includegraphics[width=0.9\linewidth]{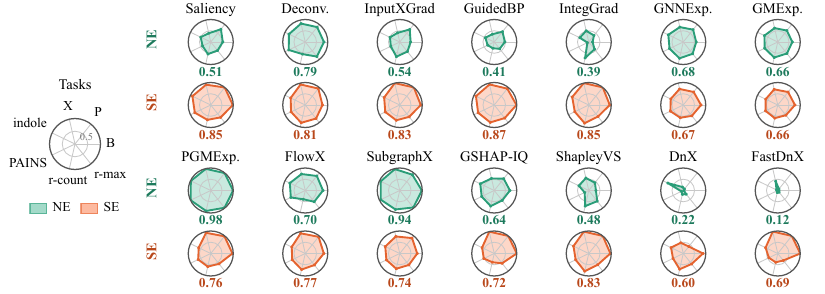}
    \vspace{-0.5\baselineskip}
    \caption{Evaluation of node-level explanations for GIN. Null explanation results are shown in green, and subgraph explanation in orange. Average scores for each method are displayed below.}
    \label{fig:node}
    \vspace{-\baselineskip}
\end{figure}

\paragraph{Edge explanations.} A similar analysis was conducted for edges, and the results are shown in Table~\ref{tab:exp-scores}. In this case, gradient-based methods do not always outperform the other methods in terms of the SE metric. The strong performance of the methods based on subgraph extraction may be caused by the fact that edges are included in the extracted subgraphs, while gradient methods often focus more on nodes. Interestingly, GNNExplainer and FlowX significantly outperform all other methods in the NE metric, while PGExplainer is best at detecting important edges (one-sided Wilcoxon test).

\begin{figure}[t]
    \centering
    \scriptsize
    \includegraphics[width=\linewidth]{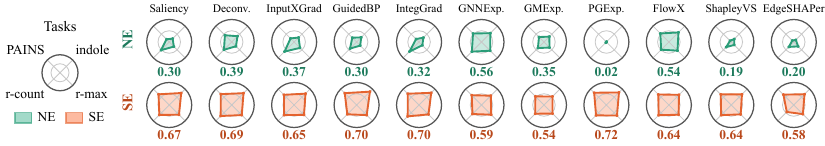}
    \vspace{-1.5\baselineskip}
     \caption{Evaluation of edge-level explanations for GIN. Null explanation results are shown in green, and subgraph explanation in orange. Average scores for each method are displayed below.}
    \label{fig:edge}
    \vspace{-2\baselineskip}
\end{figure}

The results of all explainers are presented in Figure~\ref{fig:edge}. In these results, we focus on four tasks that involve edges, excluding all tasks aiming at detecting single atoms. Also in the case of edges, tasks related to counting rings or atoms in rings appear to be more challenging for the explainers.

\begin{wraptable}[18]{r}{0.3\textwidth}
\vspace{-0.1\baselineskip}
\centering
\caption{Model F1 scores on the original molecules and after substituting true atoms with carbon.}
\label{tab:carbon_sub}
\vspace{-0.5\baselineskip}
\small
\begin{adjustbox}{width=\linewidth}
\begin{tabular}{lcc}
\toprule
Task & Original & Substituted \\
\midrule
B & 99.80±0.24 & 0.00±0.00 \\
P & 99.91±0.11 & 0.86±1.57 \\
X & 99.97±0.02 & 0.18±0.20 \\
indole & 99.33±0.13 & 8.55±2.18 \\
PAINS & 94.02±1.00 & 39.63±3.91 \\
\bottomrule
\end{tabular}
\end{adjustbox}
\vspace{1.7\baselineskip}
\caption{Out-of-Distribution Performance}
\label{tab:ood_performance}
\vspace{-0.5\baselineskip}
\small
\begin{adjustbox}{width=\linewidth}
\begin{tabular}{lccc}
\toprule
Task & ZINC & OOD$_{\rm{f}}$ & OOD \\
\midrule
X & 100\% & 99\% & 99\% \\
B & 100\% & 100\% & 100\% \\
P & 99\% & 100\% & 99\% \\
indole & 98\% & 98\% & 100\% \\
PAINS & 84\% & --- & --- \\
rings-count & 99\% & --- & --- \\
rings-max & 93\% & --- & --- \\
\bottomrule
\end{tabular}
\end{adjustbox}
\end{wraptable}

\begin{figure}[tb]
    \centering
    \includegraphics[width=0.8\linewidth]{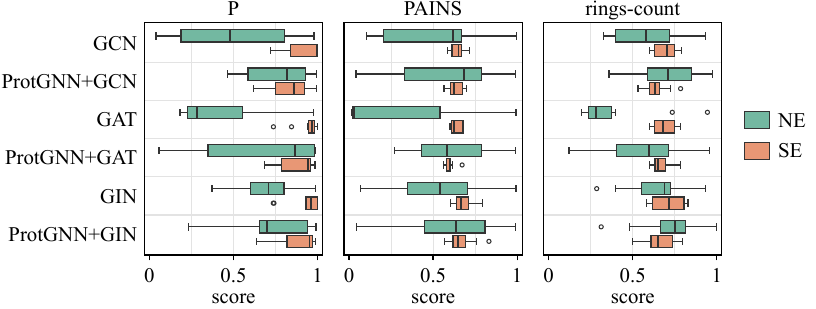}
  \vspace{-0.5\baselineskip}
    \caption{Node-level NE and SE score distributions across explainers, aggregated by model, highlighting that some models are inherently harder to explain.}
    \label{fig:boxplots}
    \vspace{-1.3\baselineskip}
\end{figure}

\paragraph{Relationship between model performance and explanation quality.}
As illustrated in Figure~\ref{fig:boxplots}, the correlation between model performance and explanation quality varies across tasks. At the node level, Pearson correlations are near zero for the atom-detection tasks B, P, and X (0.048, -0.128, and 0.199, respectively), but substantially stronger for the structural tasks indole, PAINS, rings-count, and rings-max (0.800, 0.504, 0.932, and 0.728, respectively). At the edge level, the relationship is less consistent across tasks. The full Pearson correlation results and remaining boxplots are reported in Appendix~\ref{app:f1_explanation_correlations}.

\paragraph{Showcasing that models do not overfit to the chemical space}

To test if our models learned to identify chemical patterns defining the data class, we conducted probing experiments by changing important ground-truth atoms to carbons (excluding ring tasks). The results in Table~\ref{tab:carbon_sub} show a dramatic drop in F1 scores (calculated with the same labels but with probed structure), consistent with the model relying on the ground-truth atoms for B, P, X and indole, and to a lesser degree for PAINS. Replacing them with carbon removes the information required for a correct prediction. 

To further validate this, we conducted an additional out-of-distribution (OOD) evaluation using molecules from the ZINC database, which represents a different distribution of commercially available compounds compared to our ChEMBL-based training data. The strong performance on this external dataset demonstrates genuine cross-dataset generalization (see~Table~\ref{tab:ood_performance}).

We take it a step further by generating synthetic OOD datasets with randomly created atom combinations, which provide the most rigorous generalization test. These molecules follow entirely different distributional properties while maintaining the target pattern recognition task structure. The first dataset, called “OOD$_{\rm{f}}$,” includes random atom combinations based on the empirical distribution of atom frequencies in organic molecules (e.g., carbon atoms make up 75\% of heavy atoms). The second dataset, called “OOD,” samples atoms with equal probability, resulting in completely invalid molecules, of which 50\% contain the pattern of interest. In both cases, our GNN recognizes these patterns almost perfectly despite the extremely OOD samples. 

These results establish that the models have sufficient capacity to learn the underlying patterns and generalize beyond the chemical space used for training. Therefore, the subsequent analysis of explanation quality can be interpreted as evaluating the explainers' ability to recover the information used by the model, rather than the models' inability to learn the underlying task.

\paragraph{Disentangling model capacity from explainer expressiveness}
We conducted experiments evaluating XAI methods in both SE and NE regimes, but only on instances where the GNN model made correct predictions. This approach reduces the influence of model errors on explanation scores.
As shown in Table~\ref{tab:node_explanation_corr}, which reports only the differences in node explanation performance between all predictions and correctly classified predictions, XAI methods exhibit consistent performance across the two evaluation settings, with absolute differences of at most 0.06, except for Deconvolution on PAINS (SE). The full node and edge explanation scores are provided in the Appendix \ref{ap:corr_explanations}.
Additionally, our GNN models consistently achieved high accuracy across all tasks. This high baseline performance confirms that the models possess sufficient capacity to learn the underlying chemical principles. Therefore, the observed variations in XAI effectiveness can be attributed to the XAI technique itself, rather than limitations in the GNN's ability to learn the task.

\begin{table}[tb]
\centering
\small
\caption{Difference in node explanation performance between all predictions
and correctly classified predictions from GIN, computed as
$\Delta M = M_{\text{all}} - M_{\text{correct}}$.}
\label{tab:node_explanation_corr}
\begin{adjustbox}{width=0.9\linewidth}
\begin{tabular}{llcccccccccccccc}
\toprule
Task & Type & \rotatebox{90}{Saliency} & 
\rotatebox{90}{Deconv.} & 
\rotatebox{90}{InputXGrad} & 
\rotatebox{90}{GuidedBP} & 
\rotatebox{90}{IntegGrad} & 
\rotatebox{90}{GNNExp.} & 
\rotatebox{90}{GMExp.} & 
\rotatebox{90}{PGMExp.} & 
\rotatebox{90}{FlowX} & 
\rotatebox{90}{SubgraphX} & 
\rotatebox{90}{ShapleyVS} & 
\rotatebox{90}{GSHAP-IQ} & 
\rotatebox{90}{DnX} & 
\rotatebox{90}{FastDnX} \\
\midrule
\multirow{2}{*}{\shortstack[c]{PAINS}} & NE
& -0.03 & 0.00 & 0.00 & 0.00 & 0.00
& 0.00 & 0.00 & 0.00 & -0.02 & 0.00 & 0.00 & 0.00 & 0.00 & 0.00 \\

 & SE
& -0.01 & -0.26 & 0.00 & -0.01 & 0.00
& 0.00 & 0.00 & 0.00 & 0.00 & -0.02 & 0.00 & 0.00 & 0.00 & 0.00 \\

\midrule
\multirow{2}{*}{\shortstack[c]{rings-max}} & NE
& 0.00 & 0.00 & 0.01 & -0.02 & -0.06
& 0.00 & 0.00 & 0.00 & 0.00 & 0.00 & 0.01 & 0.00 & 0.00 & 0.00 \\

 & SE
& 0.00 & -0.02 & 0.01 & 0.00 & 0.00
& -0.01 & 0.01 & 0.02 & 0.00 & -0.03 & -0.02 & 0.00 & -0.02 & 0.02 \\
\bottomrule
\end{tabular}\end{adjustbox}
\vspace{-0.9\baselineskip}
\end{table}

\begin{figure*}
    \centering
    \includegraphics[width=0.3\linewidth, trim=0cm 21cm 30cm 0cm, clip]{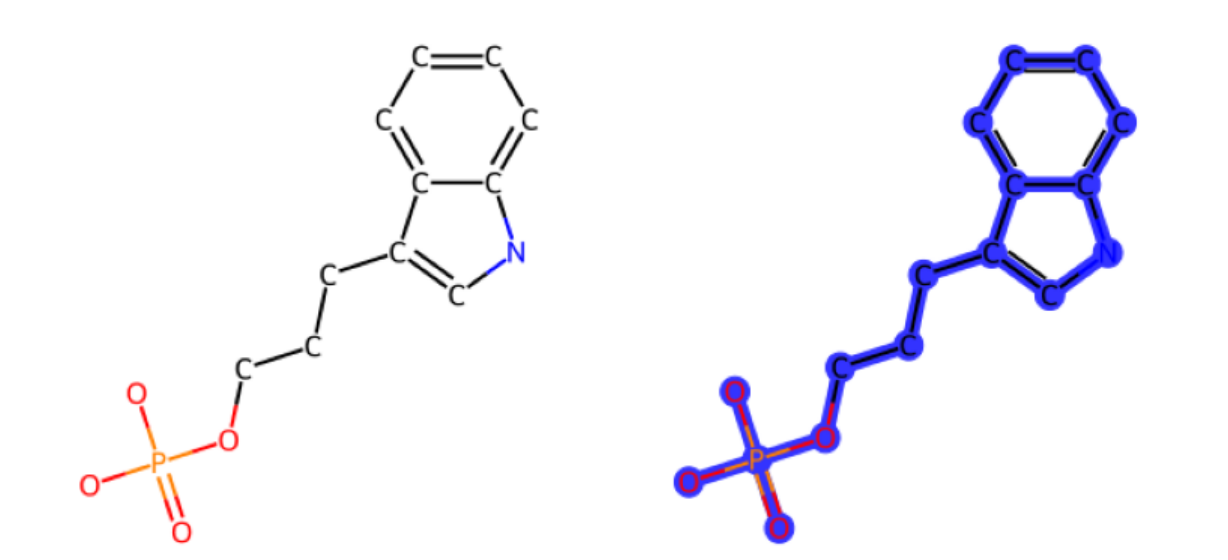} 
     \includegraphics[width=0.3\linewidth, trim=0cm 21cm 30cm 0cm, clip]{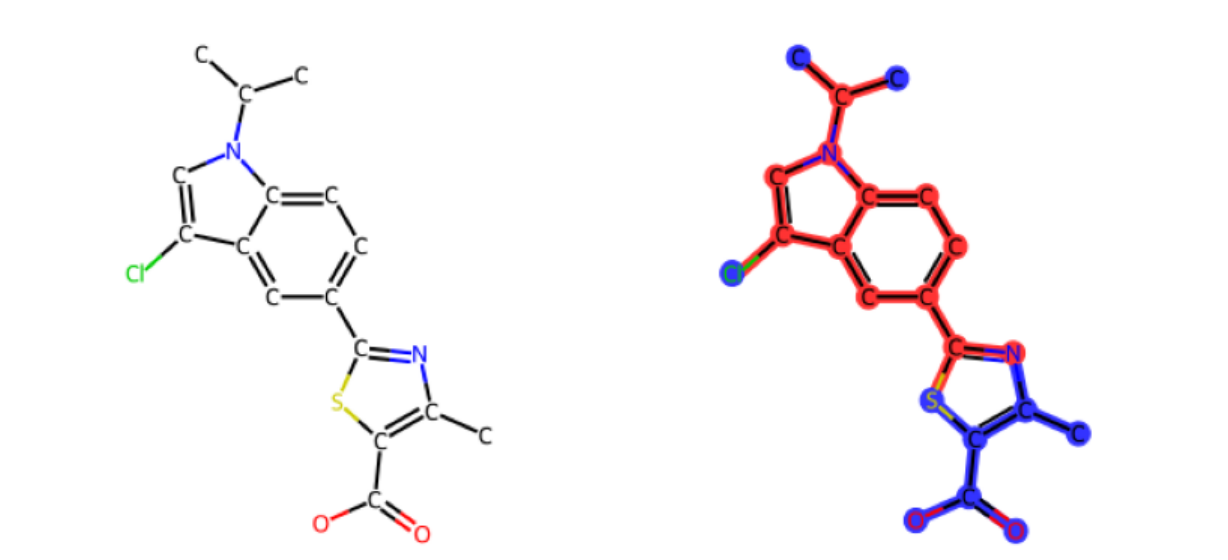} \\
     \vspace{-0.2\baselineskip}
    \includegraphics[width=\linewidth, trim=0cm 0cm 0 10cm, clip]{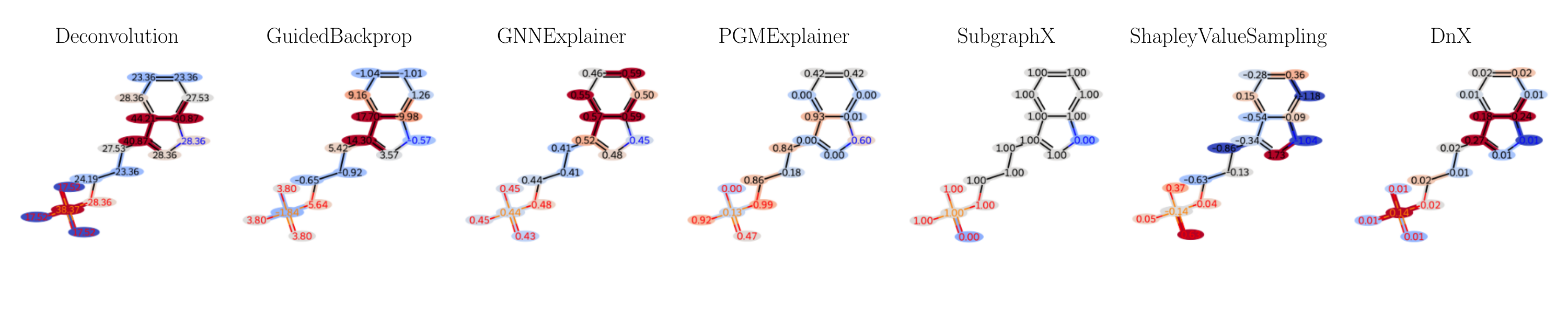} \\
    \vspace{-1.5\baselineskip}
     \includegraphics[width=\linewidth, trim=0cm 0cm 0 10cm, clip]{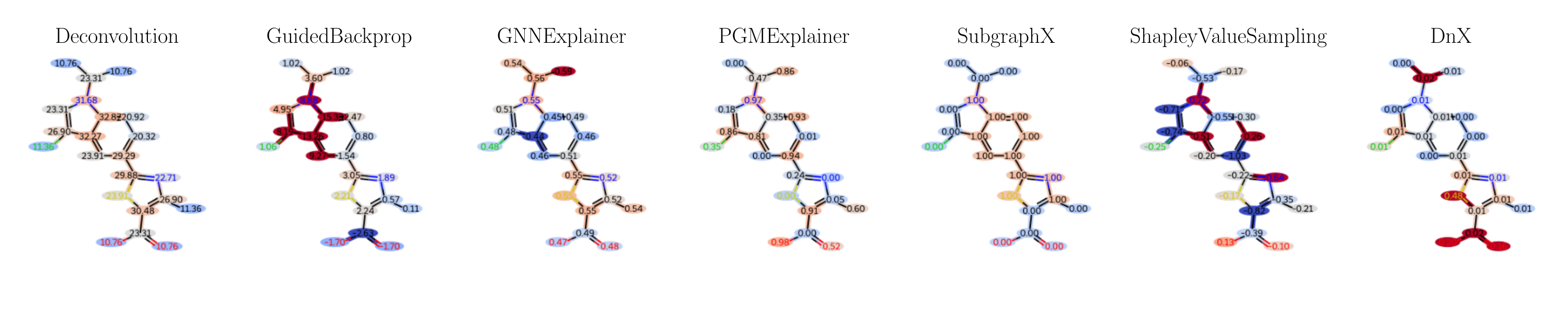}\\
     \vspace{-\baselineskip}
     \caption{Node-level explanations for graphs from both classes in the \textbf{PAINS} task using GIN. Top: input graphs and ground-truth explanations; below: explainer outputs (all explainers in Appendix~\ref{ap:examples}). Red/blue: included/excluded nodes; intermediate colors with explainer scores. PGMExplainer and SubgraphX perform best on the negative example; gradient-based methods on the positive one.}
    \label{fig:examples}
\end{figure*}

\paragraph{Explanation examples} Figure~\ref{fig:examples} illustrates representative explanations for both positive and negative graph instances. We observe that some techniques tend to highlight atoms proximal to the relevant subgraph, potentially due to limited control over the message-passing mechanism in GNNs. Additionally, we observe that even methodologically similar explanation approaches can generate markedly divergent explanations for the same graph. For the negative instances, there is no universal threshold that can be used across all methods to separate important nodes because one method can attribute weights near zero uniformly for all the nodes, while another method predicts uniform values around 4.5. In both cases no subgraph can be highlighted as predicted to be more significant. All these observations lead to the conclusion that widely used explainers struggle to highlight even simple patterns for GNNs that achieve almost perfect accuracy. This emphasizes the immense need for benchmarks like B-XAIC to accelerate research on new XAI methods for graphs.

\subsection{Discussion}

Our findings clearly demonstrate the critical need for new XAI benchmarks specifically tailored to molecular graphs. Current XAI techniques exhibit significant deficiencies in generating adequate explanations, even for the most elementary tasks proposed in our B-XAIC benchmark. Despite the GIN model achieving remarkably high performance metrics, with F1 scores exceeding 92\% for all proposed tasks, the explanations generated by the explainers consistently fail to properly identify and highlight the relevant molecular structures. Our results reveal three key failure modes.

First, \textit{sensitivity-based attribution struggles to represent feature absence}. Most gradient-based and surrogate methods locate the pattern well (high SE) but still highlight spurious fragments when no pattern is present (low NE), because the output remains unevenly sensitive to atoms even when none of them is relevant.
Second, \textit{node and edge explanations can contradict each other within the same method}. On null instances, Deconvolution's node attributions contain no outliers in most molecules, yet its edge attributions for the same molecules highlight spurious bonds.
Third, \textit{near-binary attributions may be insensitive to the evaluation threshold}. GraphMaskExplainer produces near-binary attribution scores, which can limit the extent to which changing the threshold affects the selected explanatory components. This behavior highlights the importance of considering attribution distributions when evaluating explanation quality.

While complex nonlinear interactions between atoms characterize real-world chemical applications, our benchmark reveals that these explanation failures persist even in comparatively simple tasks. The iterative message-passing layers inherent to GNNs result in information diffusion among neighboring nodes, potentially impeding precise localization of salient features. These findings highlight the need for novel approaches that can maintain predictive power while enhancing interpretability.

\paragraph{Limitations.}
\ours is a deliberately scoped \emph{unit test} for molecular GNN explainers, and several limitations follow from that scope. First, its ground truth is \emph{rule-derived} rather than expert-derived: \ours measures whether an explainer recovers the label-generating structural rule, not whether it produces chemically insightful explanations, following the diagnostic rationale of FunnyBirds~\citep{hesse2023funnybirds}. If an explainer cannot recover an unambiguous rule that the model demonstrably uses, evaluating it on complex endpoints is premature. PAINS is the expert-curated exception, as its $480$ alerts~\citep{baell2010new} carry real screening consequences. Second, all tasks are \emph{binary classification}; regression-style targets (e.g.\ continuous ring statistics or property prediction) are left to future work. Third, the benchmark is \emph{graph-level} and restricted to small, drug-like molecules (up to $\sim\!60$ atoms), so it is not aimed at large-graph or node-level explainers. Fourth, \ours admits a useful \emph{duality}: explainers can be evaluated either on a trained GNN, where a low score may reflect either the explainer or the model, or on the exact analytical game $\nu$ (Appendix~\ref{app:vt}), which is model-independent but abstracts away the learned representation; the two views are complementary. We mitigate the trained-model confound with out-of-distribution and probing experiments (Tables~\ref{tab:carbon_sub} and~\ref{tab:ood_performance}), evaluation restricted to correct predictions (Appendix~\ref{ap:corr_explanations}), and the analytical game, but cannot fully guarantee that the model relies only on the intended substructure. Finally, the harder tasks are \emph{class-imbalanced} (e.g.\ \texttt{rings-max} has $5.68\%$ positives, Table~\ref{tab:stats}), so per-task results should be read with their balance in mind, and the chemical complexity covered, while graded, is bounded by the requirement that each label admit an unambiguous structural rule. Consequently, a holistic evaluation that considers both explanation-quality metrics and the model's predictive accuracy is recommended when using this benchmark.

\section{Conclusions}

In conclusion, the B-XAIC dataset offers a valuable new resource for the GNN XAI community. By providing real-world molecular data with structure-derived ground-truth explanations, our dataset enables rigorous benchmarking of both inherently interpretable graph models and post-hoc GNN explainers. The introduction of null and subgraph explanation concepts, along with the edge-based and node-based variations, offers a more nuanced evaluation of XAI capabilities across different explanation types and graph aspects. We believe B-XAIC will serve as a crucial baseline for future research, clearly highlighting the strengths and limitations of emerging XAI methods.  Our ongoing work aims to further enrich this dataset by incorporating activity-cliff scenarios, pushing the boundaries of XAI techniques to uncover subtle but critical distinctions within graph data.


\section*{Acknowledgements}

We thank Jacek Tabor, Marek Śmieja and Marcin Sendera for helpful discussions at early stages of this project.
The work of M. Proszewska was funded by National Centre of Science (Poland) grant no. 2023/49/B/ST6/01137.
The work of T. Danel was carried out within the “AI-Based Virtual Screening Assistant” project (LIDER15/0033/2024) funded by the National Center for Research and Development (Poland) under the LIDER XV program.
The work of D. Rymarczyk was funded by "Interpretable and Interactive Multimodal Retrieval in Drug Discovery" project. The „Interpretable and Interactive Multimodal Retrieval in Drug Discovery” project (FENG.02.02-IP.05-0040/23) is carried out within the First Team programme of the Foundation for Polish Science co-financed by the European Union under the European Funds for Smart Economy 2021-2027 (FENG).
Some experiments were performed on servers purchased with funds from the Priority Research Area (Artificial Intelligence Computing Center Core Facility) under the Strategic Programme Excellence Initiative at Jagiellonian University.

\bibliographystyle{abbrv}
\bibliography{neurips2026}

@article{kakkad2023surveyexplainabilitygraphneural,
  title={A Survey on Explainability of Graph Neural Networks},
  author={Kakkad, Jaykumar and Jannu, Jaspal and Sharma, Kartik and Aggarwal, Charu and Medya, Sourav},
  journal={Bulletin of the IEEE Computer Society Technical Committee on Data Engineering},
  volume={47},
  number={2},
  year={2023},
  publisher={IEEE}
}

@inproceedings{ying2019gnnexplainergeneratingexplanationsgraph,
  author    = {Zhitao Ying and
               Dylan Bourgeois and
               Jiaxuan You and
               Marinka Zitnik and
               Jure Leskovec},
  editor    = {Hanna M. Wallach and
               Hugo Larochelle and
               Alina Beygelzimer and
               Florence d'Alch{\'{e}}{-}Buc and
               Emily B. Fox and
               Roman Garnett},
  title     = {GNNExplainer: Generating Explanations for Graph Neural Networks},
  booktitle = {Advances in Neural Information Processing Systems 32: Annual Conference
               on Neural Information Processing Systems 2019, NeurIPS 2019, December
               8-14, 2019, Vancouver, BC, Canada},
  pages     = {9240--9251},
  year      = {2019},
  url       = {https://proceedings.neurips.cc/paper/2019/hash/d80b7040b773199015de6d3b4293c8ff-Abstract.html},
  bibsource = {dblp computer science bibliography, https://dblp.org}
}

@inproceedings{luo2020parameterizedexplainergraphneural,
  author    = {Dongsheng Luo and
               Wei Cheng and
               Dongkuan Xu and
               Wenchao Yu and
               Bo Zong and
               Haifeng Chen and
               Xiang Zhang},
  editor    = {Hugo Larochelle and
               Marc'Aurelio Ranzato and
               Raia Hadsell and
               Maria{-}Florina Balcan and
               Hsuan{-}Tien Lin},
  title     = {Parameterized Explainer for Graph Neural Network},
  booktitle = {Advances in Neural Information Processing Systems 33: Annual Conference
               on Neural Information Processing Systems 2020, NeurIPS 2020, December
               6-12, 2020, virtual},
  year      = {2020},
  url       = {https://proceedings.neurips.cc/paper/2020/hash/e37b08dd3015330dcbb5d6663667b8b8-Abstract.html},
  bibsource = {dblp computer science bibliography, https://dblp.org}
}

@inproceedings{sundararajan2017axiomaticattributiondeepnetworks,
  author    = {Mukund Sundararajan and
               Ankur Taly and
               Qiqi Yan},
  editor    = {Doina Precup and
               Yee Whye Teh},
  title     = {Axiomatic Attribution for Deep Networks},
  booktitle = {Proceedings of the 34th International Conference on Machine Learning,
               {ICML} 2017, Sydney, NSW, Australia, 6-11 August 2017},
  series    = {Proceedings of Machine Learning Research},
  volume    = {70},
  pages     = {3319--3328},
  publisher = {{PMLR}},
  year      = {2017},
  url       = {http://proceedings.mlr.press/v70/sundararajan17a.html},
  bibsource = {dblp computer science bibliography, https://dblp.org}
}

@inproceedings{simonyan2014deepinsideconvolutionalnetworks,
  author       = {Karen Simonyan and
                  Andrea Vedaldi and
                  Andrew Zisserman},
  editor       = {Yoshua Bengio and
                  Yann LeCun},
  title        = {Deep Inside Convolutional Networks: Visualising Image Classification
                  Models and Saliency Maps},
  booktitle    = {2nd International Conference on Learning Representations, {ICLR} 2014,
                  Banff, AB, Canada, April 14-16, 2014, Workshop Track Proceedings},
  year         = {2014},
  url          = {https://doi.org/10.48550/arXiv.1312.6034},
  doi          = {10.48550/ARXIV.1312.6034},
  bibsource    = {dblp computer science bibliography, https://dblp.org}
}

@article{shrikumar2017justblackboxlearning,
  title={Not Just a Black Box: Learning Important Features Through Propagating Activation Differences},
  author={Avanti Shrikumar and Peyton Greenside and Anna Shcherbina and Anshul B. Kundaje},
  journal={ArXiv},
  year={2016},
  volume={abs/1605.01713},
  url={https://api.semanticscholar.org/CorpusID:8564234}
}

@inproceedings{10.1007/978-3-319-46466-4_8,
 address = {Cham},
 author = {Mahendran, Aravindh
and Vedaldi, Andrea},
 booktitle = {Computer Vision -- ECCV 2016},
 editor = {Leibe, Bastian
and Matas, Jiri
and Sebe, Nicu
and Welling, Max},
 pages = {120--135},
 publisher = {Springer International Publishing},
 title = {Salient Deconvolutional Networks},
 year = {2016}
}

@article{springenberg2015strivingsimplicityconvolutionalnet,
  title={Striving for Simplicity: The All Convolutional Net},
  author={Jost Tobias Springenberg and Alexey Dosovitskiy and Thomas Brox and Martin A. Riedmiller},
  journal={CoRR},
  year={2014},
  volume={abs/1412.6806},
  url={https://api.semanticscholar.org/CorpusID:12998557}
}

@inproceedings{schlichtkrull2022interpretinggraphneuralnetworks,
  author    = {Michael Sejr Schlichtkrull and
               Nicola De Cao and
               Ivan Titov},
  title     = {Interpreting Graph Neural Networks for {NLP} With Differentiable Edge
               Masking},
  booktitle = {9th International Conference on Learning Representations, {ICLR} 2021,
               Virtual Event, Austria, May 3-7, 2021},
  publisher = {OpenReview.net},
  year      = {2021},
  url       = {https://openreview.net/forum?id=WznmQa42ZAx},
  bibsource = {dblp computer science bibliography, https://dblp.org}
}

@article{trumbelj2010AnEE,
 author = {Erik {\v{S}}trumbelj and Igor Kononenko},
 journal = {J. Mach. Learn. Res.},
 pages = {1-18},
 title = {An Efficient Explanation of Individual Classifications using Game Theory},
 url = {https://api.semanticscholar.org/CorpusID:14451872},
 volume = {11},
 year = {2010}
}

@inproceedings{kipf2017semisupervisedclassificationgraphconvolutional,
  author    = {Thomas N. Kipf and
               Max Welling},
  title     = {Semi-Supervised Classification with Graph Convolutional Networks},
  booktitle = {5th International Conference on Learning Representations, {ICLR} 2017,
               Toulon, France, April 24-26, 2017, Conference Track Proceedings},
  publisher = {OpenReview.net},
  year      = {2017},
  url       = {https://openreview.net/forum?id=SJU4ayYgl},
  bibsource = {dblp computer science bibliography, https://dblp.org}
}

@inproceedings{velickovic2018graph,
  author    = {Petar Velickovic and
               Guillem Cucurull and
               Arantxa Casanova and
               Adriana Romero and
               Pietro Li{\`{o}} and
               Yoshua Bengio},
  title     = {Graph Attention Networks},
  booktitle = {6th International Conference on Learning Representations, {ICLR} 2018,
               Vancouver, BC, Canada, April 30 - May 3, 2018, Conference Track Proceedings},
  publisher = {OpenReview.net},
  year      = {2018},
  url       = {https://openreview.net/forum?id=rJXMpikCZ},
  bibsource = {dblp computer science bibliography, https://dblp.org}
}

@inproceedings{xu2019powerfulgraphneuralnetworks,
  author    = {Keyulu Xu and
               Weihua Hu and
               Jure Leskovec and
               Stefanie Jegelka},
  title     = {How Powerful are Graph Neural Networks?},
  booktitle = {7th International Conference on Learning Representations, {ICLR} 2019,
               New Orleans, LA, USA, May 6-9, 2019},
  publisher = {OpenReview.net},
  year      = {2019},
  url       = {https://openreview.net/forum?id=ryGs6iA5Km},
  bibsource = {dblp computer science bibliography, https://dblp.org}
}

@inproceedings{zhang2021protgnnselfexplaininggraphneural,
  author    = {Zaixi Zhang and
               Qi Liu and
               Hao Wang and
               Chengqiang Lu and
               Cheekong Lee},
  title     = {ProtGNN: Towards Self-Explaining Graph Neural Networks},
  booktitle = {Thirty-Sixth {AAAI} Conference on Artificial Intelligence, {AAAI}
               2022, Thirty-Fourth Conference on Innovative Applications of Artificial
               Intelligence, {IAAI} 2022, The Twelveth Symposium on Educational Advances
               in Artificial Intelligence, {EAAI} 2022 Virtual Event, February 22
               - March 1, 2022},
  pages     = {9127--9135},
  publisher = {{AAAI} Press},
  year      = {2022},
  url       = {https://ojs.aaai.org/index.php/AAAI/article/view/20898},
  bibsource = {dblp computer science bibliography, https://dblp.org}
}

@inproceedings{amara2024graphframexsystematicevaluationexplainability,
  title={{GraphFramEx}: Towards Systematic Evaluation of Explainability Methods for Graph Neural Networks},
  author={Kenza Amara and Rex Ying and Zi-Tao Zhang and Zhi-Hao Han and Yinan Shan and Ulrik Brandes and Sebastian Schemm and Ce Zhang},
  booktitle={Proceedings of the First Learning on Graphs Conference (LoG)},
  year={2022},
  url={https://api.semanticscholar.org/CorpusID:249889598}
}

@inproceedings{zheng2024robustfidelityevaluatingexplainability,
  author       = {Xu Zheng and
                  Farhad Shirani and
                  Tianchun Wang and
                  Wei Cheng and
                  Zhuomin Chen and
                  Haifeng Chen and
                  Hua Wei and
                  Dongsheng Luo},
  title        = {Towards Robust Fidelity for Evaluating Explainability of Graph Neural
                  Networks},
  booktitle    = {The Twelfth International Conference on Learning Representations,
                  {ICLR} 2024, Vienna, Austria, May 7-11, 2024},
  publisher    = {OpenReview.net},
  year         = {2024},
  url          = {https://openreview.net/forum?id=up6hr4hIQH},
  bibsource    = {dblp computer science bibliography, https://dblp.org}
}

@inproceedings{bajaj2022robustcounterfactualexplanationsgraph,
  author    = {Mohit Bajaj and
               Lingyang Chu and
               Zi Yu Xue and
               Jian Pei and
               Lanjun Wang and
               Peter Cho{-}Ho Lam and
               Yong Zhang},
  editor    = {Marc'Aurelio Ranzato and
               Alina Beygelzimer and
               Yann N. Dauphin and
               Percy Liang and
               Jennifer Wortman Vaughan},
  title     = {Robust Counterfactual Explanations on Graph Neural Networks},
  booktitle = {Advances in Neural Information Processing Systems 34: Annual Conference
               on Neural Information Processing Systems 2021, NeurIPS 2021, December
               6-14, 2021, virtual},
  pages     = {5644--5655},
  year      = {2021},
  url       = {https://proceedings.neurips.cc/paper/2021/hash/2c8c3a57383c63caef6724343eb62257-Abstract.html},
  bibsource = {dblp computer science bibliography, https://dblp.org}
}

@article{chen2022greasegeneratefactualcounterfactual,
  title={{GREASE}: Generate Factual and Counterfactual Explanations for +{GNN}-based Recommendations},
  author={Ziheng Chen and Fabrizio Silvestri and Jia Wang and Yong-Feng Zhang and Zhen-Hua Huang and Hongshik Ahn and Gabriele Tolomei},
  journal={ArXiv},
  year={2022},
  volume={abs/2208.04222},
  url={https://api.semanticscholar.org/CorpusID:251402395}
}

@inproceedings{yuan2021explainabilitygraphneuralnetworks,
  author    = {Hao Yuan and
               Haiyang Yu and
               Jie Wang and
               Kang Li and
               Shuiwang Ji},
  editor    = {Marina Meila and
               Tong Zhang},
  title     = {On Explainability of Graph Neural Networks via Subgraph Explorations},
  booktitle = {Proceedings of the 38th International Conference on Machine Learning,
               {ICML} 2021, 18-24 July 2021, Virtual Event},
  series    = {Proceedings of Machine Learning Research},
  volume    = {139},
  pages     = {12241--12252},
  publisher = {{PMLR}},
  year      = {2021},
  url       = {http://proceedings.mlr.press/v139/yuan21c.html},
  bibsource = {dblp computer science bibliography, https://dblp.org}
}

@inproceedings{dai2021selfexplainablegraphneuralnetwork,
author = {Dai, Enyan and Wang, Suhang},
title = {Towards Self-Explainable Graph Neural Network},
year = {2021},
isbn = {9781450384469},
publisher = {Association for Computing Machinery},
address = {New York, NY, USA},
url = {https://doi.org/10.1145/3459637.3482306},
doi = {10.1145/3459637.3482306},
booktitle = {Proceedings of the 30th ACM International Conference on Information \& Knowledge Management},
pages = {302–311},
numpages = {10},
location = {Virtual Event, Queensland, Australia},
series = {CIKM '21}
}

@inproceedings{lucic2022cfgnnexplainercounterfactualexplanationsgraph,
  author    = {Ana Lucic and
               Maartje A. ter Hoeve and
               Gabriele Tolomei and
               Maarten de Rijke and
               Fabrizio Silvestri},
  editor    = {Gustau Camps{-}Valls and
               Francisco J. R. Ruiz and
               Isabel Valera},
  title     = {CF-GNNExplainer: Counterfactual Explanations for Graph Neural Networks},
  booktitle = {International Conference on Artificial Intelligence and Statistics,
               {AISTATS} 2022, 28-30 March 2022, Virtual Event},
  series    = {Proceedings of Machine Learning Research},
  volume    = {151},
  pages     = {4499--4511},
  publisher = {{PMLR}},
  year      = {2022},
  url       = {https://proceedings.mlr.press/v151/lucic22a.html},
  bibsource = {dblp computer science bibliography, https://dblp.org}
}

@inproceedings{wu2020graphinformationbottleneck,
  author    = {Tailin Wu and
               Hongyu Ren and
               Pan Li and
               Jure Leskovec},
  editor    = {Hugo Larochelle and
               Marc'Aurelio Ranzato and
               Raia Hadsell and
               Maria{-}Florina Balcan and
               Hsuan{-}Tien Lin},
  title     = {Graph Information Bottleneck},
  booktitle = {Advances in Neural Information Processing Systems 33: Annual Conference
               on Neural Information Processing Systems 2020, NeurIPS 2020, December
               6-12, 2020, virtual},
  year      = {2020},
  url       = {https://proceedings.neurips.cc/paper/2020/hash/ebc2aa04e75e3caabda543a1317160c0-Abstract.html},
  bibsource = {dblp computer science bibliography, https://dblp.org}
}

@inproceedings{miao2022interpretablegeneralizablegraphlearning,
  author    = {Siqi Miao and
               Mia Liu and
               Pan Li},
  editor    = {Kamalika Chaudhuri and
               Stefanie Jegelka and
               Le Song and
               Csaba Szepesv{\'{a}}ri and
               Gang Niu and
               Sivan Sabato},
  title     = {Interpretable and Generalizable Graph Learning via Stochastic Attention
               Mechanism},
  booktitle = {International Conference on Machine Learning, {ICML} 2022, 17-23 July
               2022, Baltimore, Maryland, {USA}},
  series    = {Proceedings of Machine Learning Research},
  volume    = {162},
  pages     = {15524--15543},
  publisher = {{PMLR}},
  year      = {2022},
  url       = {https://proceedings.mlr.press/v162/miao22a.html},
  bibsource = {dblp computer science bibliography, https://dblp.org}
}

@inproceedings{feng2022kergnnsinterpretablegraphneural,
  author    = {Aosong Feng and
               Chenyu You and
               Shiqiang Wang and
               Leandros Tassiulas},
  title     = {KerGNNs: Interpretable Graph Neural Networks with Graph Kernels},
  booktitle = {Thirty-Sixth {AAAI} Conference on Artificial Intelligence, {AAAI}
               2022, Thirty-Fourth Conference on Innovative Applications of Artificial
               Intelligence, {IAAI} 2022, The Twelveth Symposium on Educational Advances
               in Artificial Intelligence, {EAAI} 2022 Virtual Event, February 22
               - March 1, 2022},
  pages     = {6614--6622},
  publisher = {{AAAI} Press},
  year      = {2022},
  url       = {https://ojs.aaai.org/index.php/AAAI/article/view/20615},
  bibsource = {dblp computer science bibliography, https://dblp.org}
}

@inproceedings{agarwal2021unifiedframeworkfairstable,
  author    = {Chirag Agarwal and
               Himabindu Lakkaraju and
               Marinka Zitnik},
  editor    = {Cassio P. de Campos and
               Marloes H. Maathuis and
               Erik Quaeghebeur},
  title     = {Towards a unified framework for fair and stable graph representation
               learning},
  booktitle = {Proceedings of the Thirty-Seventh Conference on Uncertainty in Artificial
               Intelligence, {UAI} 2021, Virtual Event, 27-30 July 2021},
  series    = {Proceedings of Machine Learning Research},
  volume    = {161},
  pages     = {2114--2124},
  publisher = {{AUAI} Press},
  year      = {2021},
  url       = {https://proceedings.mlr.press/v161/agarwal21b.html},
  bibsource = {dblp computer science bibliography, https://dblp.org}
}

@inproceedings{wu2022discoveringinvariantrationalesgraph,
  author    = {Yingxin Wu and
               Xiang Wang and
               An Zhang and
               Xiangnan He and
               Tat{-}Seng Chua},
  title     = {Discovering Invariant Rationales for Graph Neural Networks},
  booktitle = {The Tenth International Conference on Learning Representations, {ICLR}
               2022, Virtual Event, April 25-29, 2022},
  publisher = {OpenReview.net},
  year      = {2022},
  url       = {https://openreview.net/forum?id=hGXij5rfiHw},
  bibsource = {dblp computer science bibliography, https://dblp.org}
}

@article{agarwal2023evaluating,
 author = {Agarwal, Chirag and Queen, Owen and Lakkaraju, Himabindu and Zitnik, Marinka},
 journal = {Scientific Data},
 number = {144},
 publisher = {Nature Publishing Group},
 title = {Evaluating Explainability for Graph Neural Networks},
 url = {https://www.nature.com/articles/s41597-023-01974-x},
 volume = {10},
 year = {2023}
}

@article{doi:10.1021/jm00106a046,
 author = {Debnath, Asim Kumar and Lopez de Compadre, Rosa L. and Debnath, Gargi and Shusterman, Alan J. and Hansch, Corwin},
 doi = {10.1021/jm00106a046},
 eprint = { 
https://doi.org/10.1021/jm00106a046},
 journal = {Journal of Medicinal Chemistry},
 number = {2},
 pages = {786-797},
 title = {Structure-activity relationship of mutagenic aromatic and heteroaromatic nitro compounds. Correlation with molecular orbital energies and hydrophobicity},
 url = { 
https://doi.org/10.1021/jm00106a046},
 volume = {34},
 year = {1991}
}

@inproceedings{NEURIPS2020_417fbbf2,
  author    = {Benjamin Sanchez{-}Lengeling and
               Jennifer N. Wei and
               Brian K. Lee and
               Emily Reif and
               Peter Wang and
               Wesley Wei Qian and
               Kevin McCloskey and
               Lucy J. Colwell and
               Alexander B. Wiltschko},
  editor    = {Hugo Larochelle and
               Marc'Aurelio Ranzato and
               Raia Hadsell and
               Maria{-}Florina Balcan and
               Hsuan{-}Tien Lin},
  title     = {Evaluating Attribution for Graph Neural Networks},
  booktitle = {Advances in Neural Information Processing Systems 33: Annual Conference
               on Neural Information Processing Systems 2020, NeurIPS 2020, December
               6-12, 2020, virtual},
  year      = {2020},
  url       = {https://proceedings.neurips.cc/paper/2020/hash/417fbbf2e9d5a28a855a11894b2e795a-Abstract.html},
  bibsource = {dblp computer science bibliography, https://dblp.org}
}

@article{doi:10.1073/pnas.1820657116,
 author = {Kevin McCloskey  and Ankur Taly  and Federico Monti  and Michael P. Brenner  and Lucy J. Colwell },
 doi = {10.1073/pnas.1820657116},
 eprint = {https://www.pnas.org/doi/pdf/10.1073/pnas.1820657116},
 journal = {Proceedings of the National Academy of Sciences},
 number = {24},
 pages = {11624-11629},
 title = {Using attribution to decode binding mechanism in neural network models for chemistry},
 url = {https://www.pnas.org/doi/abs/10.1073/pnas.1820657116},
 volume = {116},
 year = {2019}
}

@inproceedings{4053093,
 author = {Wale, Nikil and Karypis, George},
 booktitle = {Sixth International Conference on Data Mining (ICDM'06)},
 doi = {10.1109/ICDM.2006.39},
 number = {},
 pages = {678-689},
 title = {Comparison of Descriptor Spaces for Chemical Compound Retrieval and Classification},
 volume = {},
 year = {2006}
}

@article{doi:10.1021/ci300124c,
 author = {Martins, Ines Filipa and Teixeira, Ana L. and Pinheiro, Luis and Falcao, Andre O.},
 doi = {10.1021/ci300124c},
 eprint = { 
https://doi.org/10.1021/ci300124c},
 journal = {Journal of Chemical Information and Modeling},
 number = {6},
 pages = {1686-1697},
 title = {A Bayesian Approach to in Silico Blood-Brain Barrier Penetration Modeling},
 url = { 
https://doi.org/10.1021/ci300124c},
 volume = {52},
 year = {2012}
}

@article{10.3389/fenvs.2015.00080,
 author = {Mayr, Andreas  and Klambauer, Günter  and Unterthiner, Thomas  and Hochreiter, Sepp },
 doi = {10.3389/fenvs.2015.00080},
 issn = {2296-665X},
 journal = {Frontiers in Environmental Science},
 title = {{DeepTox}: Toxicity Prediction using Deep Learning},
 url = {https://www.frontiersin.org/journals/environmental-science/articles/10.3389/fenvs.2015.00080},
 volume = {3},
 year = {2015}
}

@article{10.1093/bioinformatics/bti1007,
 address = {USA},
 author = {Borgwardt, Karsten M. and Ong, Cheng Soon and Sch\"{o}nauer, Stefan and Vishwanathan, S. V. N. and Smola, Alex J. and Kriegel, Hans-Peter},
 doi = {10.1093/bioinformatics/bti1007},
 issn = {1367-4803},
 issue_date = {January 2005},
 journal = {Bioinformatics},
 number = {1},
 numpages = {10},
 pages = {47–56},
 publisher = {Oxford University Press, Inc.},
 title = {Protein function prediction via graph kernels},
 url = {https://doi.org/10.1093/bioinformatics/bti1007},
 volume = {21},
 year = {2005}
}

@inproceedings{knyazev2019understandingattentiongeneralizationgraph,
  author    = {Boris Knyazev and
               Graham W. Taylor and
               Mohamed R. Amer},
  editor    = {Hanna M. Wallach and
               Hugo Larochelle and
               Alina Beygelzimer and
               Florence d'Alch{\'{e}}{-}Buc and
               Emily B. Fox and
               Roman Garnett},
  title     = {Understanding Attention and Generalization in Graph Neural Networks},
  booktitle = {Advances in Neural Information Processing Systems 32: Annual Conference
               on Neural Information Processing Systems 2019, NeurIPS 2019, December
               8-14, 2019, Vancouver, BC, Canada},
  pages     = {4204--4214},
  year      = {2019},
  url       = {https://proceedings.neurips.cc/paper/2019/hash/4c5bcfec8584af0d967f1ab10179ca4b-Abstract.html},
  bibsource = {dblp computer science bibliography, https://dblp.org}
}

@article{yuan2022explainabilitygraphneuralnetworks,
  author={Yuan, Hao and Yu, Haiyang and Gui, Shurui and Ji, Shuiwang},
  title={Explainability in Graph Neural Networks: A Taxonomic Survey},
  journal={IEEE Transactions on Pattern Analysis and Machine Intelligence},
  year={2023},
  volume={45},
  number={5},
  month={May},
  pages={5782--5799},
  publisher={IEEE},
  doi={10.1109/TPAMI.2022.3204236},
  url={https://doi.org/10.1109/TPAMI.2022.3204236}
}

@article{yu2022improvingsubgraphrecognitionvariational,
  title={Improving Subgraph Recognition with Variational Graph Information Bottleneck},
  author={Junchi Yu and Jie Cao and Ran He},
  journal={2022 IEEE/CVF Conference on Computer Vision and Pattern Recognition (CVPR)},
  year={2021},
  pages={19374-19383},
  url={https://api.semanticscholar.org/CorpusID:245335028}
}

@inproceedings{azzolin2023globalexplainabilitygnnslogic,
  author       = {Steve Azzolin and
                  Antonio Longa and
                  Pietro Barbiero and
                  Pietro Li{\`{o}} and
                  Andrea Passerini},
  title        = {Global Explainability of GNNs via Logic Combination of Learned Concepts},
  booktitle    = {The Eleventh International Conference on Learning Representations,
                  {ICLR} 2023, Kigali, Rwanda, May 1-5, 2023},
  publisher    = {OpenReview.net},
  year         = {2023},
  url          = {https://openreview.net/pdf?id=OTbRTIY4YS},
  bibsource    = {dblp computer science bibliography, https://dblp.org}
}

@inproceedings{zhang2020relexmodelagnosticrelationalmodel,
author = {Zhang, Yue and Defazio, David and Ramesh, Arti},
title = {{RelEx}: A Model-Agnostic Relational Model Explainer},
year = {2021},
isbn = {9781450384735},
publisher = {Association for Computing Machinery},
address = {New York, NY, USA},
url = {https://doi.org/10.1145/3461702.3462562},
doi = {10.1145/3461702.3462562},
booktitle = {Proceedings of the 2021 AAAI/ACM Conference on AI, Ethics, and Society},
pages = {1042–1049},
numpages = {8},
location = {Virtual Event, USA},
series = {AIES '21}
}

@inproceedings{rymarczyk2023progrest,
 author = {Rymarczyk, Dawid and Dobrowolski, Daniel and Danel, Tomasz},
 booktitle = {Proceedings of the 2023 SIAM International Conference on Data Mining (SDM)},
 organization = {SIAM},
 pages = {379--387},
 title = {Progrest: Prototypical graph regression soft trees for molecular property prediction},
 year = {2023}
}

@article{wieder2020compact,
 author = {Wieder, Oliver and Kohlbacher, Stefan and Kuenemann, M{\'e}laine and Garon, Arthur and Ducrot, Pierre and Seidel, Thomas and Langer, Thierry},
 journal = {Drug Discovery Today: Technologies},
 pages = {1--12},
 publisher = {Elsevier},
 title = {A compact review of molecular property prediction with graph neural networks},
 volume = {37},
 year = {2020}
}

@article{jimenez2020drug,
 author = {Jim{\'e}nez-Luna, Jos{\'e} and Grisoni, Francesca and Schneider, Gisbert},
 journal = {Nature Machine Intelligence},
 number = {10},
 pages = {573--584},
 publisher = {Nature Publishing Group UK London},
 title = {Drug discovery with explainable artificial intelligence},
 volume = {2},
 year = {2020}
}

@article{wu2023black,
 author = {Wu, Zhenxing and Chen, Jihong and Li, Yitong and Deng, Yafeng and Zhao, Haitao and Hsieh, Chang-Yu and Hou, Tingjun},
 journal = {Journal of Chemical Information and Modeling},
 number = {24},
 pages = {7617--7627},
 publisher = {ACS Publications},
 title = {From black boxes to actionable insights: a perspective on explainable artificial intelligence for scientific discovery},
 volume = {63},
 year = {2023}
}

@inproceedings{faber2021comparing,
  author       = {Lukas Faber and
                  Amin K. Moghaddam and
                  Roger Wattenhofer},
  editor       = {Feida Zhu and
                  Beng Chin Ooi and
                  Chunyan Miao},
  title        = {When Comparing to Ground Truth is Wrong: On Evaluating {GNN} Explanation
                  Methods},
  booktitle    = {{KDD} '21: The 27th {ACM} {SIGKDD} Conference on Knowledge Discovery
                  and Data Mining, Virtual Event, Singapore, August 14-18, 2021},
  pages        = {332--341},
  publisher    = {{ACM}},
  year         = {2021},
  url          = {https://doi.org/10.1145/3447548.3467283},
  doi          = {10.1145/3447548.3467283},
  bibsource    = {dblp computer science bibliography, https://dblp.org}
}

@article{baell2010new,
 author = {Baell, Jonathan B and Holloway, Georgina A},
 journal = {Journal of medicinal chemistry},
 number = {7},
 pages = {2719--2740},
 publisher = {ACS Publications},
 title = {New substructure filters for removal of pan assay interference compounds (PAINS) from screening libraries and for their exclusion in bioassays},
 volume = {53},
 year = {2010}
}

@article{gaulton2012chembl,
 author = {Gaulton, Anna and Bellis, Louisa J and Bento, A Patricia and Chambers, Jon and Davies, Mark and Hersey, Anne and Light, Yvonne and McGlinchey, Shaun and Michalovich, David and Al-Lazikani, Bissan and others},
 journal = {Nucleic acids research},
 number = {D1},
 pages = {D1100--D1107},
 publisher = {Oxford University Press},
 title = {ChEMBL: a large-scale bioactivity database for drug discovery},
 volume = {40},
 year = {2012}
}

@inproceedings{elhadri2025looks,
  title={This looks like what? Challenges and Future Research Directions for Part-Prototype Models},
  author={Elhadri, Khawla and Michalski, Tomasz and Wr{\'o}bel, Adam and Schl{\"o}tterer, J{\"o}rg and Zieli{\'n}ski, Bartosz and Seifert, Christin},
  booktitle={Proceedings of the 4th World Conference on eXplainable Artificial Intelligence (XAI)},
  year={2026},
  eprint={2502.09340},
  archivePrefix={arXiv},
  url={https://arxiv.org/abs/2502.09340}
}

@inproceedings{Tan_2022,
  author       = {Juntao Tan and
                  Shijie Geng and
                  Zuohui Fu and
                  Yingqiang Ge and
                  Shuyuan Xu and
                  Yunqi Li and
                  Yongfeng Zhang},
  editor       = {Fr{\'{e}}d{\'{e}}rique Laforest and
                  Rapha{\"{e}}l Troncy and
                  Elena Simperl and
                  Deepak Agarwal and
                  Aristides Gionis and
                  Ivan Herman and
                  Lionel M{\'{e}}dini},
  title        = {Learning and Evaluating Graph Neural Network Explanations based on
                  Counterfactual and Factual Reasoning},
  booktitle    = {{WWW} '22: The {ACM} Web Conference 2022, Virtual Event, Lyon, France,
                  April 25 - 29, 2022},
  pages        = {1018--1027},
  publisher    = {{ACM}},
  year         = {2022},
  url          = {https://doi.org/10.1145/3485447.3511948},
  doi          = {10.1145/3485447.3511948},
  bibsource    = {dblp computer science bibliography, https://dblp.org}
}

@article{Cosmo_2025,
   title={Graph Kernel Neural Networks},
   volume={36},
   ISSN={2162-2388},
   url={http://dx.doi.org/10.1109/TNNLS.2024.3400850},
   DOI={10.1109/tnnls.2024.3400850},
   number={4},
   journal={IEEE Transactions on Neural Networks and Learning Systems},
   publisher={Institute of Electrical and Electronics Engineers (IEEE)},
   author={Cosmo, Luca and Minello, Giorgia and Bicciato, Alessandro and Bronstein, Michael M. and Rodolà, Emanuele and Rossi, Luca and Torsello, Andrea},
   year={2025},
   month=apr, pages={6257–6270} }

@article{Longa_2025,
   title={Explaining the Explainers in Graph Neural Networks: a Comparative Study},
   volume={57},
   ISSN={1557-7341},
   url={http://dx.doi.org/10.1145/3696444},
   DOI={10.1145/3696444},
   number={5},
   journal={ACM Computing Surveys},
   publisher={Association for Computing Machinery (ACM)},
   author={Longa, Antonio and Azzolin, Steve and Santin, Gabriele and Cencetti, Giulia and Lio, Pietro and Lepri, Bruno and Passerini, Andrea},
   year={2025},
   month=jan, pages={1–37} }

@article{Nauta_2023,
   title={From Anecdotal Evidence to Quantitative Evaluation Methods: A Systematic Review on Evaluating Explainable AI},
   volume={55},
   ISSN={1557-7341},
   url={http://dx.doi.org/10.1145/3583558},
   DOI={10.1145/3583558},
   number={13s},
   journal={ACM Computing Surveys},
   publisher={Association for Computing Machinery (ACM)},
   author={Nauta, Meike and Trienes, Jan and Pathak, Shreyasi and Nguyen, Elisa and Peters, Michelle and Schmitt, Yasmin and Schlötterer, Jörg and van Keulen, Maurice and Seifert, Christin},
   year={2023},
   month=jul, pages={1–42} }

@article{gui2023flowxexplainablegraphneural,
  author={Gui, Shurui and Yuan, Hao and Wang, Jie and Lao, Qicheng and Li, Kang and Ji, Shuiwang},
  title={{FlowX}: Towards Explainable Graph Neural Networks via Message Flows},
  journal={IEEE Transactions on Pattern Analysis and Machine Intelligence},
  year={2024},
  volume={46},
  number={7},
  month={Jul.},
  pages={4567--4578},
  doi={10.1109/TPAMI.2023.3347470},
  url={https://doi.org/10.1109/TPAMI.2023.3347470}
}

@inproceedings{vu2020pgmexplainerprobabilisticgraphicalmodel,
  author    = {Minh N. Vu and
               My T. Thai},
  editor    = {Hugo Larochelle and
               Marc'Aurelio Ranzato and
               Raia Hadsell and
               Maria{-}Florina Balcan and
               Hsuan{-}Tien Lin},
  title     = {PGM-Explainer: Probabilistic Graphical Model Explanations for Graph
               Neural Networks},
  booktitle = {Advances in Neural Information Processing Systems 33: Annual Conference
               on Neural Information Processing Systems 2020, NeurIPS 2020, December
               6-12, 2020, virtual},
  year      = {2020},
  url       = {https://proceedings.neurips.cc/paper/2020/hash/8fb134f258b1f7865a6ab2d935a897c9-Abstract.html},
  bibsource = {dblp computer science bibliography, https://dblp.org},
  archivePrefix={arXiv},
  eprint={2010.05788},
}

@inproceedings{hesse2023funnybirds,
  author       = {Robin Hesse and
                  Simone Schaub{-}Meyer and
                  Stefan Roth},
  title        = {FunnyBirds: {A} Synthetic Vision Dataset for a Part-Based Analysis
                  of Explainable {AI} Methods},
  booktitle    = {{IEEE/CVF} International Conference on Computer Vision, {ICCV} 2023,
                  Paris, France, October 1-6, 2023},
  pages        = {3958--3968},
  publisher    = {{IEEE}},
  year         = {2023},
  url          = {https://doi.org/10.1109/ICCV51070.2023.00368},
  doi          = {10.1109/ICCV51070.2023.00368},
  bibsource    = {dblp computer science bibliography, https://dblp.org}
}

@inproceedings{muschalik2025exact,
  author       = {Maximilian Muschalik and
                  Fabian Fumagalli and
                  Paolo Frazzetto and
                  Janine Strotherm and
                  Luca Hermes and
                  Alessandro Sperduti and
                  Eyke H{\"{u}}llermeier and
                  Barbara Hammer},
  title        = {Exact Computation of Any-Order Shapley Interactions for Graph Neural
                  Networks},
  booktitle    = {The Thirteenth International Conference on Learning Representations,
                  {ICLR} 2025, Singapore, April 24-28, 2025},
  publisher    = {OpenReview.net},
  year         = {2025},
  url          = {https://openreview.net/forum?id=9tKC0YM8sX},
  bibsource    = {dblp computer science bibliography, https://dblp.org}
}

@article{mastropietro2022edgeshaper,
  title={{EdgeSHAPer}: Bond-centric Shapley value-based explanation method for graph neural networks},
  author={Mastropietro, Andrea and Pasculli, Giuseppe and Feldmann, Christian and Rodr{\'\i}guez-P{\'e}rez, Raquel and Bajorath, J{\"u}rgen},
  journal={iScience},
  volume={25},
  number={10},
  year={2022},
  publisher={Elsevier}
}

@article{proszewska2026aiming,
  author       = {Magdalena Proszewska and
                  N. Siddharth},
  title        = {AIMing for Standardised Explainability Evaluation in GNNs: {A} Framework
                  and Case Study on Graph Kernel Networks},
  journal      = {Trans. Mach. Learn. Res.},
  volume       = {2026},
  year         = {2026},
  url          = {https://openreview.net/forum?id=onZkYXI7oe},
  bibsource    = {dblp computer science bibliography, https://dblp.org}
}

@inproceedings{jacovi-goldberg-2020-towards,
    title = "Towards Faithfully Interpretable {NLP} Systems: How Should We Define and Evaluate Faithfulness?",
    author = "Jacovi, Alon  and
      Goldberg, Yoav",
    editor = "Jurafsky, Dan  and
      Chai, Joyce  and
      Schluter, Natalie  and
      Tetreault, Joel",
    booktitle = "Proceedings of the 58th Annual Meeting of the Association for Computational Linguistics",
    month = jul,
    year = "2020",
    address = "Online",
    publisher = "Association for Computational Linguistics",
    url = "https://aclanthology.org/2020.acl-main.386/",
    doi = "10.18653/v1/2020.acl-main.386",
    pages = "4198--4205"
}

@inproceedings{chen2020graphneuralnetworkscount,
  author    = {Zhengdao Chen and
               Lei Chen and
               Soledad Villar and
               Joan Bruna},
  editor    = {Hugo Larochelle and
               Marc'Aurelio Ranzato and
               Raia Hadsell and
               Maria{-}Florina Balcan and
               Hsuan{-}Tien Lin},
  title     = {Can Graph Neural Networks Count Substructures?},
  booktitle = {Advances in Neural Information Processing Systems 33: Annual Conference
               on Neural Information Processing Systems 2020, NeurIPS 2020, December
               6-12, 2020, virtual},
  year      = {2020},
  url       = {https://proceedings.neurips.cc/paper/2020/hash/75877cb75154206c4e65e76b88a12712-Abstract.html},
  bibsource = {dblp computer science bibliography, https://dblp.org},
  archivePrefix={arXiv},
  eprint={2002.04025},
}

@inproceedings{pereira2024distillnexplainexplaining,
  author       = {Tamara A. Pereira and
                  Erik Nascimento and
                  Lucas E. Resck and
                  Diego Mesquita and
                  Amauri H. Souza},
  editor       = {Francisco J. R. Ruiz and
                  Jennifer G. Dy and
                  Jan{-}Willem van de Meent},
  title        = {Distill n' Explain: explaining graph neural networks using simple
                  surrogates},
  booktitle    = {International Conference on Artificial Intelligence and Statistics,
                  25-27 April 2023, Palau de Congressos, Valencia, Spain},
  series       = {Proceedings of Machine Learning Research},
  volume       = {206},
  pages        = {6199--6214},
  publisher    = {{PMLR}},
  year         = {2023},
  url          = {https://proceedings.mlr.press/v206/pereira23a.html},
  bibsource    = {dblp computer science bibliography, https://dblp.org},
  archivePrefix={arXiv},
  eprint={2303.10139},
}

@article{huang2020graphlimelocalinterpretablemodel,
  author={Huang, Qiang and Yamada, Makoto and Tian, Yuan and Singh, Dinesh and Chang, Yi},
  title={{GraphLIME}: Local Interpretable Model Explanations for Graph Neural Networks},
  journal={IEEE Transactions on Knowledge and Data Engineering},
  year={2023},
  volume={35},
  number={7},
  month={Jul.},
  pages={6968--6972},
  publisher={IEEE},
  doi={10.1109/TKDE.2022.3187455},
  url={https://doi.org/10.1109/TKDE.2022.3187455}
}

@book{Tukey1977,
  author    = {Tukey, J. W.},
  title     = {Exploratory Data Analysis},
  year      = {1977},
  publisher = {Addison-Wesley},
  address   = {Reading, MA}
}

\newpage

\appendix

\section{Appendix}

\subsection{Reproducibility} 
\label{app:reproducibility}
To ensure accessibility and encourage community engagement, we have hosted the B-XAIC dataset on Hugging Face (\url{https://huggingface.co/datasets/mproszewska/B-XAIC}) and provided open-source code (\url{https://github.com/mproszewska/B-XAIC}) for its execution. Furthermore, the careful design of our molecule selection process and data hosting infrastructure allows us to effectively mitigate the risk of data misuse. Importantly, the dataset is released under the CC-BY-SA license, empowering the community to leverage this resource while ensuring proper attribution and continued sharing.

\paragraph{Resources}
We conduct all experiments on an NVIDIA H100 GPU with 80 GB of HBM3 memory. Training each model (GCN, GAT, GIN, and ProtGNN variants) takes less than one hour. The runtime of explainers varies substantially depending on the method, the availability of efficient implementations, and used hyperparameters.

\subsection{Ethics Statement: Impact and Broader Impact} This work contributes to the broader field of XAI, specifically within the context of GNNs applied to drug discovery and molecular modeling. The B-XAIC dataset offers the community a standardized benchmark for evaluating novel XAI techniques dedicated to small molecules. Beyond this specific domain, we anticipate its utility for assessing XAI methods on graphs of moderate size (up to 60 nodes), a common scale in various real-world applications. More generally, this research provides a valuable example for the broader XAI community, demonstrating how real-world data and carefully designed tasks of increasing complexity can be leveraged for effective and insightful XAI benchmarking. Ultimately, we envision that B-XAIC will facilitate the development of more robust and transparent XAI methods for graph data. This advancement holds a promise for enhancing the interpretability and trustworthiness of GNNs, allowing for their wider adoption in critical production environments, especially in scientific discovery and the design of new therapeutics.

\paragraph{Safeguards.}
The careful design of our molecule selection process and data hosting infrastructure allows us to effectively mitigate the risk of data misuse. The public release of the dataset and source code further supports accessibility and community engagement.

\paragraph{LLMs usage}
We used LLMs only for editorial purposes (grammar and tone).


\subsection{Justification of the dataset size}
\label{app:dataset-size}

Our design philosophy for this benchmark, intentionally balances the simplicity-complexity trade-off. Our aim was to create a dataset that is:
\begin{itemize}
\item \textbf{Sufficiently Compact for Rapid Iteration:} A smaller dataset allows researchers to quickly develop, test, and iterate on new models and XAI methods without extensive computational resources or long training times. This accelerates the research cycle.
\item \textbf{Complex Enough to Reflect Real-World Challenges:} While not encompassing the entirety of chemical space, the seven carefully selected tasks represent a diverse set of common challenges in AI for small molecules. These tasks, ranging from classification to regression-like predictions (e.g., ring count), cover fundamental chemical principles and allow for a robust evaluation of XAI methods' ability to identify relevant features.
\end{itemize}

To further justify our chosen dataset size, we conducted an analysis evaluating model performance across different dataset scales, ranging from 5,000 to 100,000 samples. The results (see Table~\ref{tab:dataset_scales}), presented in the table below, demonstrate that for most tasks, performance largely stabilizes with 50,000 samples.

\begin{table}[htb]
\centering
\caption{Model performance across different dataset scales}
\label{tab:dataset_scales}
\begin{tabular}{lrrrrr}
\toprule
Task & 5K Samples & 10K Samples & 25K Samples & 50K Samples & 100K Samples \\
\midrule
B & 99.92 & 99.92 & 99.96 & 99.96 & 99.94 \\
P & 99.97 & 99.99 & 99.98 & 99.98 & 100.00 \\
X & 99.77 & 99.85 & 99.94 & 99.94 & 99.96 \\
indole & 94.11 & 95.54 & 97.65 & 98.32 & 98.24 \\
PAINS & 82.79 & 85.81 & 91.91 & 92.90 & 93.20 \\
rings-count & 86.86 & 93.14 & 99.64 & 99.62 & 99.94 \\
rings-max & 92.12 & 91.96 & 92.54 & 92.98 & 92.89 \\
\bottomrule
\end{tabular}
\end{table}

The core assumption when creating our dataset was to include simple and moderately complex patterns to achieve two key objectives: (1) to ensure that any GNN can easily learn these patterns, putting focus on the evaluation of XAI methods, and (2) to provide high-quality ground-truth annotations for substructures that directly relate to the instance label. Although simple, these tasks remain relevant for real-world scenarios. For example, the same set of PAINS rules is used in drug discovery to determine if a molecule may interfere with the assay, helping reduce screening costs by filtering out questionable candidates. In practice, evaluating XAI methods with experimental data is challenging due to significant noise and complex interactions between chemical groups, such as intramolecular interactions or steric effects. Therefore, initial evaluation on simple, well-defined tasks is essential for developing effective explainers, especially since many popular methods fail on our straightforward benchmarks.

\subsection{Choice of F1 for classification evaluation}
\label{app:f1_metrics}

In the main paper, we report weighted-F1 and macro-F1 to evaluate
classification performance from complementary perspectives. Weighted-F1
accounts for class imbalance by aggregating per-class performance in
proportion to class frequency, whereas macro-F1 assigns equal importance
to each class, providing a balanced assessment across classes. In this
appendix, we additionally report micro-F1, which aggregates contributions
across all instances, and provide results for model architectures not
included in the main comparison. Table~\ref{tab:f1all}
presents the complete results across all evaluated model architectures,
F1 metrics, and seven classification tasks.

\begin{table}[htb]
\centering
\caption{Complete F1-score results (\%) across all tasks and model
architectures. Results are reported as mean $\pm$ standard deviation.}
\label{tab:f1all}
\scriptsize
\setlength{\tabcolsep}{3pt}
\renewcommand{\arraystretch}{1.2}

\begin{adjustbox}{width=\linewidth}
\begin{tabular}{lccc ccc ccc}
\toprule
& \multicolumn{3}{c}{\textbf{GCN}}
& \multicolumn{3}{c}{\textbf{GAT}}
& \multicolumn{3}{c}{\textbf{GIN}} \\
\cmidrule(lr){2-4}
\cmidrule(lr){5-7}
\cmidrule(lr){8-10}
\textbf{Task}
& Weighted-F1 & Micro-F1 & Macro-F1
& Weighted-F1 & Micro-F1 & Macro-F1
& Weighted-F1 & Micro-F1 & Macro-F1 \\
\midrule
B
& 99.94 $\pm$ 0.05 & 99.99 $\pm$ 0.02 & 99.82 $\pm$ 0.19
& 99.11 $\pm$ 0.19 & 99.98 $\pm$ 0.05 & 99.77 $\pm$ 0.15
& 99.96 $\pm$ 0.05 & 99.95 $\pm$ 0.04 & 99.50 $\pm$ 0.04 \\

P
& 99.98 $\pm$ 0.02 & 99.97 $\pm$ 0.02 & 99.92 $\pm$ 0.05
& 99.97 $\pm$ 0.01 & 99.97 $\pm$ 0.09 & 99.93 $\pm$ 0.08
& 99.98 $\pm$ 0.03 & 99.97 $\pm$ 0.03 & 99.94 $\pm$ 0.07 \\

X
& 99.84 $\pm$ 0.05 & 99.99 $\pm$ 0.02 & 99.99 $\pm$ 0.02
& 99.18 $\pm$ 0.23 & 99.85 $\pm$ 0.13 & 99.85 $\pm$ 0.10
& 99.94 $\pm$ 0.02 & 99.94 $\pm$ 0.02 & 99.94 $\pm$ 0.02 \\

Indole
& 88.33 $\pm$ 1.74 & 84.46 $\pm$ 1.16 & 83.76 $\pm$ 0.95
& 67.30 $\pm$ 2.21 & 67.50 $\pm$ 1.49 & 67.18 $\pm$ 1.79
& 98.32 $\pm$ 0.36 & 98.31 $\pm$ 0.36 & 98.19 $\pm$ 0.38 \\

PAINS
& 79.89 $\pm$ 0.47 & 78.37 $\pm$ 0.91 & 76.10 $\pm$ 0.42
& 65.02 $\pm$ 1.65 & 62.70 $\pm$ 2.91 & 60.67 $\pm$ 0.97
& 92.90 $\pm$ 0.54 & 92.91 $\pm$ 0.51 & 91.94 $\pm$ 0.63 \\

Rings-count
& 87.27 $\pm$ 1.61 & 85.45 $\pm$ 0.74 & 82.75 $\pm$ 0.70
& 82.88 $\pm$ 0.59 & 82.67 $\pm$ 0.83 & 79.92 $\pm$ 0.95
& 99.62 $\pm$ 0.21 & 99.62 $\pm$ 0.21 & 99.54 $\pm$ 0.25 \\

Rings-max
& 91.25 $\pm$ 1.13 & 87.78 $\pm$ 3.02 & 64.45 $\pm$ 2.16
& 91.03 $\pm$ 0.79 & 88.50 $\pm$ 0.79 & 63.55 $\pm$ 1.19
& 92.98 $\pm$ 0.84 & 92.36 $\pm$ 1.46 & 70.01 $\pm$ 2.22 \\
\bottomrule
\end{tabular}
\end{adjustbox}
\vspace{8mm}
\begin{adjustbox}{width=\linewidth}
\begin{tabular}{lccc ccc ccc}
\toprule
& \multicolumn{3}{c}{\textbf{ProtGNN+GCN}}
& \multicolumn{3}{c}{\textbf{ProtGNN+GAT}}
& \multicolumn{3}{c}{\textbf{ProtGNN+GIN}} \\
\cmidrule(lr){2-4}
\cmidrule(lr){5-7}
\cmidrule(lr){8-10}
\textbf{Task}
& Weighted-F1 & Micro-F1 & Macro-F1
& Weighted-F1 & Micro-F1 & Macro-F1
& Weighted-F1 & Micro-F1 & Macro-F1 \\
\midrule
B
& 97.38 $\pm$ 1.42 & 99.43 $\pm$ 0.94 & 87.35 $\pm$ 21.34
& 98.52 $\pm$ 1.69 & 97.83 $\pm$ 0.18 & 62.43 $\pm$ 1.54
& 96.98 $\pm$ 0.69 & 98.15 $\pm$ 0.67 & 82.10 $\pm$ 10.45 \\

P
& 95.52 $\pm$ 7.98 & 98.63 $\pm$ 1.86 & 96.55 $\pm$ 4.00
& 99.51 $\pm$ 0.22 & 96.53 $\pm$ 5.50 & 86.20 $\pm$ 22.81
& 99.77 $\pm$ 0.14 & 99.68 $\pm$ 0.11 & 99.10 $\pm$ 0.23 \\

X
& 99.68 $\pm$ 0.05 & 99.95 $\pm$ 0.09 & 99.99 $\pm$ 0.02
& 99.89 $\pm$ 0.05 & 99.99 $\pm$ 0.02 & 99.99 $\pm$ 0.02
& 98.74 $\pm$ 0.56 & 98.58 $\pm$ 1.07 & 98.55 $\pm$ 1.11 \\

Indole
& 73.80 $\pm$ 16.62 & 74.33 $\pm$ 11.15 & 65.65 $\pm$ 22.96
& 76.17 $\pm$ 2.52 & 76.65 $\pm$ 1.07 & 75.78 $\pm$ 1.24
& 95.76 $\pm$ 0.89 & 95.85 $\pm$ 0.92 & 95.55 $\pm$ 0.96 \\

PAINS
& 56.34 $\pm$ 5.57 & 67.10 $\pm$ 0.16 & 50.18 $\pm$ 1.08
& 54.27 $\pm$ 0.64 & 67.10 $\pm$ 0.16 & 45.17 $\pm$ 0.06
& 85.88 $\pm$ 4.26 & 74.38 $\pm$ 9.02 & 62.63 $\pm$ 19.85 \\

Rings-count
& 68.74 $\pm$ 8.31 & 74.83 $\pm$ 4.83 & 63.58 $\pm$ 16.48
& 71.04 $\pm$ 4.89 & 72.83 $\pm$ 2.26 & 63.48 $\pm$ 11.85
& 83.48 $\pm$ 0.69 & 82.58 $\pm$ 0.58 & 78.65 $\pm$ 0.69 \\

Rings-max
& 91.63 $\pm$ 0.19 & 94.35 $\pm$ 0.11 & 51.52 $\pm$ 1.05
& 91.63 $\pm$ 0.18 & 94.33 $\pm$ 0.15 & 52.64 $\pm$ 0.06
& 91.63 $\pm$ 0.19 & 94.35 $\pm$ 0.51 & 57.55 $\pm$ 1.25 \\
\bottomrule
\end{tabular}
\end{adjustbox}
\end{table}

\subsection{Affine invariance of the IQR-based NE metric}
\label{app:iqr-affine}

We formalize the claim that the NE score, defined through the standard interquartile-range (IQR) outlier rule, is invariant under affine rescaling of the input attributions.

\paragraph{Setup.} Let $x = (x_1, \dots, x_n) \in \mathbb{R}^n$ be a vector of attributions assigned by an explainer to the nodes (or edges) of a graph. We write $Q_p(x)$ for the $p$-th sample quantile of $x$ under the standard linear-interpolation convention, and we extend any function $\psi : \mathbb{R} \to \mathbb{R}$ elementwise to vectors via $\psi(x) = (\psi(x_1), \dots, \psi(x_n))$. Define the interquartile range, Tukey fences, outlier set, and NE score
\begin{align*}
\mathrm{IQR}(x) &\;=\; Q_{0.75}(x) - Q_{0.25}(x), \\
L(x) &\;=\; Q_{0.25}(x) - 1.5\,\mathrm{IQR}(x), \\
U(x) &\;=\; Q_{0.75}(x) + 1.5\,\mathrm{IQR}(x), \\
\mathcal{O}(x) &\;=\; \bigl\{\, i \in \{1, \dots, n\} : x_i < L(x) \text{ or } x_i > U(x) \,\bigr\} \;\subseteq\; \{1, \dots, n\}, \\
s(x) &\;=\; \mathbf{1}\!\left[\mathcal{O}(x) = \varnothing\right].
\end{align*}
Thus $\mathcal{O}(x)$ is the set of indices whose attributions fall outside the Tukey fences, and $s(x) = 1$ exactly when no entry of $x$ is flagged as an outlier.

\begin{proposition}[Affine invariance of the NE metric]
\label{prop:affine-invariance}
For every $x \in \mathbb{R}^n$ and every affine map $\varphi(t) = a t + b$ with $a \neq 0$, the outlier set is preserved:
\begin{equation*}
\mathcal{O}(\varphi(x)) \;=\; \mathcal{O}(x), \qquad \text{and consequently} \qquad s(\varphi(x)) \;=\; s(x).
\end{equation*}
\end{proposition}

\begin{proof}
Under linear interpolation, sample quantiles are equivariant under affine maps $\varphi(t) = at + b$ with $a \neq 0$:
\begin{equation}
Q_p(\varphi(x)) =
\begin{cases}
\varphi\bigl(Q_{p}(x)\bigr) & \text{if } \varphi \text{ is strictly increasing,}\\[2pt]
\varphi\bigl(Q_{1-p}(x)\bigr) & \text{if } \varphi \text{ is strictly decreasing.}
\end{cases}
\label{eq:quantile-equivariance}
\end{equation}

\emph{Case 1: $a > 0$.} The map $\varphi$ is strictly increasing, so by~\eqref{eq:quantile-equivariance}, $Q_p(\varphi(x)) = a\,Q_p(x) + b$ for every $p$. Hence $\mathrm{IQR}(\varphi(x)) = a\,\mathrm{IQR}(x)$, and a direct computation gives $L(\varphi(x)) = \varphi(L(x))$ and $U(\varphi(x)) = \varphi(U(x))$. Because $\varphi$ preserves the order of $\mathbb{R}$, for every index $i \in \{1, \dots, n\}$,
\begin{equation*}
\varphi(x_i) < L(\varphi(x)) \;\Longleftrightarrow\; x_i < L(x), \qquad
\varphi(x_i) > U(\varphi(x)) \;\Longleftrightarrow\; x_i > U(x),
\end{equation*}
so $\mathcal{O}(\varphi(x)) = \mathcal{O}(x)$.

\emph{Case 2: $a < 0$.} Write $a = -|a|$, so that $\varphi$ is strictly decreasing. By~\eqref{eq:quantile-equivariance}, $Q_p(\varphi(x)) = a\,Q_{1-p}(x) + b$, and in particular
\begin{equation*}
Q_{0.25}(\varphi(x)) = a\,Q_{0.75}(x) + b, \qquad Q_{0.75}(\varphi(x)) = a\,Q_{0.25}(x) + b.
\end{equation*}
Subtracting, $\mathrm{IQR}(\varphi(x)) = a\,\bigl(Q_{0.25}(x) - Q_{0.75}(x)\bigr) = -a\,\mathrm{IQR}(x) = |a|\,\mathrm{IQR}(x)$. Substituting $-1.5\,|a| = 1.5\,a$ (valid since $a < 0$),
\begin{align*}
L(\varphi(x)) &= Q_{0.25}(\varphi(x)) - 1.5\,\mathrm{IQR}(\varphi(x)) \\
&= a\,Q_{0.75}(x) + b - 1.5\,|a|\,\mathrm{IQR}(x) \\
&= a\,Q_{0.75}(x) + 1.5\,a\,\mathrm{IQR}(x) + b \\
&= a\bigl(Q_{0.75}(x) + 1.5\,\mathrm{IQR}(x)\bigr) + b \;=\; \varphi(U(x)),
\end{align*}
and symmetrically $U(\varphi(x)) = \varphi(L(x))$: the lower and upper fences are exchanged by $\varphi$. Because $\varphi$ reverses the order of $\mathbb{R}$, for every index $i \in \{1, \dots, n\}$,
\begin{equation*}
\varphi(x_i) < L(\varphi(x)) = \varphi(U(x)) \;\Longleftrightarrow\; x_i > U(x),
\end{equation*}
and likewise $\varphi(x_i) > U(\varphi(x)) = \varphi(L(x)) \Longleftrightarrow x_i < L(x)$. The two outlier conditions thus swap, but their union is the same set of indices, so $\mathcal{O}(\varphi(x)) = \mathcal{O}(x)$.

In either case, $\mathcal{O}(\varphi(x)) = \mathcal{O}(x)$, and consequently $s(\varphi(x)) = s(x)$.
\end{proof}

\paragraph{Implication for normalization.} Two common postprocessing steps for explainer outputs are min-max normalization, $x \mapsto (x - \min x)/(\max x - \min x)$, and standardization, $x \mapsto (x - \overline{x})/\mathrm{std}(x)$. Both are affine transformations with $a > 0$, so by Proposition~\ref{prop:affine-invariance} they leave the NE score invariant. In particular, the IQR-based NE metric requires no per-explainer normalization scheme to be comparable across methods.

\paragraph{Empirical sensitivity analysis}

We evaluate the null-explanation (NE) metric using Tukey fence multipliers of $1.0\times$, $1.5\times$, $2.0\times$, and $3.0\times$ IQR. Although absolute scores increase with wider fences, explainer rankings remain largely stable (Tables~\ref{tab:ne_sensitivity_node}--\ref{tab:ne_sensitivity_ranks}). PGMExplainer consistently ranks first at the node level, while IntegratedGradients ranks last; PGExplainer remains last at the edge level. 
GraphMaskExplainer's edge-level NE score increases only from 0.308 to 0.365 across multipliers, as its near-binary attributions limit the effect of wider fences. Consequently, its ranking drops from first at $1.0\times$ IQR to second to last at $3.0\times$ IQR, driving much of the edge-level ranking variation.
Overall, the results support the robustness of our conclusions across multiplier choices, without implying that the standard $1.5\times$ IQR convention~\citep{Tukey1977} is optimal.

\begin{table}[htb]
\centering
\small
\caption{Node-level NE scores (mean $\pm$ standard deviation) across
seven tasks for different Tukey fence multipliers. Ordered based on $1.5\times IQR$ scores.}
\label{tab:ne_sensitivity_node}
\begin{tabular}{lcccc}
\toprule
Explainer & $1.0\times$ IQR & $1.5\times$ IQR &
$2.0\times$ IQR & $3.0\times$ IQR \\
\midrule
PGMExplainer & $0.918 \pm 0.066$ & $0.979 \pm 0.020$ & $0.989 \pm 0.016$ & $0.997 \pm 0.007$ \\
Deconvolution & $0.519 \pm 0.069$ & $0.788 \pm 0.100$ & $0.888 \pm 0.092$ & $0.956 \pm 0.066$ \\
FlowX & $0.389 \pm 0.103$ & $0.720 \pm 0.127$ & $0.873 \pm 0.097$ & $0.969 \pm 0.044$ \\
GNNExplainer & $0.363 \pm 0.150$ & $0.683 \pm 0.116$ & $0.874 \pm 0.028$ & $0.980 \pm 0.016$ \\
GraphMaskExplainer & $0.266 \pm 0.048$ & $0.656 \pm 0.027$ & $0.855 \pm 0.017$ & $0.987 \pm 0.005$ \\
InputXGradient & $0.273 \pm 0.125$ & $0.536 \pm 0.137$ & $0.713 \pm 0.118$ & $0.894 \pm 0.074$ \\
Saliency & $0.286 \pm 0.112$ & $0.513 \pm 0.124$ & $0.696 \pm 0.121$ & $0.883 \pm 0.069$ \\
ShapleyValueSampling & $0.242 \pm 0.122$ & $0.476 \pm 0.230$ & $0.715 \pm 0.187$ & $0.887 \pm 0.123$ \\
GuidedBackprop & $0.216 \pm 0.081$ & $0.408 \pm 0.121$ & $0.609 \pm 0.145$ & $0.820 \pm 0.094$ \\
IntegratedGradients & $0.207 \pm 0.165$ & $0.386 \pm 0.260$ & $0.509 \pm 0.254$ & $0.695 \pm 0.228$ \\
\bottomrule
\end{tabular}
\end{table}

\begin{table}[htb]
\centering
\small
\caption{Edge-level NE scores (mean $\pm$ standard deviation) across
four tasks for different Tukey fence multipliers. Ordered based on $1.5\times IQR$ scores.}
\label{tab:ne_sensitivity_edge}
\begin{tabular}{lcccc}
\toprule
Explainer & $1.0\times$ IQR & $1.5\times$ IQR &
$2.0\times$ IQR & $3.0\times$ IQR \\
\midrule
GNNExplainer & $0.263 \pm 0.257$ & $0.561 \pm 0.113$ & $0.837 \pm 0.127$ & $0.952 \pm 0.087$ \\
FlowX & $0.181 \pm 0.041$ & $0.550 \pm 0.071$ & $0.807 \pm 0.063$ & $0.951 \pm 0.010$ \\
Deconvolution & $0.163 \pm 0.125$ & $0.387 \pm 0.150$ & $0.527 \pm 0.149$ & $0.863 \pm 0.127$ \\
InputXGradient & $0.152 \pm 0.084$ & $0.369 \pm 0.171$ & $0.580 \pm 0.106$ & $0.880 \pm 0.058$ \\
GraphMaskExplainer & $0.308 \pm 0.043$ & $0.345 \pm 0.039$ & $0.362 \pm 0.047$ & $0.365 \pm 0.047$ \\
IntegratedGradients & $0.157 \pm 0.112$ & $0.317 \pm 0.183$ & $0.525 \pm 0.211$ & $0.769 \pm 0.224$ \\
GuidedBackprop & $0.150 \pm 0.108$ & $0.300 \pm 0.087$ & $0.497 \pm 0.077$ & $0.769 \pm 0.087$ \\
Saliency & $0.135 \pm 0.103$ & $0.297 \pm 0.155$ & $0.490 \pm 0.108$ & $0.775 \pm 0.068$ \\
ShapleyValueSampling & $0.033 \pm 0.023$ & $0.193 \pm 0.097$ & $0.480 \pm 0.075$ & $0.737 \pm 0.054$ \\
PGExplainer & $0.011 \pm 0.018$ & $0.016 \pm 0.023$ & $0.017 \pm 0.027$ & $0.026 \pm 0.029$ \\
\bottomrule
\end{tabular}
\end{table}

\begin{table}[htb]
\centering
\small
\caption{Rank stability relative to the $1.5\times$ IQR setting,
measured by Spearman's $\rho$ and Kendall's $\tau$.}
\label{tab:ne_sensitivity_ranks}
\begin{tabular}{lcccc}
\toprule
& \multicolumn{2}{c}{Node} & \multicolumn{2}{c}{Edge} \\
\cmidrule(lr){2-3} \cmidrule(lr){4-5}
Multiplier & $\rho$ & $\tau$ & $\rho$ & $\tau$ \\
\midrule
$1.0\times$ IQR & 0.952 & 0.867 & 0.855 & 0.778 \\
$2.0\times$ IQR & 0.952 & 0.867 & 0.867 & 0.778 \\
$3.0\times$ IQR & 0.867 & 0.689 & 0.818 & 0.644 \\
\bottomrule
\end{tabular}
\end{table}

\subsection{Motivation for the null explanation evaluation}
\label{app:ne_motiv}

\paragraph{Why coarseness is deliberate}
For negative instances, no target substructure is present, so non-zero attributions represent spurious importance. Since attribution scales vary substantially across explainers, value-based or top-$k$ thresholding would require arbitrary method-specific normalization. Instead, our null-explanation (NE) metric asks whether the attribution vector contains statistical outliers, using Tukey's interquartile range (IQR) fences. This provides a scale-free assessment that is invariant to positive affine transformations of attribution values (Appendix~\ref{app:iqr-affine}).

\paragraph{NE as a diagnostic floor}
The binary NE metric is intended as an assumption-light necessary condition for reliable explanations, not a complete measure of explanation quality. Failing the test indicates that an explainer produces localized attribution spikes on negative instances, where no target structure exists. Passing the test is necessary but not sufficient to establish faithfulness on positive instances.

\paragraph{Continuous extension: Gini coefficient}
To complement the binary NE metric, we examine the Gini coefficient of attribution inequality on negative instances. Lower Gini values indicate more evenly distributed attributions, whereas values closer to one indicate extreme concentration on a small number of nodes or edges despite the absence of a target pattern. Tables~\ref{tab:gini_node} and~\ref{tab:gini_edge} report the node- and edge-level results, respectively.

\begin{table}[t]
\centering
\small
\begin{minipage}[t]{0.48\textwidth}
\centering
\caption{Node-level Gini coefficients on negative instances (mean $\pm$ standard deviation).}
\label{tab:gini_node}
\begin{tabular}{lc}
\toprule
Explainer & Gini coefficient \\
\midrule
Saliency             & 0.429 $\pm$ 0.130 \\
Deconvolution        & 0.810 $\pm$ 0.186 \\
InputXGradient       & 0.947 $\pm$ 0.035 \\
GuidedBackprop       & 0.743 $\pm$ 0.149 \\
IntegratedGradients  & 0.978 $\pm$ 0.013 \\
GNNExplainer         & 1.000 $\pm$ 0.000 \\
GraphMaskExplainer   & 1.000 $\pm$ 0.000 \\
PGMExplainer         & 0.996 $\pm$ 0.001 \\
FlowX                & 1.000 $\pm$ 0.000 \\
ShapleyValueSampling & 0.976 $\pm$ 0.013 \\
\bottomrule
\end{tabular}
\end{minipage}
\hfill
\begin{minipage}[t]{0.48\textwidth}
\centering
\caption{Edge-level Gini coefficients on negative instances (mean $\pm$ standard deviation).}
\label{tab:gini_edge}
\begin{tabular}{lc}
\toprule
Explainer & Gini coefficients \\
\midrule
Saliency             & 0.979 $\pm$ 0.019 \\
Deconvolution        & 0.991 $\pm$ 0.005 \\
InputXGradient       & 0.988 $\pm$ 0.005 \\
GuidedBackprop       & 0.979 $\pm$ 0.010 \\
IntegratedGradients  & 0.999 $\pm$ 0.000 \\
GNNExplainer         & 1.000 $\pm$ 0.000 \\
GraphMaskExplainer   & 1.000 $\pm$ 0.000 \\
PGExplainer          & 1.000 $\pm$ 0.000 \\
FlowX                & 1.000 $\pm$ 0.000 \\
ShapleyValueSampling & 0.999 $\pm$ 0.001 \\
\bottomrule
\end{tabular}
\end{minipage}
\end{table}

The two metrics capture complementary failure modes. Gini measures global attribution concentration, revealing explainers that concentrate importance on a few components even for negative instances. In contrast, NE detects localized outliers, including isolated gradient peaks that may occur even when attribution is globally diffuse. Consequently, gradient-based methods can exhibit relatively low Gini values while still triggering NE alarms, whereas mask-based and other structure-selecting methods may exhibit high attribution concentration. Together, NE and Gini distinguish localized spurious peaks from global attribution concentration, providing a more comprehensive diagnostic of explanation behavior on negative instances.

\subsection{Explanation performance on correct predictions}
\label{ap:corr_explanations}

Table~\ref{tab:node_explanation_corr_full} and
Table~\ref{tab:edge_explanation_corr_full} report node and edge explanation
performance, respectively, for all predictions and for correctly classified
predictions only. These results complement the performance differences
reported in the main paper and provide a more detailed comparison of XAI methods
across both evaluation settings.

\begin{table}[htb]
\centering
\small
\caption{Node explanation performance for all predictions vs. only correct predictions from GIN.}\label{tab:node_explanation_corr_full}
\begin{adjustbox}{width=\linewidth}
\begin{tabular}{llccccccccccccccc}
\toprule
Task & 
Type & 
Subset & 
\rotatebox{90}{Saliency} & 
\rotatebox{90}{Deconv.} & 
\rotatebox{90}{InputXGrad} & 
\rotatebox{90}{GuidedBP} & 
\rotatebox{90}{IntegGrad} & 
\rotatebox{90}{GNNExp.} & 
\rotatebox{90}{GMExp.} & 
\rotatebox{90}{PGMExp.} & 
\rotatebox{90}{FlowX} & 
\rotatebox{90}{SubgraphX} & 
\rotatebox{90}{ShapleyVS} & 
\rotatebox{90}{GSHAP-IQ} & 
\rotatebox{90}{DnX} & 
\rotatebox{90}{FastDnX} \\
\midrule
\multirow{4}{*}{PAINS} & \multirow{2}{*}{NE} & all & 0.39     & 0.81   & 0.44   & 0.32     & 0.07      & 0.68   & 0.64  & 0.99  & 0.7   & 0.99 & 0.21 & 0.62 & 0.09 & 0.03 \\
& & pred=target &  0.42     & 0.81   & 0.44   & 0.32     & 0.07      & 0.68   & 0.64  & 0.99  & 0.72  & 0.99 & 0.21 & 0.62 & 0.09 & 0.03 \\
\cmidrule(lr){2-17}
& \multirow{2}{*}{SE} & all & 0.79     & 0.44   & 0.72   & 0.79     & 0.69      & 0.63   & 0.60   & 0.62  & 0.64  & 0.64 & 0.64 & 0.52 & 0.50 & 0.44 \\
& & pred=target & 0.80      & 0.70    & 0.72   & 0.80      & 0.69      & 0.63   & 0.60   & 0.62  & 0.64  & 0.66 & 0.64 & 0.52 & 0.51 & 0.44 \\
\midrule
\multirow{4}{*}{rings-max} & \multirow{2}{*}{NE} & all         & 0.48     & 0.83   & 0.55   & 0.40      & 0.38      & 0.67   & 0.65  & 0.98  & 0.64  & 0.99 & 0.60 & 0.57 & 0.21 & 0.10 \\
          &    & pred=target & 0.48     & 0.83   & 0.54   & 0.42     & 0.44      & 0.67   & 0.65  & 0.98  & 0.64  & 0.99 & 0.59 & 0.57 & 0.21 & 0.10 \\
\cmidrule(lr){2-17}
          & \multirow{2}{*}{SE} & all         & 0.70      & 0.77   & 0.67   & 0.75     & 0.67      & 0.61   & 0.58  & 0.59  & 0.65  & 0.57 & 0.63 & 0.56 & 0.60 & 0.40 \\
          &    & pred=target & 0.70      & 0.79   & 0.66   & 0.75     & 0.67      & 0.62   & 0.57  & 0.57  & 0.65  & 0.60 & 0.65 & 0.56 & 0.62 & 0.37 \\
\bottomrule
\end{tabular}\end{adjustbox}
\end{table}

\begin{table}[htb]
\centering
\small
\caption{Edge explanation performance for all predictions vs. only correct predictions from GIN.}
\label{tab:edge_explanation_corr_full}
\begin{adjustbox}{width=\linewidth}
\begin{tabular}{llcccccccccccc}
\toprule
Task & 
Type & 
Subset & 
\rotatebox{90}{Saliency} & 
\rotatebox{90}{Deconv.} & 
\rotatebox{90}{InputXGrad} & 
\rotatebox{90}{GuidedBP} & 
\rotatebox{90}{IntegGrad} & 
\rotatebox{90}{GNNExp.} & 
\rotatebox{90}{GMExp.} & 
\rotatebox{90}{PGExp.} & 
\rotatebox{90}{FlowX} & 
\rotatebox{90}{ShapleyVS} & 
\rotatebox{90}{EdgeSHAPer} \\
\midrule
\multirow{4}{*}{PAINS}     & \multirow{2}{*}{NE} & all         & 0.17     & 0.39   & 0.23   & 0.28     & 0.19      & 0.56   & 0.31  & 0.00     & 0.54  & 0.08 & 0.09 \\
          &    & pred=target & 0.18     & 0.44   & 0.23   & 0.28     & 0.19      & 0.50    & 0.31  & 0.00     & 0.55  & 0.08 & 0.09 \\
\cmidrule(lr){2-14}
          & \multirow{2}{*}{SE} & all         & 0.67     & 0.68   & 0.64   & 0.74     & 0.71      & 0.62   & 0.54  & 0.77  & 0.66  & 0.67 & 0.65 \\
          &    & pred=target & 0.68     & 0.69   & 0.64   & 0.75     & 0.72      & 0.62   & 0.54  & 0.78  & 0.67  & 0.67 & 0.66 \\
\cmidrule(lr){1-14}
\multirow{4}{*}{rings-max} & \multirow{2}{*}{NE} & all         & 0.28     & 0.26   & 0.34   & 0.23     & 0.19      & 0.55   & 0.35  & 0.00     & 0.52  & 0.18 & 0.28 \\
          &    & pred=target & 0.28     & 0.27   & 0.37   & 0.23     & 0.22      & 0.55   & 0.35  & 0.00     & 0.50   & 0.17 & 0.28 \\
\cmidrule(lr){2-14}
          & \multirow{2}{*}{SE} & all         & 0.61     & 0.63   & 0.62   & 0.63     & 0.59      & 0.60    & 0.54  & 0.66  & 0.62  & 0.59 & 0.57 \\
          &    & pred=target & 0.60      & 0.64   & 0.61   & 0.64     & 0.59      & 0.60    & 0.54  & 0.66  & 0.58  & 0.59 & 0.59 \\
\bottomrule
\end{tabular}\vspace{-\baselineskip}
\end{adjustbox}
\end{table}

\begin{wraptable}[12]{r}{0.32\textwidth}
\vspace{-\baselineskip}
\centering
\small
\caption{Pearson correlations between F1 and mean explanation score,
$(\mathrm{SE}+\mathrm{NE})/2$.}
\label{tab:f1_explanation_correlations}
\vspace{-0.5\baselineskip}
\begin{tabular}{lrr}
\toprule
Task & Node & Edge \\
\midrule
B & 0.048 & --- \\
P & -0.128 & --- \\
X & 0.199 & --- \\
indole & 0.800 & 0.020 \\
PAINS & 0.504 & -0.269 \\
rings-count & 0.932 & 0.925 \\
rings-max & 0.728 & 0.139 \\
\bottomrule
\end{tabular}
\vspace{-0.5\baselineskip}
\end{wraptable}

\newpage
\subsection{Correlation of Explanation Quality and Model Performance}
\label{app:f1_explanation_correlations}

\begin{figure}[htb]
    \centering
    \includegraphics{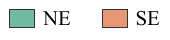}\par\smallskip
    \begin{subfigure}[b]{\linewidth}
        \centering
        \includegraphics[width=\linewidth]{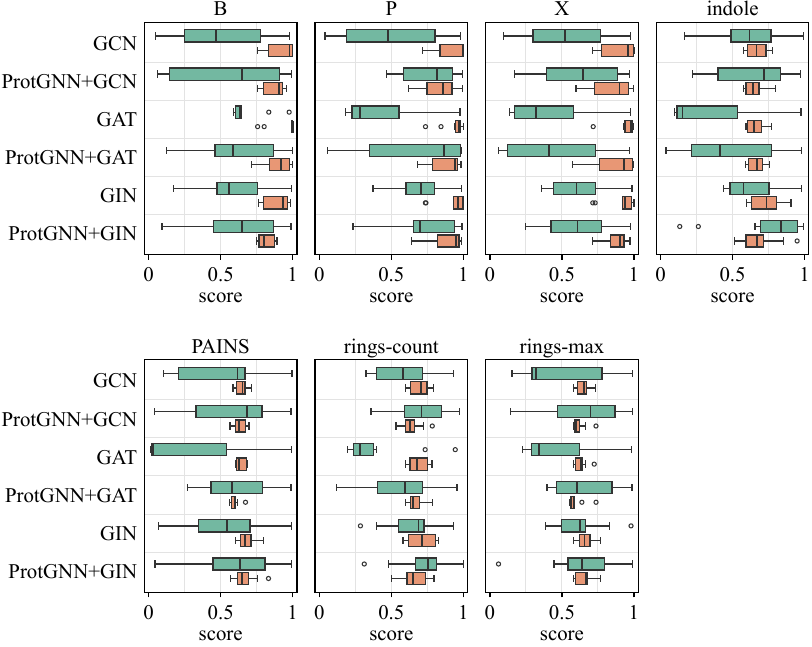}
        \caption{Node-based explanations}
    \end{subfigure}
    \begin{subfigure}[b]{\linewidth}
        \centering
        \includegraphics[width=\linewidth]{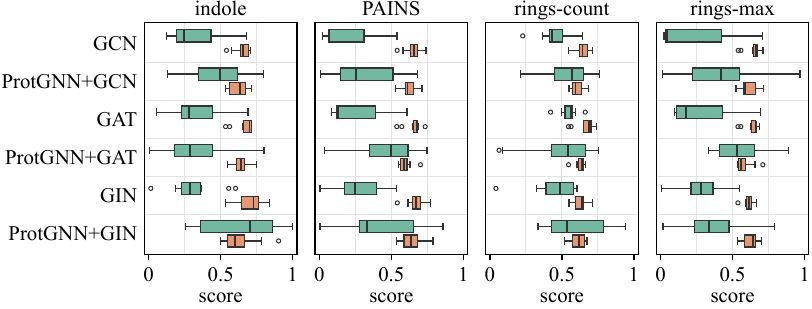}
        \caption{Edge-based explanations}
    \end{subfigure}
    \caption{Boxplots showing the distribution of null explanation (NE) and subgraph explanation (SE) scores across different explainers for each model. Results are aggregated per model, highlighting that some models are inherently more difficult to explain than others.}
    \label{fig:boxplots-all}
\end{figure}

Figure~\ref{fig:boxplots-all} compares NE and SE scores across evaluated
models and illustrates their relationship with model performance across
tasks. To quantify this relationship, we compute the Pearson correlation
between F1 and the mean explanation score, defined as
$(\mathrm{SE}+\mathrm{NE})/2$, for each task and explanation level.
Table~\ref{tab:f1_explanation_correlations} reports the results.

\subsection{Ranking by evaluation scores}

Table~\ref{tab:node-details} and Table~\ref{tab:edge-details} report the evaluation metrics for all model–explainer combinations, averaged across tasks and sorted by the aggregated scores for node and edge explanations, respectively.

\begin{table}[htb]
\centering
\footnotesize
\caption{Ranking of all model–explainer combinations based on the evaluation of \textbf{node explanations}, sorted by overall scores. The best score and all scores not significantly lower (according to a one-sided Wilcoxon test) are highlighted in bold. For NE and SE, we report the mean and standard deviation; for the overall score (avg), we additionally provide the standard error of the mean (SEM) to highlight the trade-off between NE and SE performance.}
\label{tab:node-details}
\begin{adjustbox}{width=0.63\linewidth}
\begin{tabular}{llccc}
\toprule
Model       & Explainer            & NE & SE & avg \\
\midrule 
ProtGNN+GCN & PGMExplainer & \textbf{0.98$\pm$0.02} & 0.81$\pm$0.20 & \textbf{0.89$\pm$0.10} \\ 
GIN & PGMExplainer & \textbf{0.98$\pm$0.02} & 0.76$\pm$0.18 &\textbf{ 0.87$\pm$0.10} \\ 
GAT & PGMExplainer & \textbf{0.98$\pm$0.02} & 0.75$\pm$0.16 & \textbf{0.86$\pm$0.08} \\ 
ProtGNN+GIN & PGMExplainer & \textbf{0.98$\pm$0.02} & 0.73$\pm$0.17 & \textbf{0.85$\pm$0.09} \\ 
GCN & PGMExplainer & \textbf{0.98$\pm$0.02} & 0.72$\pm$0.11 & \textbf{0.85$\pm$0.06} \\ 
ProtGNN+GAT & PGMExplainer & \textbf{0.98$\pm$0.01} & 0.70$\pm$0.19 & \textbf{0.84$\pm$0.10} \\ 
GIN & SubgraphX & \textbf{0.94$\pm$0.05} & 0.74$\pm$0.10 & \textbf{0.84$\pm$0.05} \\
GAT & SubgraphX & \textbf{0.92$\pm$0.13} & 0.71$\pm$0.12 & \textbf{0.82$\pm$0.10} \\
GIN & Deconvolution & 0.79$\pm$0.10 & 0.81$\pm$0.09 & 0.80$\pm$0.07 \\ 
GCN & SubgraphX & \textbf{0.91$\pm$0.15} & 0.69$\pm$0.08 & \textbf{0.80$\pm$0.10} \\
ProtGNN+GCN & FlowX & 0.81$\pm$0.15 & 0.79$\pm$0.06 & 0.80$\pm$0.09 \\ 
ProtGNN+GIN & FlowX & 0.78$\pm$0.14 & 0.79$\pm$0.11 & 0.79$\pm$0.09 \\ 
ProtGNN+GCN & Deconvolution & 0.80$\pm$0.11 & 0.77$\pm$0.16 & 0.79$\pm$0.10 \\ 
ProtGNN+GIN & Deconvolution & 0.78$\pm$0.13 & 0.79$\pm$0.11 & 0.79$\pm$0.10 \\ 
ProtGNN+GAT & FlowX & 0.79$\pm$0.15 & 0.80$\pm$0.08 & 0.79$\pm$0.09 \\ 
GCN & Deconvolution & 0.78$\pm$0.11 & 0.77$\pm$0.16 & 0.78$\pm$0.10 \\ 
ProtGNN+GCN & GuidedBackprop & 0.90$\pm$0.16 & 0.64$\pm$0.11 & 0.77$\pm$0.11 \\ 
GCN & FlowX & 0.75$\pm$0.14 & 0.79$\pm$0.10 & 0.77$\pm$0.09 \\ 
GAT & FlowX & 0.73$\pm$0.10 & 0.79$\pm$0.11 & 0.76$\pm$0.08 \\ 
ProtGNN+GIN & Saliency & 0.63$\pm$0.21 & 0.84$\pm$0.10 & 0.74$\pm$0.12 \\ 
ProtGNN+GIN & GuidedBackprop & 0.80$\pm$0.26 & 0.67$\pm$0.13 & 0.74$\pm$0.12 \\ 
GIN & FlowX & 0.70$\pm$0.15 & 0.77$\pm$0.12 & 0.73$\pm$0.09 \\ 
ProtGNN+GAT & GuidedBackprop & 0.74$\pm$0.33 & 0.66$\pm$0.15 & 0.70$\pm$0.21 \\ 
ProtGNN+GCN & GNNExplainer & 0.77$\pm$0.17 & 0.63$\pm$0.10 & 0.70$\pm$0.06 \\ 
ProtGNN+GIN & GNNExplainer & 0.74$\pm$0.18 & 0.64$\pm$0.11 & 0.69$\pm$0.07 \\ 
GIN & GraphSHAP-IQ & 0.64$\pm$0.09 & 0.72$\pm$0.25 & 0.68$\pm$0.13 \\ 
GIN & Saliency & 0.51$\pm$0.12 & \textbf{0.85$\pm$0.12} & 0.68$\pm$0.10 \\ 
GIN & InputXGradient & 0.54$\pm$0.14 & 0.83$\pm$0.14 & 0.68$\pm$0.10 \\ 
GIN & GNNExplainer & 0.68$\pm$0.12 & 0.67$\pm$0.08 & 0.68$\pm$0.06 \\ 
GCN & GuidedBackprop & 0.51$\pm$0.22 & \textbf{0.84$\pm$0.13} & 0.68$\pm$0.10 \\ 
ProtGNN+GCN & Saliency & 0.62$\pm$0.23 & 0.74$\pm$0.14 & 0.68$\pm$0.11 \\ 
ProtGNN+GIN & InputXGradient & 0.55$\pm$0.27 & 0.80$\pm$0.13 & 0.68$\pm$0.14 \\ 
ProtGNN+GAT & GNNExplainer & 0.70$\pm$0.18 & 0.65$\pm$0.09 & 0.68$\pm$0.10 \\ 
ProtGNN+GAT & Saliency & 0.53$\pm$0.35 & 0.80$\pm$0.18 & 0.67$\pm$0.19 \\ 
GIN & GraphMaskExplainer & 0.66$\pm$0.03 & 0.66$\pm$0.07 & 0.66$\pm$0.03 \\ 
ProtGNN+GIN & GraphMaskExplainer & 0.66$\pm$0.03 & 0.66$\pm$0.07 & 0.66$\pm$0.03 \\ 
GCN & GraphSHAP-IQ & 0.58$\pm$0.15 & 0.73$\pm$0.24 & 0.66$\pm$0.11 \\ 
GCN & GraphMaskExplainer & 0.65$\pm$0.04 & 0.66$\pm$0.07 & 0.65$\pm$0.03 \\ 
ProtGNN+GCN & GraphMaskExplainer & 0.65$\pm$0.04 & 0.66$\pm$0.07 & 0.65$\pm$0.03 \\ 
GAT & GraphMaskExplainer & 0.65$\pm$0.04 & 0.66$\pm$0.07 & 0.65$\pm$0.03 \\ 
ProtGNN+GAT & GraphMaskExplainer & 0.65$\pm$0.04 & 0.66$\pm$0.07 & 0.65$\pm$0.03 \\ 
GIN & ShapleyValueSampling & 0.48$\pm$0.23 & 0.83$\pm$0.15 & 0.65$\pm$0.14 \\ 
ProtGNN+GAT & InputXGradient & 0.49$\pm$0.38 & 0.79$\pm$0.18 & 0.64$\pm$0.21 \\ 
GIN & GuidedBackprop & 0.41$\pm$0.12 & \textbf{0.87$\pm$0.09} & 0.64$\pm$0.09 \\ 
GCN & Saliency & 0.43$\pm$0.13 & \textbf{0.84$\pm$0.14} & 0.63$\pm$0.08 \\ 
ProtGNN+GIN & IntegratedGradients & 0.44$\pm$0.24 & 0.82$\pm$0.12 & 0.63$\pm$0.13 \\ 
GIN & IntegratedGradients & 0.39$\pm$0.26 & \textbf{0.85$\pm$0.13} & 0.62$\pm$0.16 \\ 
ProtGNN+GAT & IntegratedGradients & 0.41$\pm$0.36 & 0.79$\pm$0.18 & 0.60$\pm$0.20 \\ 
GCN & GNNExplainer & 0.53$\pm$0.17 & 0.67$\pm$0.08 & 0.60$\pm$0.09 \\ 
ProtGNN+GCN & InputXGradient & 0.36$\pm$0.32 & 0.76$\pm$0.17 & 0.56$\pm$0.14 \\ 
GAT & GNNExplainer & 0.37$\pm$0.17 & 0.76$\pm$0.16 & 0.56$\pm$0.13 \\ 
GAT & InputXGradient & 0.26$\pm$0.22 & 0.83$\pm$0.15 & 0.55$\pm$0.16 \\ 
GAT & ShapleyValueSampling & 0.27$\pm$0.20 & 0.82$\pm$0.15 & 0.54$\pm$0.15 \\ 
ProtGNN+GCN & IntegratedGradients & 0.29$\pm$0.27 & 0.80$\pm$0.15 & 0.54$\pm$0.14 \\ 
GAT & IntegratedGradients & 0.25$\pm$0.20 & 0.82$\pm$0.15 & 0.54$\pm$0.15 \\ 
GCN & ShapleyValueSampling & 0.26$\pm$0.15 & 0.80$\pm$0.17 & 0.53$\pm$0.10 \\ 
GCN & InputXGradient & 0.21$\pm$0.19 & 0.83$\pm$0.16 & 0.52$\pm$0.10 \\ 
GAT & GuidedBackprop & 0.26$\pm$0.23 & 0.78$\pm$0.18 & 0.52$\pm$0.17 \\ 
GAT & Deconvolution & 0.26$\pm$0.23 & 0.78$\pm$0.18 & 0.52$\pm$0.17 \\ 
GAT & Saliency & 0.24$\pm$0.22 & 0.79$\pm$0.18 & 0.51$\pm$0.17 \\ 
ProtGNN+GAT & Deconvolution & 0.26$\pm$0.20 & 0.74$\pm$0.14 & 0.50$\pm$0.10 \\
GAT & FastDnX & 0.31$\pm$0.19 & 0.67$\pm$0.30 & 0.49$\pm$0.18 \\
ProtGNN+GIN & ShapleyValueSampling & 0.24$\pm$0.24 & 0.74$\pm$0.14 & 0.49$\pm$0.14 \\ 
ProtGNN+GCN & ShapleyValueSampling & 0.25$\pm$0.27 & 0.73$\pm$0.15 & 0.49$\pm$0.14 \\ 
GCN & IntegratedGradients & 0.16$\pm$0.13 & 0.82$\pm$0.16 & 0.49$\pm$0.07 \\ 
ProtGNN+GAT & ShapleyValueSampling & 0.21$\pm$0.31 & 0.75$\pm$0.18 & 0.48$\pm$0.17 \\ 
GCN & DnX & 0.32$\pm$0.24 & 0.56$\pm$0.13 & 0.44$\pm$0.15 \\
GCN & FastDnX & 0.14$\pm$0.15 & 0.73$\pm$0.24 & 0.44$\pm$0.16 \\
GAT & GraphSHAP-IQ & 0.18$\pm$0.14 & 0.70$\pm$0.29 & 0.44$\pm$0.14 \\ 
GIN & DnX & 0.22$\pm$0.24 & 0.60$\pm$0.16 & 0.41$\pm$0.15 \\
GIN & FastDnX & 0.12$\pm$0.15 & 0.69$\pm$0.28 & 0.41$\pm$0.18 \\
GAT & DnX & 0.16$\pm$0.22 & 0.40$\pm$0.20 & 0.28$\pm$0.18 \\
\bottomrule
\end{tabular}
\end{adjustbox}
\end{table}

\begin{table}[htb]
\centering
\caption{Ranking of all model–explainer combinations based on the evaluation of \textbf{edge explanations}, sorted by overall scores, following the format of Table~\ref{tab:node-details}.}
\label{tab:edge-details}
\footnotesize
\begin{adjustbox}{width=0.75\linewidth}
\begin{tabular}{llccc}
\toprule
Model       & Explainer            & NE & SE & avg  \\
\midrule 
ProtGNN+GIN & GuidedBackprop & \textbf{0.83$\pm$0.16} & 0.59$\pm$0.05 & \textbf{0.71$\pm$0.07} \\ 
ProtGNN+GIN & FlowX & 0.68$\pm$0.14 & 0.68$\pm$0.01 & \textbf{0.68$\pm$0.07} \\ 
ProtGNN+GCN & GuidedBackprop & 0.77$\pm$0.24 & 0.59$\pm$0.07 & \textbf{0.68$\pm$0.13} \\ 
ProtGNN+GAT & GuidedBackprop & 0.76$\pm$0.22 & 0.59$\pm$0.06 & 0.67$\pm$0.11 \\ 
ProtGNN+GIN & GNNExplainer & 0.80$\pm$0.22 & 0.54$\pm$0.04 & 0.67$\pm$0.10 \\ 
ProtGNN+GIN & Saliency & 0.58$\pm$0.30 & \textbf{0.75$\pm$0.11} & 0.67$\pm$0.18 \\ 
ProtGNN+GAT & FlowX & 0.61$\pm$0.08 & 0.68$\pm$0.02 & 0.64$\pm$0.05 \\ 
ProtGNN+GCN & FlowX & 0.61$\pm$0.07 & 0.68$\pm$0.03 & 0.64$\pm$0.04 \\ 
GAT & FlowX & 0.60$\pm$0.08 & 0.67$\pm$0.02 & 0.63$\pm$0.05 \\ 
GCN & FlowX & 0.59$\pm$0.11 & 0.67$\pm$0.03 & 0.63$\pm$0.06 \\ 
ProtGNN+GIN & InputXGradient & 0.58$\pm$0.30 & 0.66$\pm$0.05 & 0.62$\pm$0.16 \\ 
ProtGNN+GCN & GNNExplainer & 0.65$\pm$0.23 & 0.57$\pm$0.05 & 0.61$\pm$0.10 \\ 
GIN & FlowX & 0.54$\pm$0.06 & 0.64$\pm$0.03 & 0.59$\pm$0.04 \\ 
ProtGNN+GIN & IntegratedGradients & 0.49$\pm$0.30 & 0.70$\pm$0.07 & 0.59$\pm$0.16 \\ 
ProtGNN+GAT & GNNExplainer & 0.61$\pm$0.24 & 0.58$\pm$0.04 & 0.59$\pm$0.11 \\ 
GIN & GNNExplainer & 0.56$\pm$0.11 & 0.59$\pm$0.03 & 0.58$\pm$0.05 \\ 
ProtGNN+GAT & PGExplainer & 0.51$\pm$0.32 & 0.63$\pm$0.09 & 0.57$\pm$0.13 \\ 
GAT & GNNExplainer & 0.56$\pm$0.04 & 0.56$\pm$0.01 & 0.56$\pm$0.02 \\ 
ProtGNN+GCN & Deconvolution & 0.42$\pm$0.21 & 0.68$\pm$0.03 & 0.55$\pm$0.10 \\ 
ProtGNN+GAT & Saliency & 0.46$\pm$0.32 & 0.64$\pm$0.09 & 0.55$\pm$0.13 \\ 
ProtGNN+GCN & GraphMaskExplainer & 0.55$\pm$0.05 & 0.54$\pm$0.01 & 0.54$\pm$0.03 \\ 
GIN & Deconvolution & 0.39$\pm$0.15 & 0.69$\pm$0.04 & 0.54$\pm$0.08 \\ 
ProtGNN+GAT & GraphMaskExplainer & 0.51$\pm$0.05 & 0.55$\pm$0.00 & 0.53$\pm$0.03 \\ 
GCN & GraphMaskExplainer & 0.50$\pm$0.06 & 0.54$\pm$0.00 & 0.52$\pm$0.03 \\ 
ProtGNN+GAT & InputXGradient & 0.45$\pm$0.32 & 0.59$\pm$0.06 & 0.52$\pm$0.15 \\ 
ProtGNN+GAT & Deconvolution & 0.39$\pm$0.26 & 0.65$\pm$0.04 & 0.52$\pm$0.13 \\ 
ProtGNN+GCN & Saliency & 0.39$\pm$0.32 & 0.64$\pm$0.10 & 0.51$\pm$0.14 \\ 
GAT & GraphMaskExplainer & 0.48$\pm$0.05 & 0.54$\pm$0.00 & 0.51$\pm$0.02 \\ 
GIN & InputXGradient & 0.37$\pm$0.17 & 0.65$\pm$0.05 & 0.51$\pm$0.08 \\ 
GIN & IntegratedGradients & 0.32$\pm$0.18 & 0.70$\pm$0.07 & 0.51$\pm$0.11 \\ 
GCN & Deconvolution & 0.34$\pm$0.22 & 0.67$\pm$0.02 & 0.50$\pm$0.12 \\ 
GAT & Saliency & 0.34$\pm$0.28 & 0.67$\pm$0.03 & 0.50$\pm$0.14 \\ 
GIN & GuidedBackprop & 0.30$\pm$0.09 & 0.70$\pm$0.10 & 0.50$\pm$0.05 \\ 
ProtGNN+GCN & InputXGradient & 0.38$\pm$0.32 & 0.60$\pm$0.07 & 0.49$\pm$0.15 \\ 
GIN & Saliency & 0.30$\pm$0.15 & 0.67$\pm$0.06 & 0.48$\pm$0.07 \\ 
GCN & GuidedBackprop & 0.30$\pm$0.27 & 0.66$\pm$0.03 & 0.48$\pm$0.14 \\ 
ProtGNN+GIN & Deconvolution & 0.32$\pm$0.18 & 0.63$\pm$0.04 & 0.48$\pm$0.09 \\ 
GAT & InputXGradient & 0.26$\pm$0.23 & 0.69$\pm$0.03 & 0.48$\pm$0.12 \\ 
GAT & Deconvolution & 0.26$\pm$0.23 & 0.69$\pm$0.03 & 0.48$\pm$0.12 \\ 
GAT & GuidedBackprop & 0.26$\pm$0.23 & 0.69$\pm$0.03 & 0.48$\pm$0.12 \\ 
ProtGNN+GCN & PGExplainer & 0.32$\pm$0.34 & 0.63$\pm$0.07 & 0.47$\pm$0.15 \\ 
GAT & IntegratedGradients & 0.26$\pm$0.23 & 0.69$\pm$0.02 & 0.47$\pm$0.12 \\ 
GAT & ShapleyValueSampling & 0.28$\pm$0.24 & 0.66$\pm$0.03 & 0.47$\pm$0.12 \\ 
GCN & GNNExplainer & 0.36$\pm$0.23 & 0.57$\pm$0.02 & 0.47$\pm$0.11 \\ 
GAT & PGExplainer & 0.22$\pm$0.23 & 0.72$\pm$0.03 & 0.47$\pm$0.12 \\ 
GCN & InputXGradient & 0.21$\pm$0.20 & 0.68$\pm$0.04 & 0.45$\pm$0.11 \\ 
ProtGNN+GIN & GraphMaskExplainer & 0.35$\pm$0.04 & 0.54$\pm$0.01 & 0.44$\pm$0.02 \\ 
GIN & GraphMaskExplainer & 0.35$\pm$0.04 & 0.54$\pm$0.01 & 0.44$\pm$0.02 \\ 
GCN & IntegratedGradients & 0.20$\pm$0.16 & 0.66$\pm$0.03 & 0.43$\pm$0.08 \\ 
ProtGNN+GCN & IntegratedGradients & 0.19$\pm$0.24 & 0.65$\pm$0.06 & 0.42$\pm$0.12 \\ 
GIN & ShapleyValueSampling & 0.19$\pm$0.10 & 0.64$\pm$0.03 & 0.42$\pm$0.05 \\ 
GCN & Saliency & 0.17$\pm$0.15 & 0.65$\pm$0.04 & 0.41$\pm$0.08 \\ 
GCN & PGExplainer & 0.11$\pm$0.10 & 0.71$\pm$0.03 & 0.41$\pm$0.05 \\ 
GCN & ShapleyValueSampling & 0.16$\pm$0.17 & 0.65$\pm$0.03 & 0.41$\pm$0.08 \\ 
ProtGNN+GIN & ShapleyValueSampling & 0.21$\pm$0.27 & 0.58$\pm$0.02 & 0.40$\pm$0.13 \\ 
GIN & EdgeSHAPer & 0.20$\pm$0.11 & 0.58$\pm$0.09 & 0.39$\pm$0.03 \\ 
ProtGNN+GIN & PGExplainer & 0.16$\pm$0.19 & 0.62$\pm$0.05 & 0.39$\pm$0.09 \\ 
ProtGNN+GAT & IntegratedGradients & 0.15$\pm$0.29 & 0.62$\pm$0.05 & 0.38$\pm$0.13 \\ 
GIN & PGExplainer & 0.02$\pm$0.02 & 0.72$\pm$0.06 & 0.37$\pm$0.03 \\ 
ProtGNN+GAT & ShapleyValueSampling & 0.13$\pm$0.29 & 0.59$\pm$0.04 & 0.36$\pm$0.13 \\ 
GCN & EdgeSHAPer & 0.15$\pm$0.15 & 0.56$\pm$0.08 & 0.36$\pm$0.05 \\ 
ProtGNN+GCN & ShapleyValueSampling & 0.12$\pm$0.23 & 0.59$\pm$0.04 & 0.35$\pm$0.11 \\ 
GAT & EdgeSHAPer & 0.19$\pm$0.22 & 0.48$\pm$0.12 & 0.34$\pm$0.10 \\ 
\bottomrule
\end{tabular}
\end{adjustbox}
\end{table}

\subsection{Analytical ground-truth game and exact interaction evaluation}
\label{app:vt}
 
The benchmark so far evaluates explainers of a \emph{trained} GNN. Because every \ours task is defined by an explicit structural rule, we can go further and evaluate explainers against an \emph{exact}, model-independent ground truth, removing the model-capacity confound entirely. This makes B-XAIC usable as a unit test for Shapley- and interaction-based explainers.
 
\paragraph{Game formulation.}
Let $G=(V,E)$ be a molecular graph and $f$ the Boolean labeling rule of a task. For a coalition of retained atoms $S\subseteq V$, let $G[S]$ be the node-induced subgraph (atoms $V\setminus S$ and their incident edges deleted). We define the cooperative game
\begin{equation}
\nu(S) \;=\; f\bigl(G[S]\bigr), \qquad \nu:2^{V}\to\{0,1\}.
\label{eq:vt}
\end{equation}
Because $\nu$ \emph{is} the rule, its exact Shapley values and M\"obius (interaction) transform are, by construction, the ground-truth attribution and the ground-truth higher-order interactions of the task: no model is trained and no attribution is thresholded. Any Shapley or interaction explainer can thus be unit-tested against an exact target on every task, not only single-atom ones. We evaluate $\nu$ on the graph (NetworkX induced subgraph with VF2 subgraph matching for pattern tasks, circuit rank for ring tasks) rather than by re-sanitizing an RDKit fragment, which fails on broken aromatic rings. For the element/pattern/large-ring tasks $\nu$ is \emph{local}: it depends only on the pattern atoms, so the exact $2^{n}$ M\"obius transform is taken over a small support and all other atoms provably receive zero interaction, making the construction exact for molecules of any size.
 
\paragraph{Intrinsic interaction order.}
We define the \emph{intrinsic interaction order} of a task on a molecule as the largest cardinality of a non-zero M\"obius interaction of $\nu$. It is exact, model-independent, and measures how high an interaction order any explainer must represent to be faithful. For the tasks whose game is an OR of unanimity games (element presence, pattern presence, large-ring presence) the largest non-zero M\"obius term is the union of all ground-truth atoms, so the intrinsic order equals the number of ground-truth atoms. Table~\ref{tab:vt-io} reports the full-dataset distribution over all $50{,}000$ molecules. On every molecule, $\nu(V)$ reproduces the published label ($100\%$ agreement), confirming the rule implementation. \texttt{rings-count} is excluded: its game is circuit-rank based and genuinely global, so the exact $2^{n}$ transform is tractable only for molecules with $\lesssim 20$ ring atoms (a sparse-M\"obius extension over ring occurrences is left to future work).
 
\begin{table}[htb]
\centering
\small
\caption{Full-dataset (50K molecules) intrinsic interaction order of the analytical game $\nu$. On all molecules $\nu(V)$ matches the published label.}
\label{tab:vt-io}
\begin{tabular}{lccccc}
\toprule
 & & \multicolumn{4}{c}{Intrinsic interaction order} \\
\cmidrule(lr){3-6}
Task & \# positive & min & median & mean & max \\
\midrule
B          & 1{,}116  & 1 & 1 & 1.0  & 5 \\
P          & 6{,}612  & 1 & 1 & 1.5  & 29 \\
X          & 27{,}934 & 1 & 1 & 1.8  & 20 \\
indole     & 18{,}403 & 9 & 9 & 10.3 & 108 \\
PAINS      & 16{,}446 & 2 & 9 & 9.7  & 90 \\
rings-max  & 2{,}841  & 7 & 9 & 18.1 & 490 \\
\bottomrule
\end{tabular}
\end{table}
 
\paragraph{Sanity check.} As an independent verification, we recomputed the M\"obius interactions of $\nu$ with the external library \texttt{shapiq}~\citep{muschalik2025exact} on representative molecules spanning all game shapes (single-atom OR, connected conjunction, PAINS, large ring); the maximum discrepancy against our closed-form transform was $0$.
 
\paragraph{Four failure modes of interaction methods.} The exact game makes the limitations of Shapley-interaction explainers (e.g.\ GraphSHAP-IQ~\citep{muschalik2025exact}) precise rather than empirical.
\begin{enumerate}[leftmargin=*]
\item \textbf{Order truncation.} Any method capped at interaction order $L$ below a task's intrinsic order cannot represent it: the truncated reconstruction of $\nu$ has strictly positive error for every such $L$ and is exactly zero at the intrinsic order (constant across $L$ for a single-term conjunction such as indole). Truncation also \emph{fabricates} low-order terms: a pure $k$-way conjunction has zero true pairwise interaction, yet any order-2 method must spread mass over all $\binom{k}{2}$ pairs to fit it ($36$ ground-truth-zero pairs for indole, $k=9$).
\item \textbf{Cost of exactness.} To be exact, the order budget $\lambda$ must reach the intrinsic order; on ring and indole tasks that order is comparable to the receptive-field size, so exactness collides with GraphSHAP-IQ's own exponential receptive-field complexity.
\item \textbf{Masking mismatch.} GraphSHAP-IQ uses BShap baseline-feature imputation, whereas the rule (and $\nu$) uses node deletion. These are different games, so even an exact GraphSHAP-IQ explains a different set function than the one that defines the label.
\item \textbf{Assumption violated by ProtGNN.} GraphSHAP-IQ's exactness assumes linear global pooling and readout (its Assumption~3.4), which excludes ProtGNN and other deep-readout backbones evaluated in \ours.
\end{enumerate}
 
\paragraph{Exact explainer evaluation on a trained GIN.}
We also computed the exact Shapley values and M\"obius interactions of the trained GIN under node-deletion masking, on a small-molecule subset per task so that the $2^{n}$ game is exact ($22$--$100$ molecules/task). Because both $\nu$ and the model use deletion masking, the masking-mismatch term vanishes by construction, so the gap to the reference decomposes as model-error plus, at order cap $\lambda$, $\lambda$-truncation (reported as norms over the interaction vector). Table~\ref{tab:vt-onmodel} shows that atom-level recovery against the exact ground truth is near-perfect for low-order tasks but degrades sharply for the high-order ones (PAINS, rings-max), with individual PAINS molecules reaching AUROC $0.10$ (the model's Shapley \emph{anti-correlates} with the true substructure), despite F1$\ge 0.92$. Even where recovery is perfect, the model-error and $\lambda=2$ truncation norms are large, showing the GIN represents each rule through a broad spread of higher-order interactions rather than the clean ground-truth term.
 
\begin{table}[htb]
\centering
\small
\caption{Exact Shapley/M\"obius evaluation of the trained GIN against the analytical reference $\nu$ (node-deletion masking; small-molecule subset so $2^{n}$ is exact). Norms are over the interaction vector.}
\label{tab:vt-onmodel}
\begin{tabular}{lcccccc}
\toprule
Task & GIN F1 & \# mols & \shortstack{intrinsic\\order} &
\shortstack{atom-recovery\\AUROC (mean / min)} &
\shortstack{model-error\\$\lVert\cdot\rVert$} &
\shortstack{$\lambda{=}2$ trunc.\\$\lVert\cdot\rVert$} \\
\midrule
B         & 1.00 & 87  & 1    & 1.00 / 1.00 & 2.07 & 1.61 \\
P         & 1.00 & 100 & 1    & 1.00 / 1.00 & 2.42 & 1.96 \\
X         & 1.00 & 100 & 1    & 1.00 / 0.91 & 1.52 & 0.98 \\
indole    & 0.98 & 78  & 9    & 0.98 / 0.82 & 1.81 & 1.64 \\
PAINS     & 0.93 & 100 & 2--12 & 0.77 / 0.10 & 3.41 & 3.12 \\
rings-max & 0.92 & 22  & 7    & 0.76 / 0.54 & 1.84 & 1.47 \\
\bottomrule
\end{tabular}
\end{table}
 
The analytical-game unit-testing code is released with the benchmark. Results here use the GIN backbone.

\subsection{Visual summary of evaluation scores}
\begin{figure*}[htb]
    \centering
    \scriptsize

    \begin{tikzpicture}[x=1cm, y=1cm] 
        \node[anchor=east, minimum height=1.5cm, minimum width=2cm, align=center] at (-3, 0) {GCN}; 
        \node at (2.5, 0) {\includegraphics[width=0.8\linewidth]{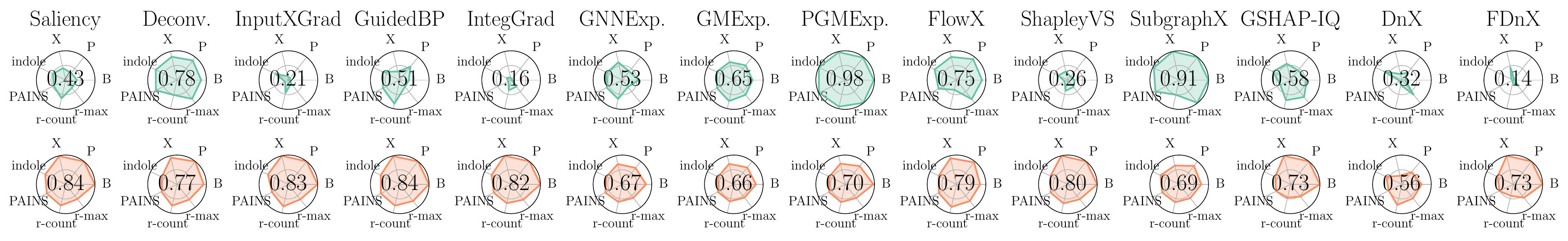}};
    \end{tikzpicture}

    \begin{tikzpicture}[x=1cm, y=1cm]
        \node[anchor=east, minimum height=1.5cm, minimum width=2cm, align=center] at (-3, 0) {ProtGNN\\+GCN};  
        \node at (2.5, 0) {\includegraphics[width=0.8\linewidth]{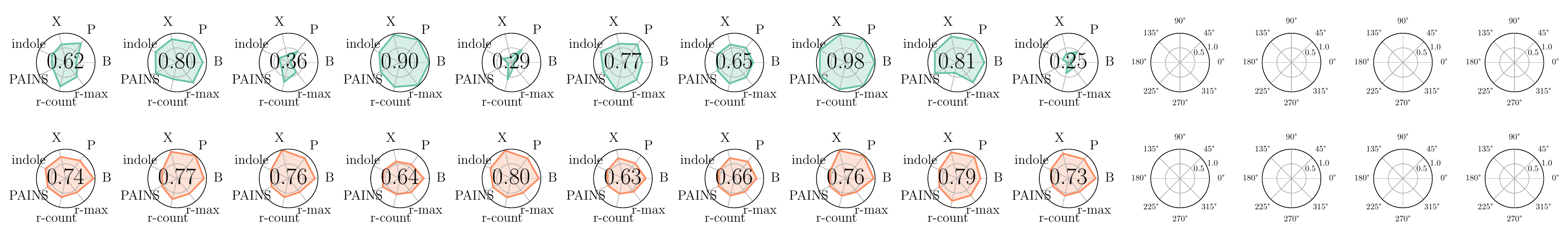}};
    \end{tikzpicture}

    \begin{tikzpicture}[x=1cm, y=1cm]
        \node[anchor=east, minimum height=1.5cm, minimum width=2cm, align=center] at (-3, 0) {GAT};  
        \node at (2.5, 0) {\includegraphics[width=0.8\linewidth]{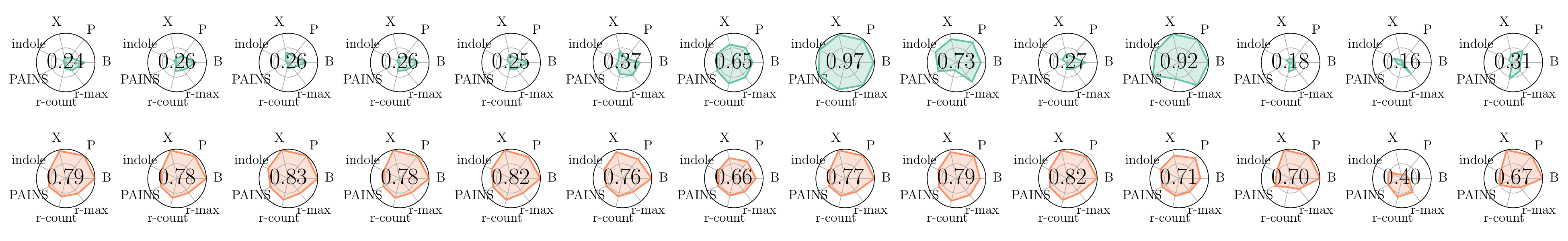}};
    \end{tikzpicture}

    \begin{tikzpicture}[x=1cm, y=1cm]
        \node[anchor=east, minimum height=1.5cm, minimum width=2cm, align=center] at (-3, 0) {ProtGNN\\+GAT};  
        \node at (2.5, 0) {\includegraphics[width=0.8\linewidth]{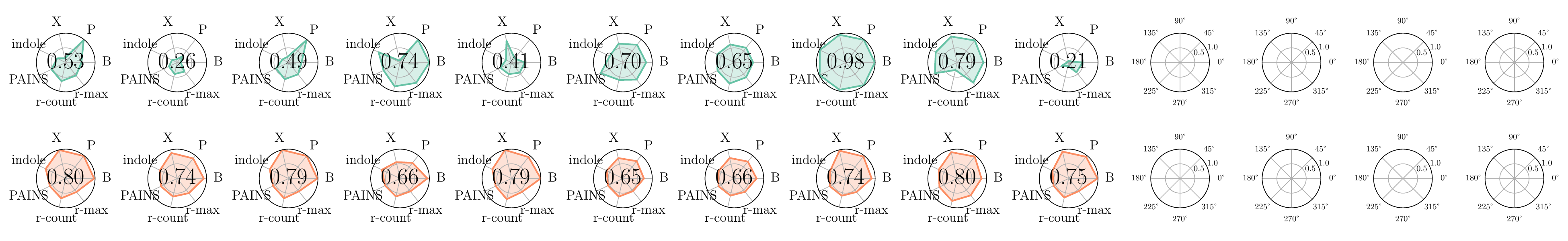}};
    \end{tikzpicture}

    \begin{tikzpicture}[x=1cm, y=1cm]
        \node[anchor=east, minimum height=1.5cm, minimum width=2cm, align=center] at (-3, 0) {GIN};  
        \node at (2.5, 0) {\includegraphics[width=0.8\linewidth]{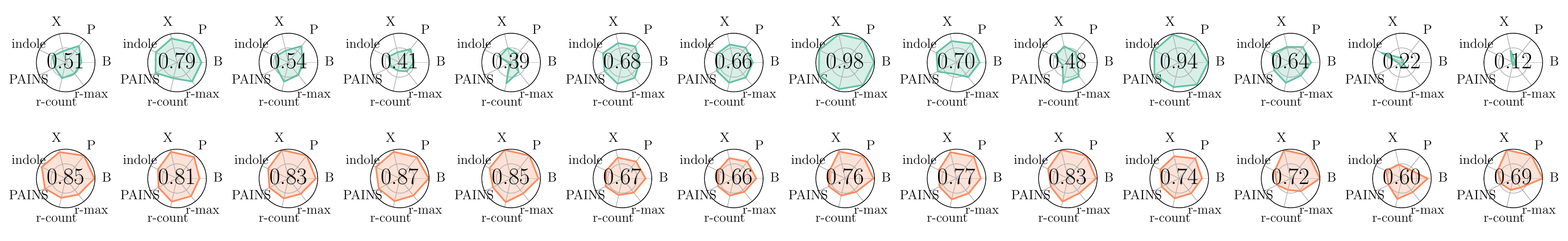}};
    \end{tikzpicture}

    \begin{tikzpicture}[x=1cm, y=1cm]
        \node[anchor=east, minimum height=1.5cm, minimum width=2cm, align=center] at (-3, 0) {ProtGNN\\+GIN};  
        \node at (2.5, 0) {\includegraphics[width=0.8\linewidth]{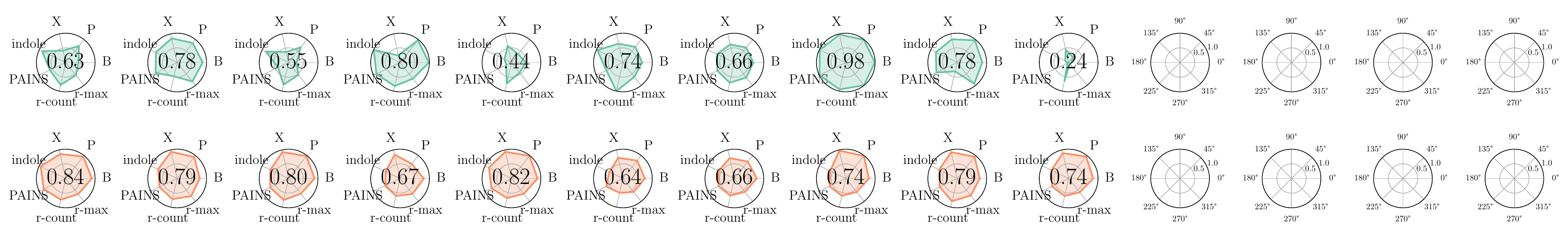}};
    \end{tikzpicture}

    \caption{Evaluation of \textbf{node explanations} for all model-explainer combinations. Null explanation results are shown in green, and subgraph explanation results in orange. Overall average scores for each method are displayed in the center.}
    \label{fig:radar-node}
\end{figure*}

\begin{figure*}[htb]
    \centering
    \scriptsize

    \begin{tikzpicture}[x=1cm, y=1cm] 
        \node[anchor=east, minimum height=1.5cm, minimum width=2cm, align=center] at (-3, 0) {GCN}; 
        \node at (2.5, 0) {\includegraphics[width=0.8\linewidth]{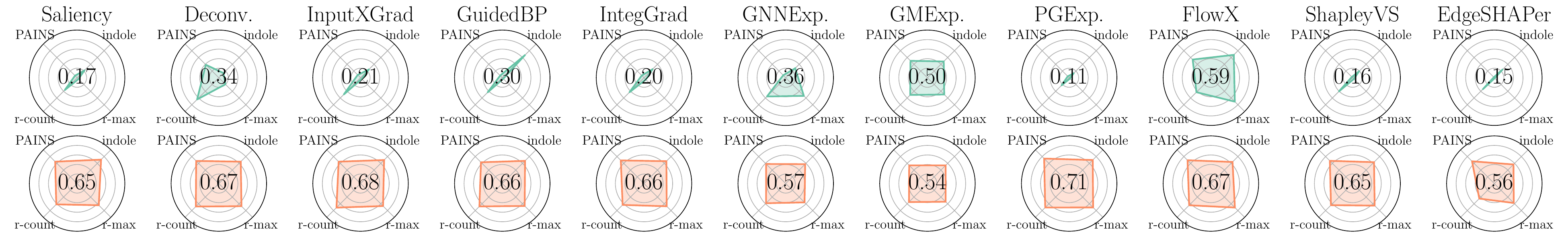}};
    \end{tikzpicture}

    \begin{tikzpicture}[x=1cm, y=1cm]
        \node[anchor=east, minimum height=1.5cm, minimum width=2cm, align=center] at (-3, 0) {ProtGNN\\+GCN};  
        \node at (2.5, 0) {\includegraphics[width=0.8\linewidth]{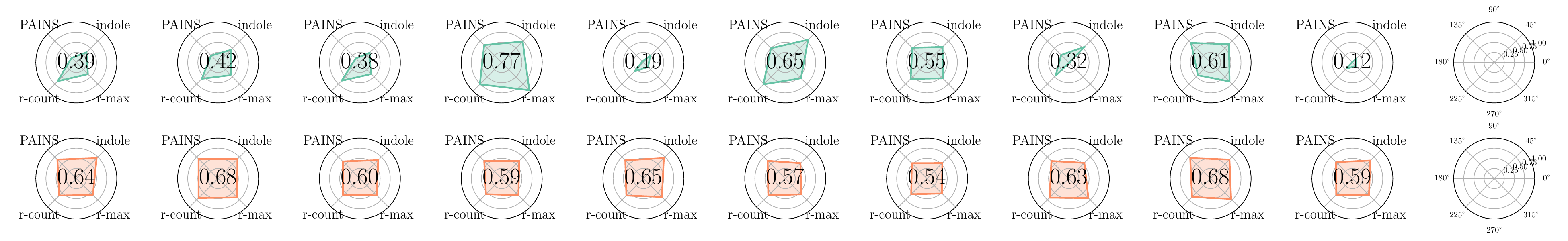}};
    \end{tikzpicture}

    \begin{tikzpicture}[x=1cm, y=1cm]
        \node[anchor=east, minimum height=1.5cm, minimum width=2cm, align=center] at (-3, 0) {GAT};  
        \node at (2.5, 0) {\includegraphics[width=0.8\linewidth]{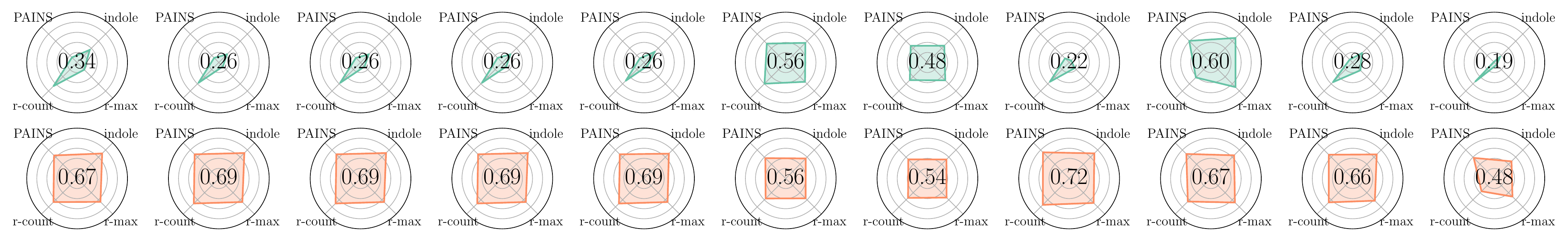}};
    \end{tikzpicture}

    \begin{tikzpicture}[x=1cm, y=1cm]
        \node[anchor=east, minimum height=1.5cm, minimum width=2cm, align=center] at (-3, 0) {ProtGNN\\+GAT};  
        \node at (2.5, 0) {\includegraphics[width=0.8\linewidth]{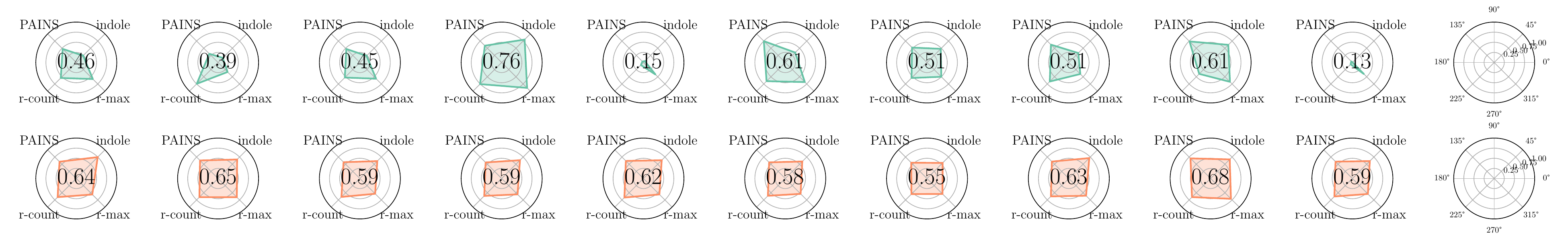}};
    \end{tikzpicture}

    \begin{tikzpicture}[x=1cm, y=1cm]
        \node[anchor=east, minimum height=1.5cm, minimum width=2cm, align=center] at (-3, 0) {GIN};  
        \node at (2.5, 0) {\includegraphics[width=0.8\linewidth]{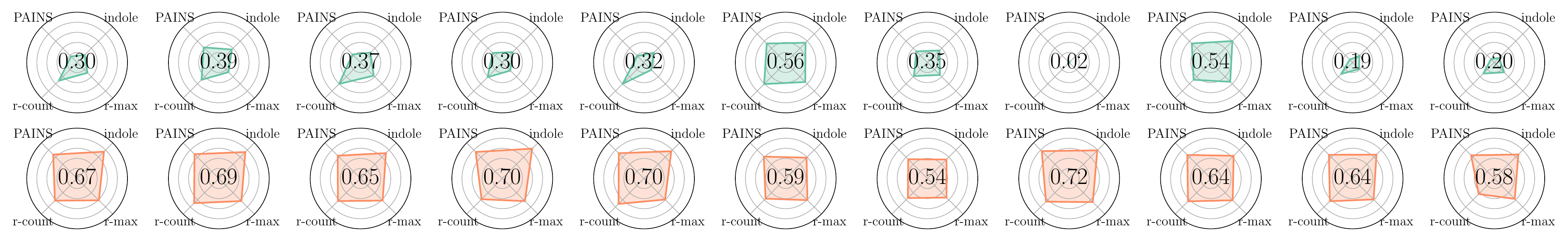}};
    \end{tikzpicture}

    \begin{tikzpicture}[x=1cm, y=1cm]
        \node[anchor=east, minimum height=1.5cm, minimum width=2cm, align=center] at (-3, 0) {ProtGNN\\+GIN};  
        \node at (2.5, 0) {\includegraphics[width=0.8\linewidth]{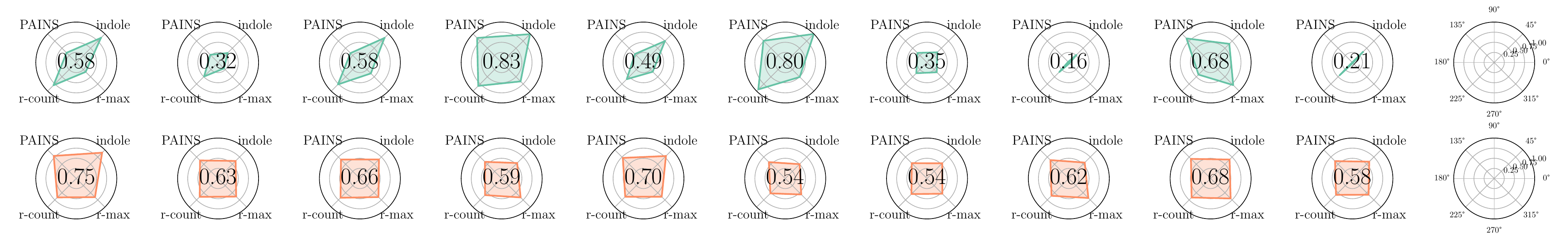}};
    \end{tikzpicture}
    \caption{Evaluation of \textbf{edge explanations} for all model-explainer combinations. Null explanation results are shown in green, and subgraph explanation results in orange. Overall average scores for each method are displayed in the center.}
    \label{fig:radar-edges}
\end{figure*}

The radar plots in Figure~\ref{fig:radar-node} and Figure~\ref{fig:radar-edges} illustrate the evaluation scores of each model–explainer combination across all $7$ tasks, providing a visual comparison of their performance.

\subsection{Examples of explanations} \label{ap:examples}

\begin{figure}[htb]
    \centering
    \includegraphics[width=\linewidth]{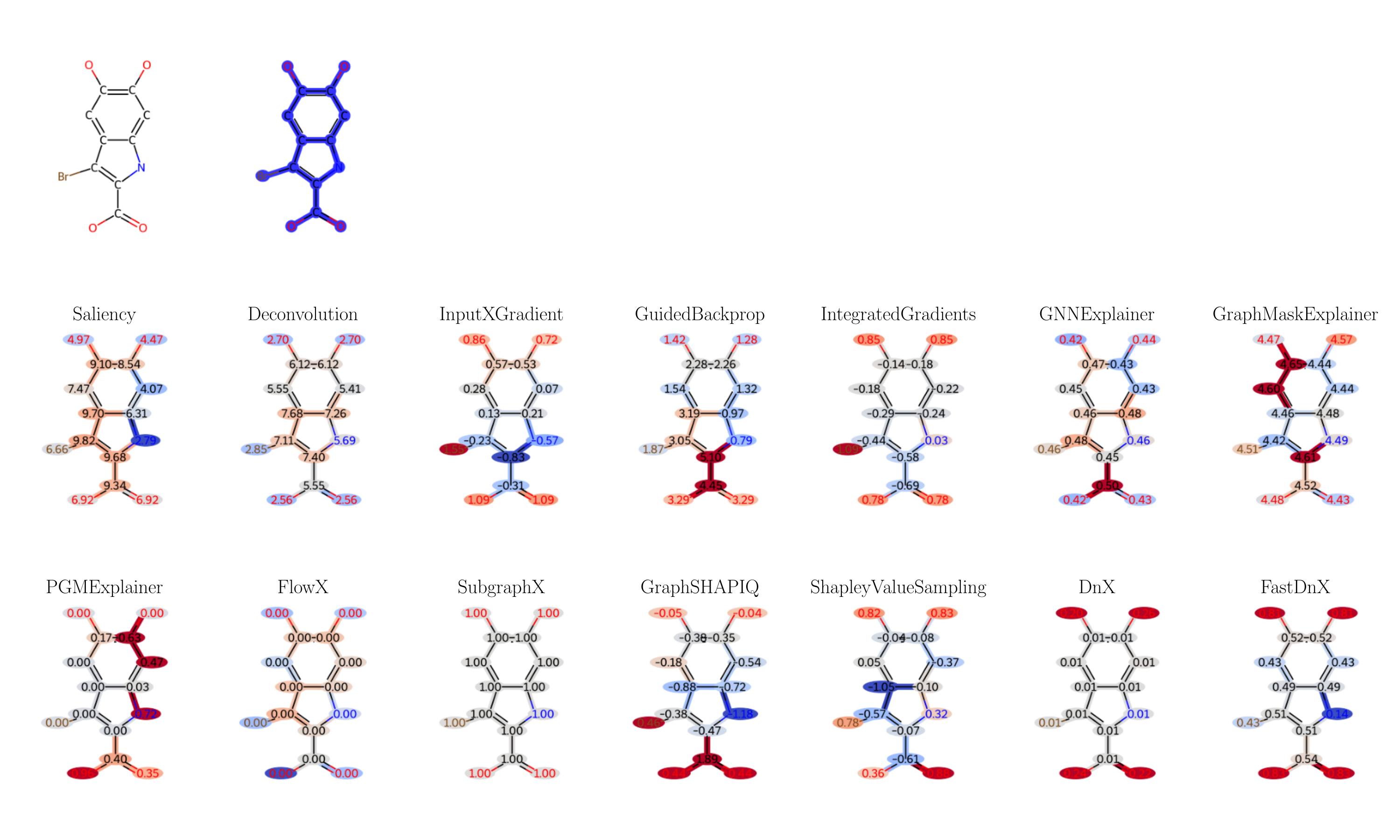} \\
    \includegraphics[width=\linewidth]{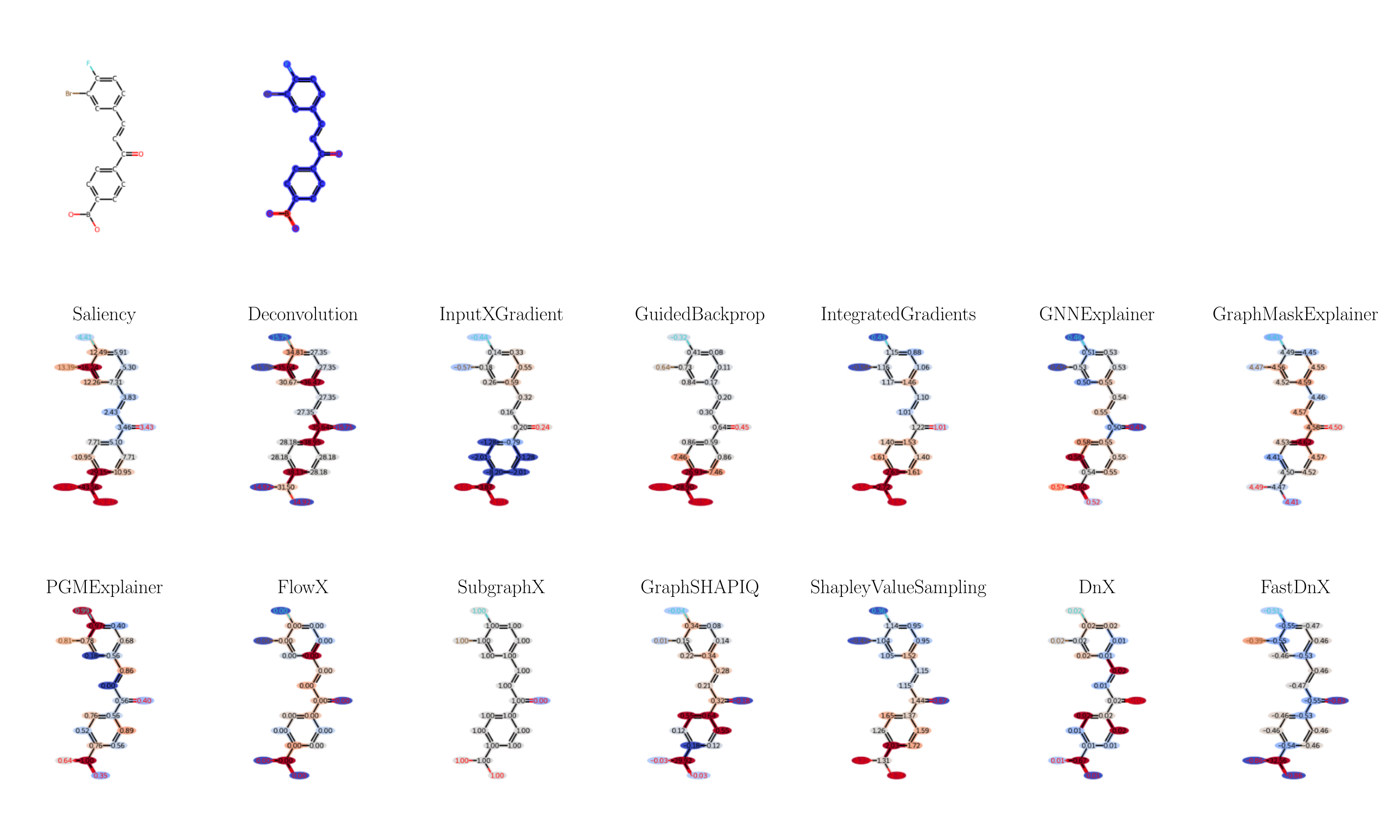}
    \caption{Node-level explanation examples on graphs from different classes in the \textbf{B} task, using the GIN model and different explanation methods.}
    \label{fig:vis-b}
\end{figure}

\begin{figure}[htb]
    \centering
    \includegraphics[width=\linewidth]{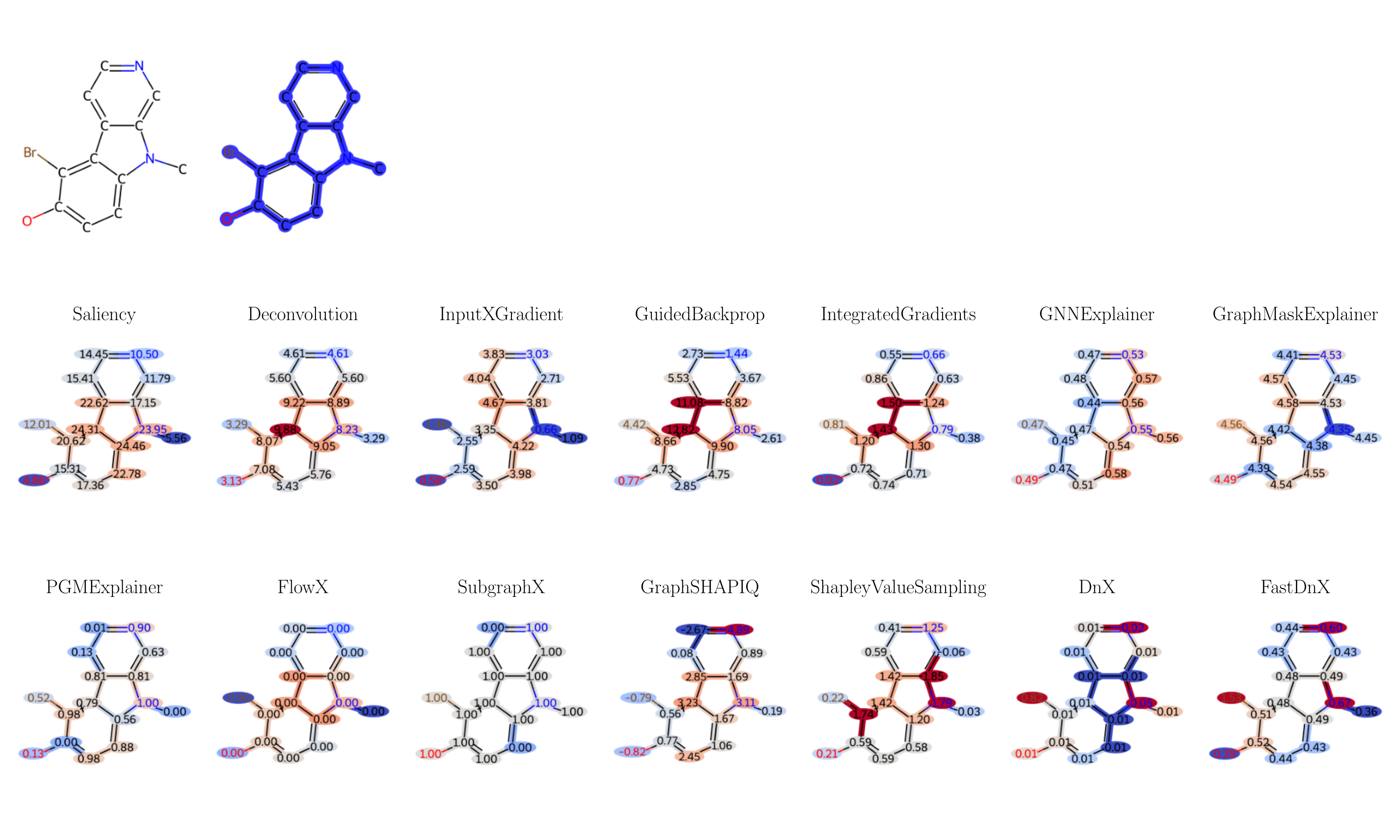} \\
    \includegraphics[width=\linewidth]{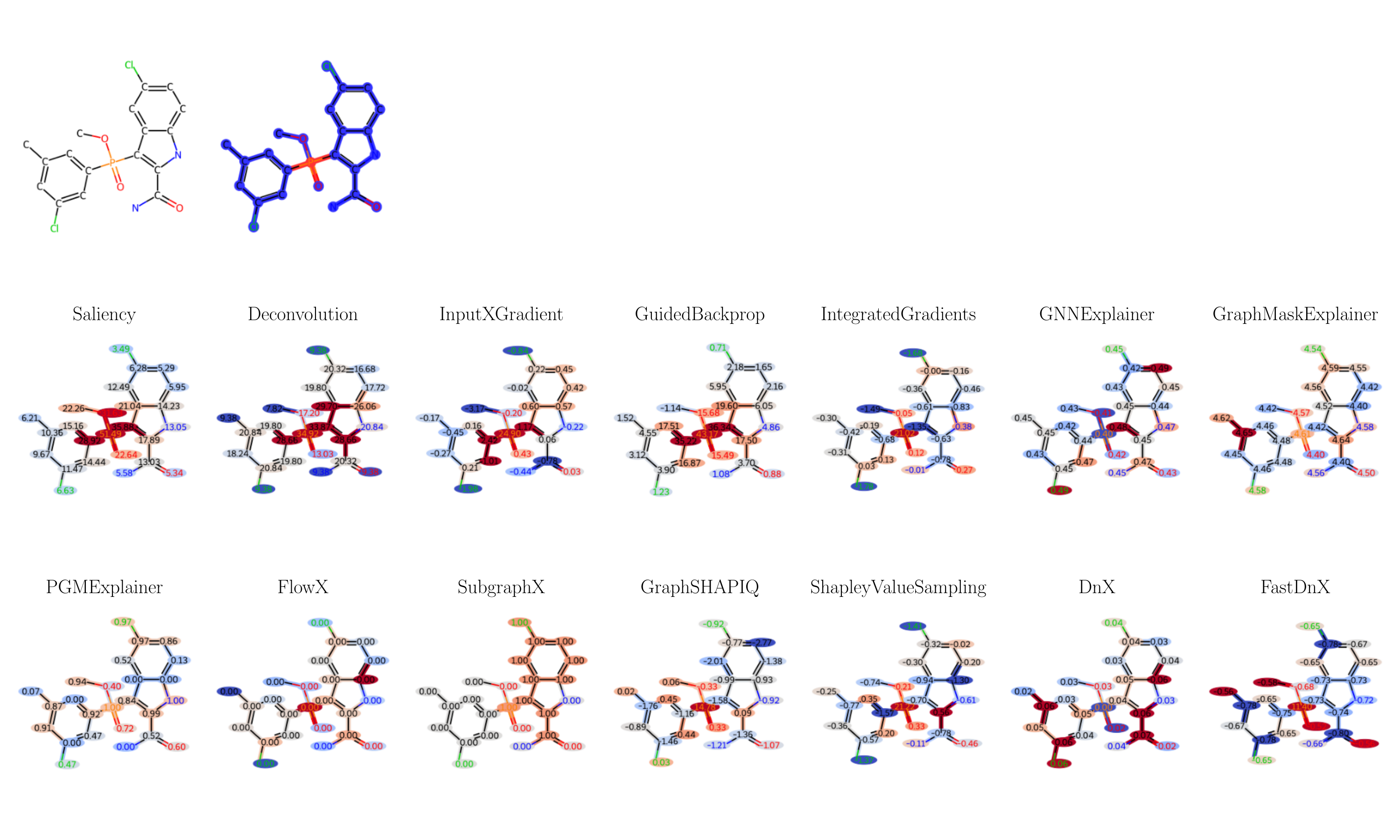}
    \caption{Node-level explanation examples on graphs from different classes in the \textbf{P} task, using the GIN model and different explanation methods.}
    \label{fig:vis-p}
\end{figure}

\begin{figure}[htb]
    \centering
    \includegraphics[width=\linewidth]{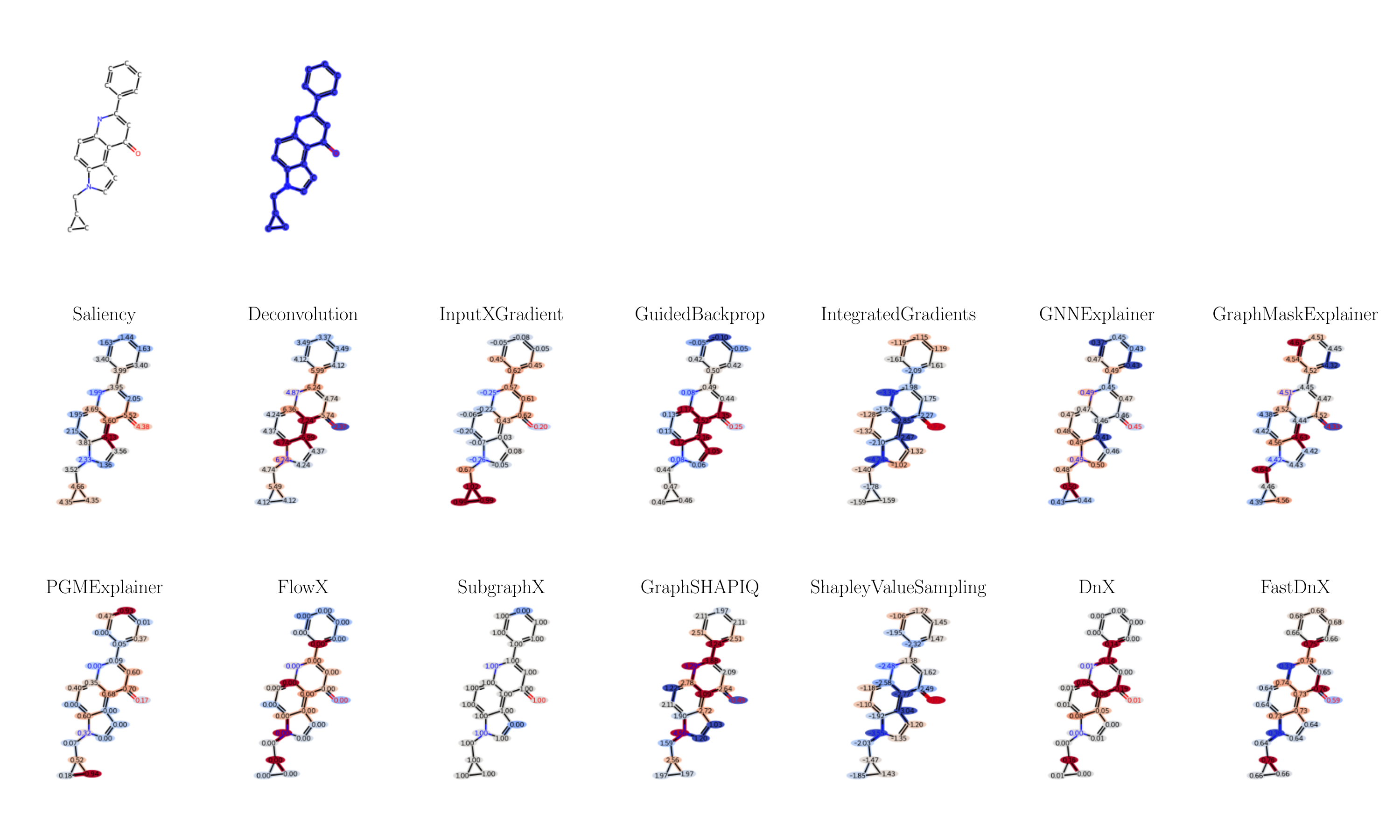} \\
    \includegraphics[width=\linewidth]{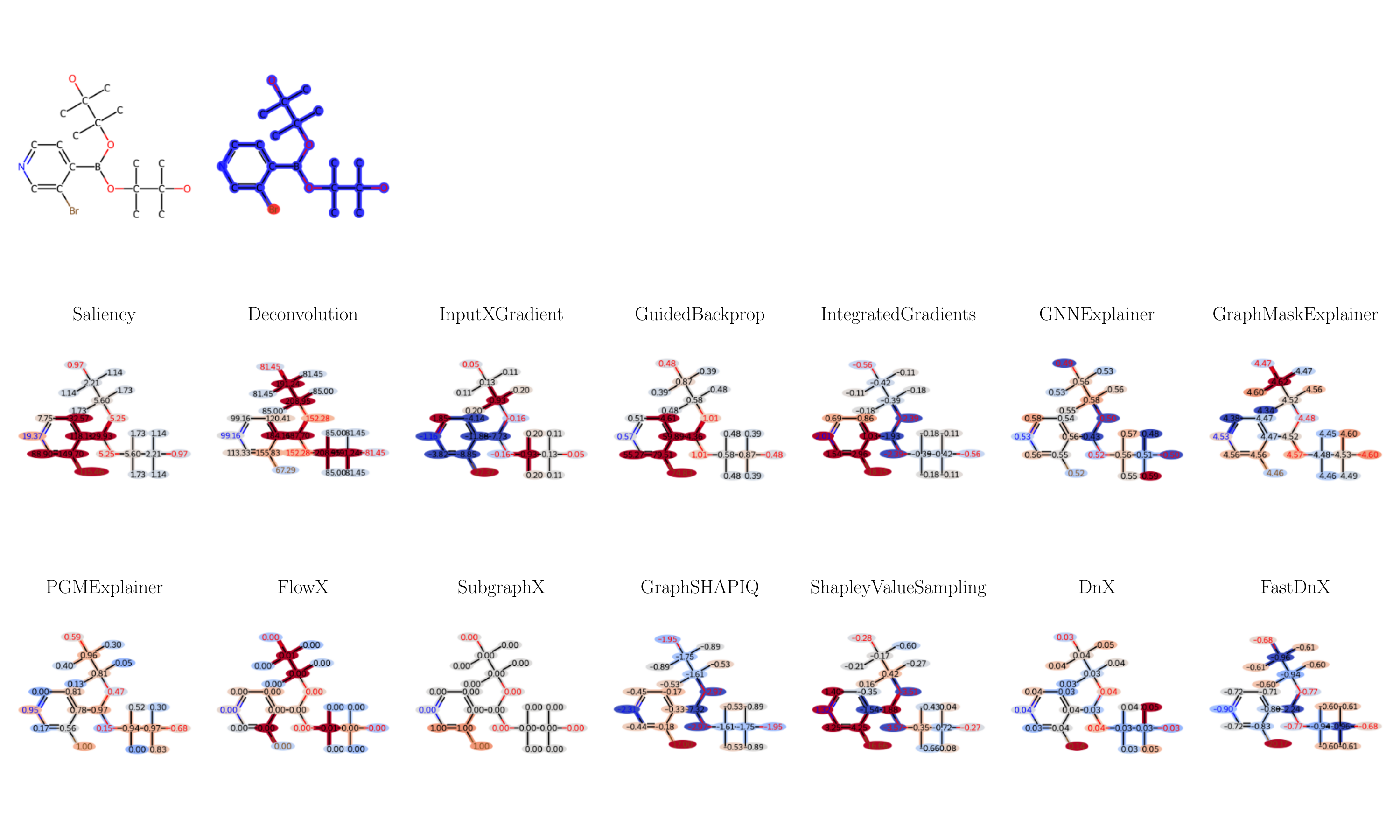}
    \caption{Node-level explanation examples on graphs from different classes in the \textbf{X} task, using the GIN model and different explanation methods.}
    \label{fig:vis-x}
\end{figure}

\begin{figure}[htb]
    \centering
    \includegraphics[width=\linewidth]{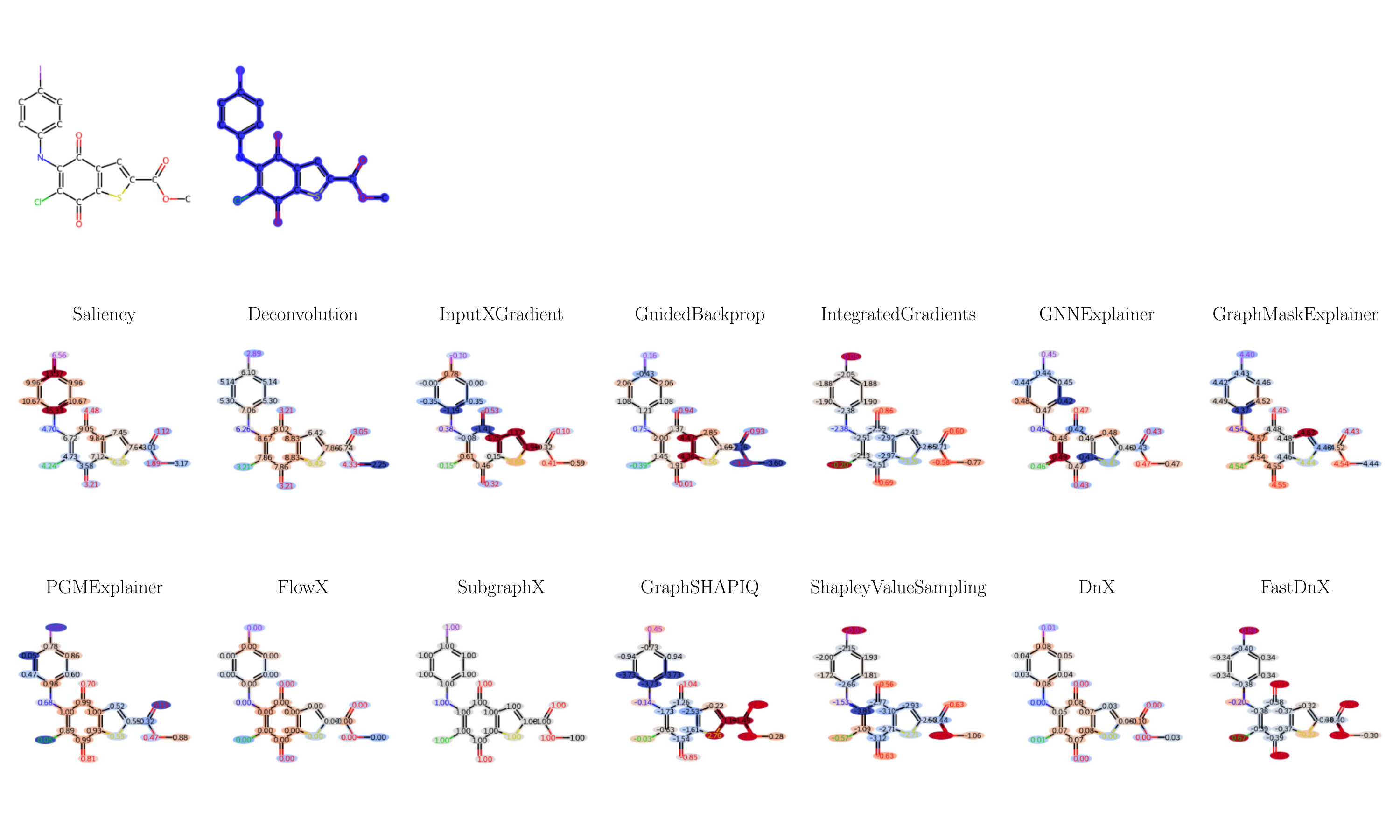} \\
    \includegraphics[width=\linewidth]{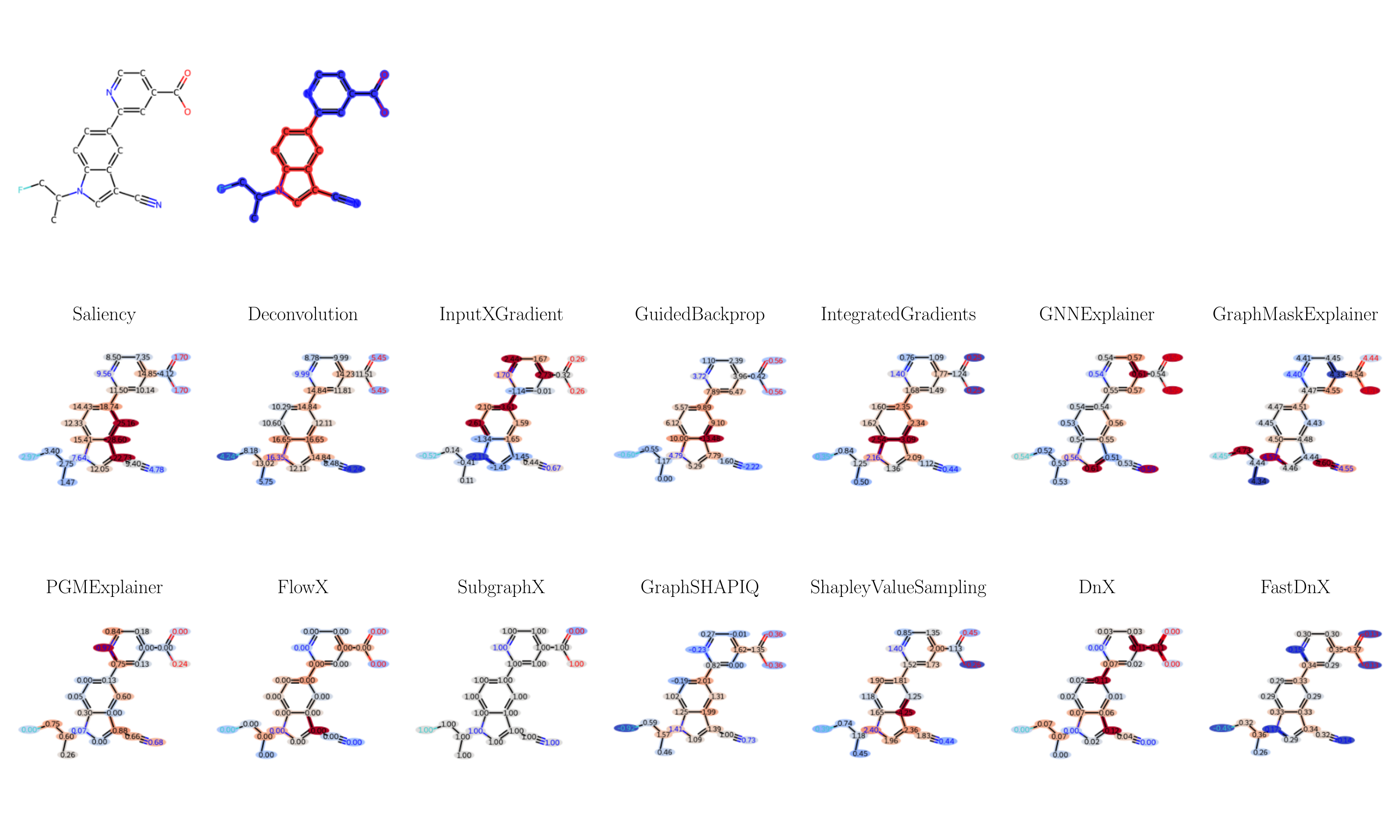}
    \caption{Node-level explanation examples on graphs from different classes in the \textbf{indole} task, using the GIN model and different explanation methods.}
    \label{fig:vis-indole}
\end{figure}

\begin{figure}[htb]
    \centering
    \includegraphics[width=\linewidth]{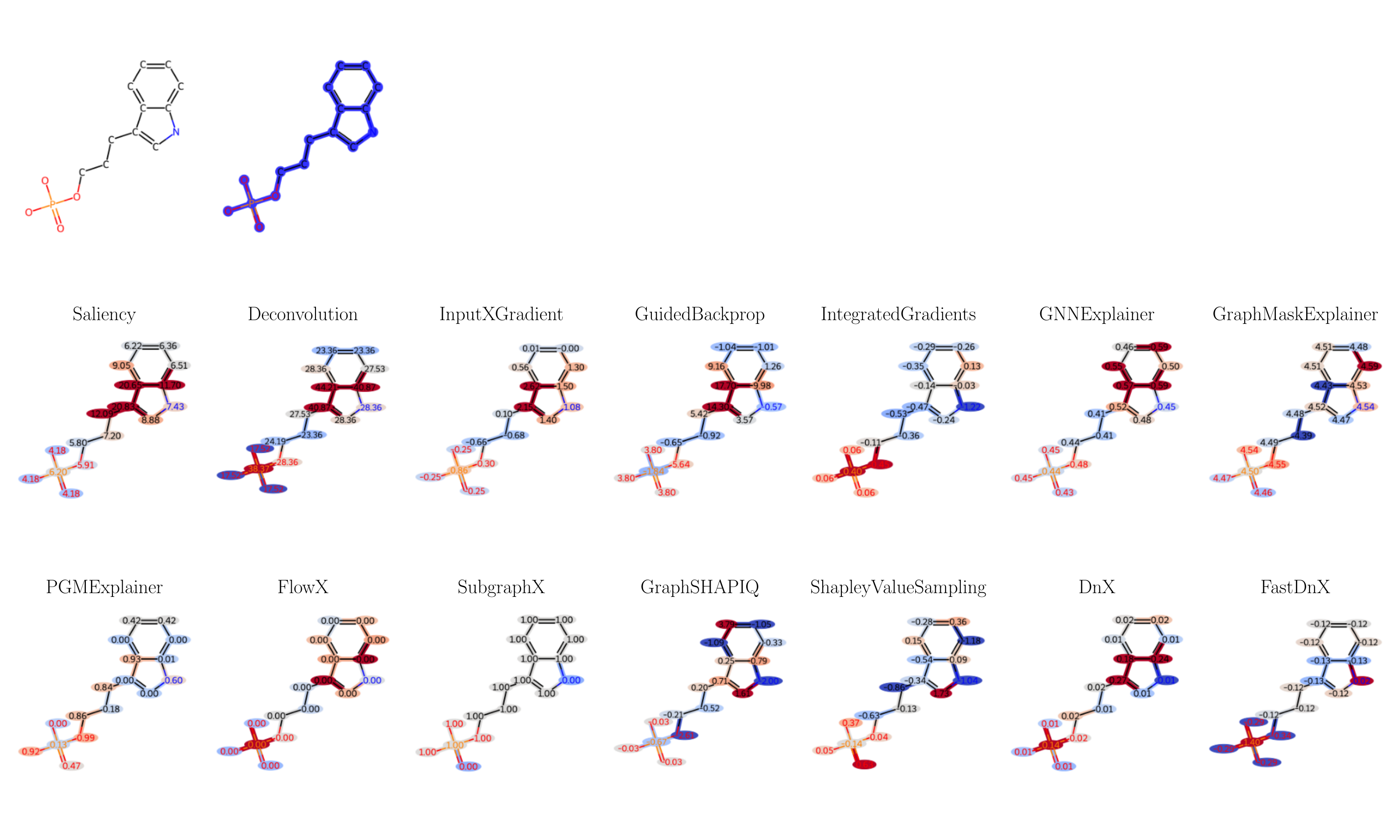} \\
    \includegraphics[width=\linewidth]{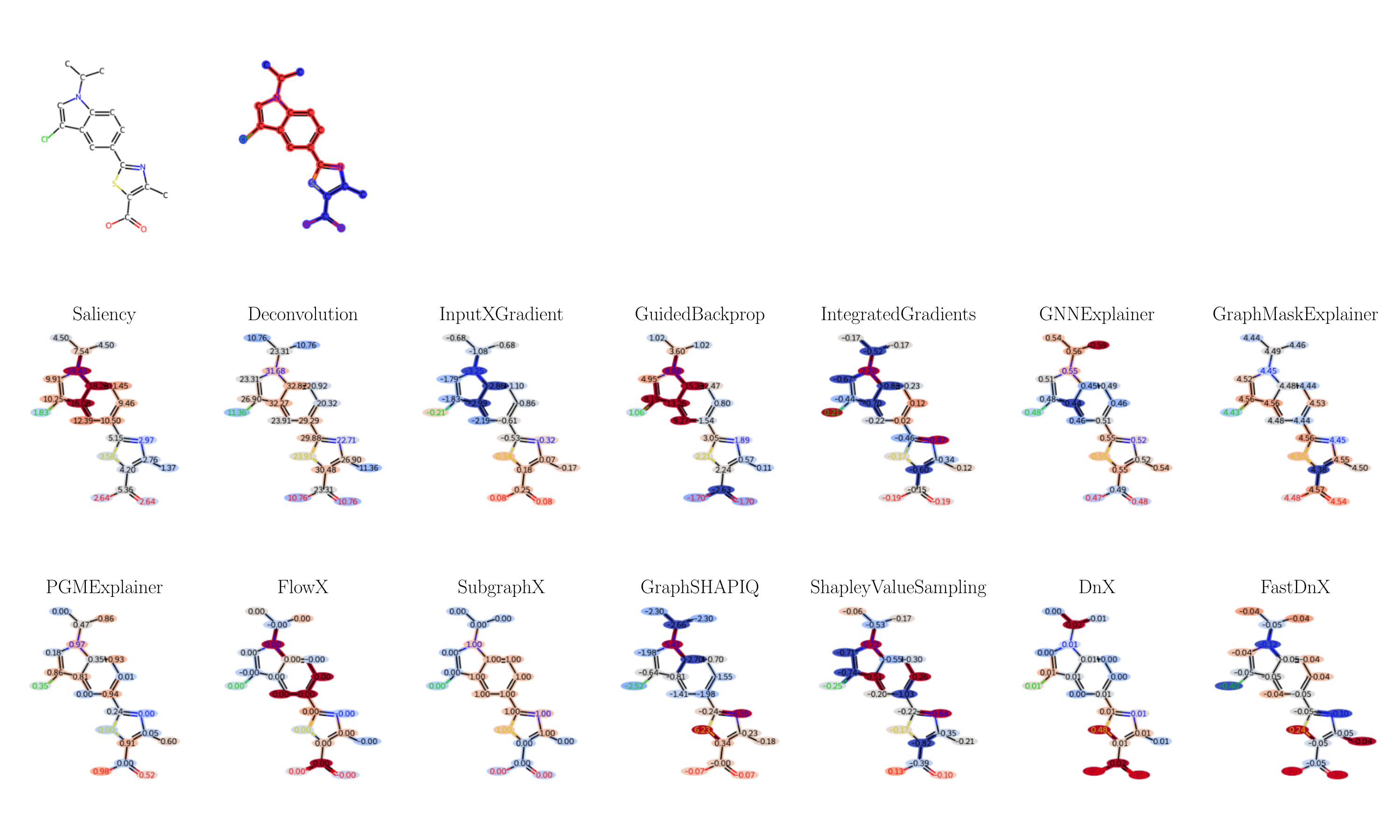}
    \caption{Node-level explanation examples on graphs from different classes in the \textbf{PAINS} task, using the GIN model and different explanation methods.}
    \label{fig:vis-pains}
\end{figure}

\begin{figure}[htb]
    \centering
    \includegraphics[width=\linewidth]{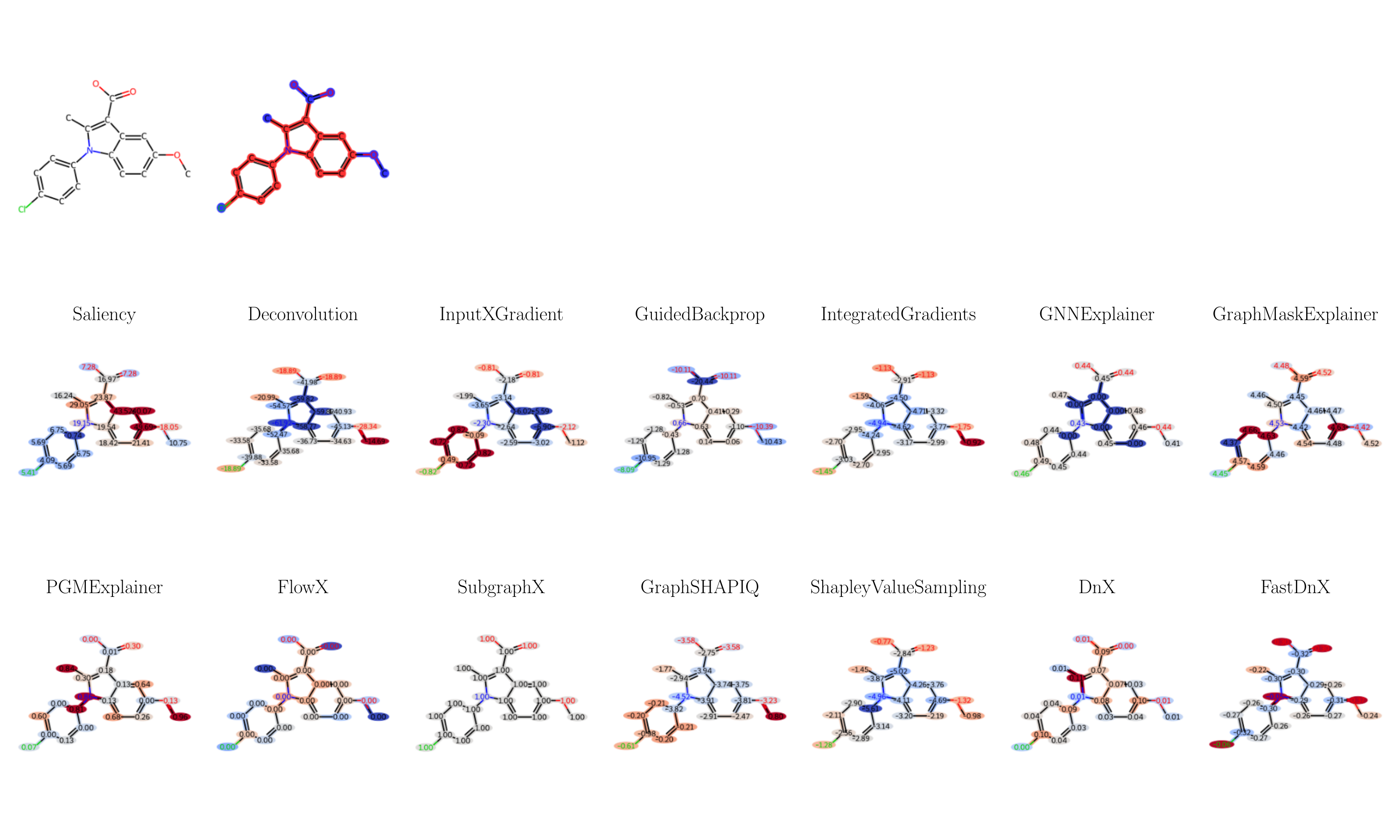} \\
    \includegraphics[width=\linewidth]{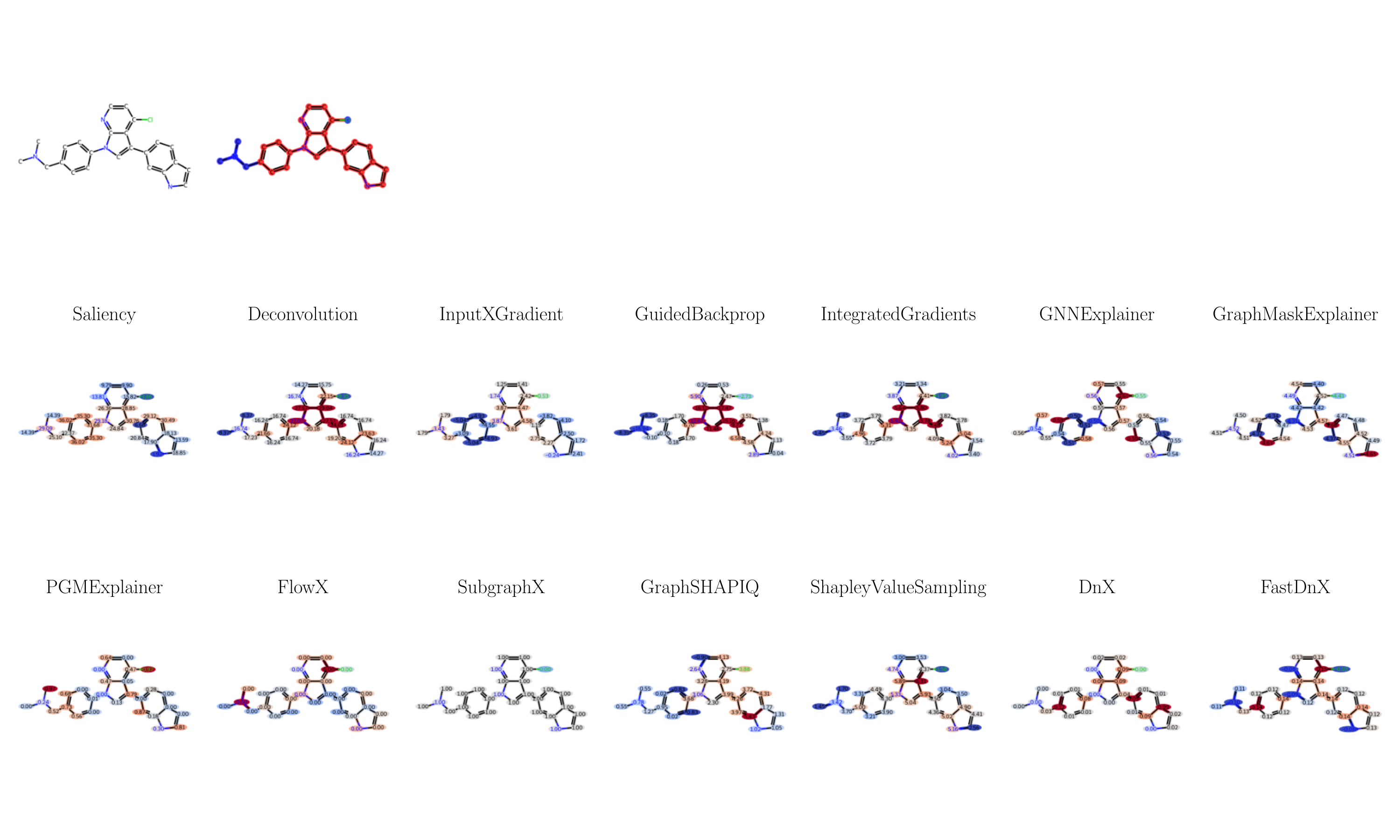}
    \caption{Node-level explanation examples on graphs from different classes in the \textbf{rings-count} task, using the GIN model and different explanation methods.}
    \label{fig:vis-count}
\end{figure}

\begin{figure}[htb]
    \centering
    \includegraphics[width=\linewidth]{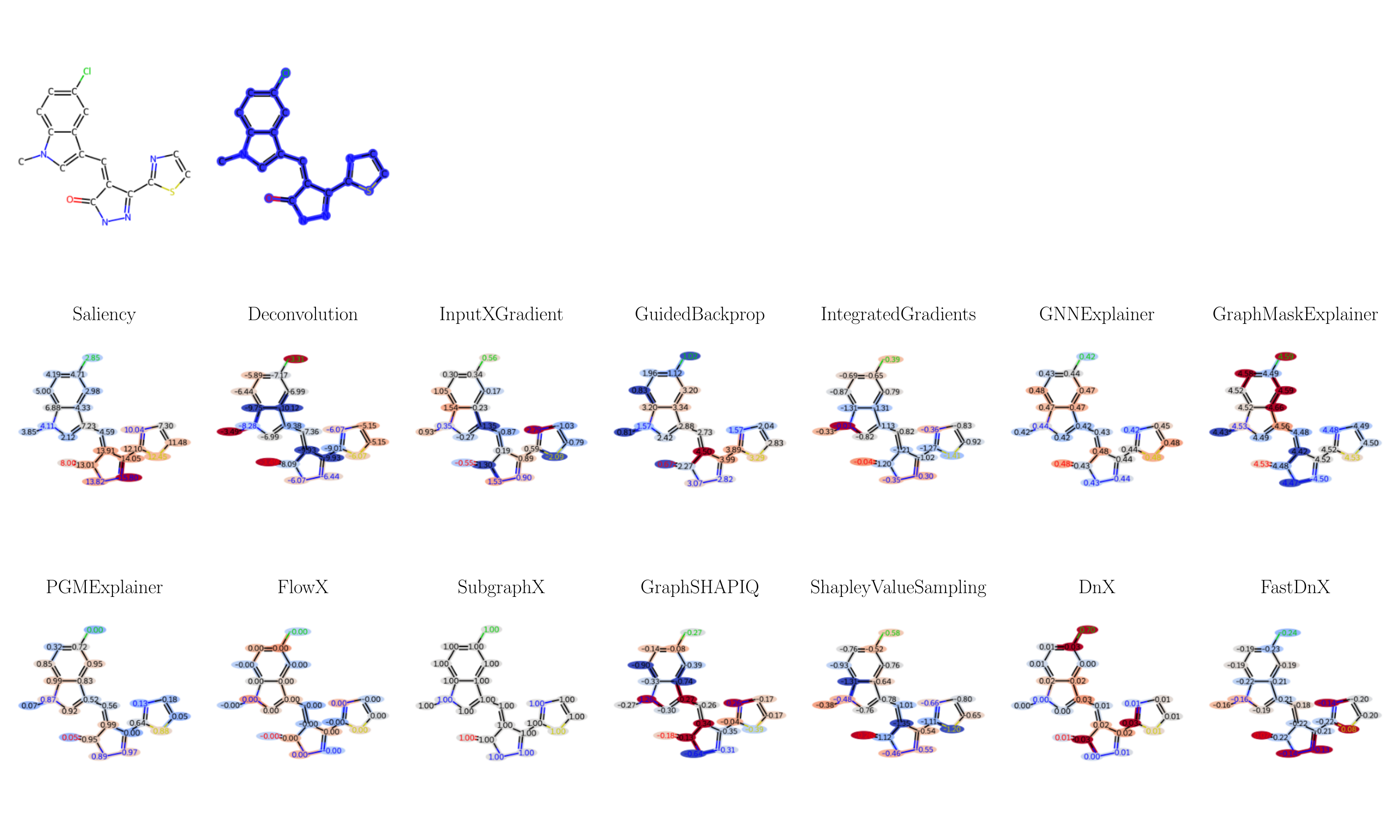} \\
    \includegraphics[width=\linewidth]{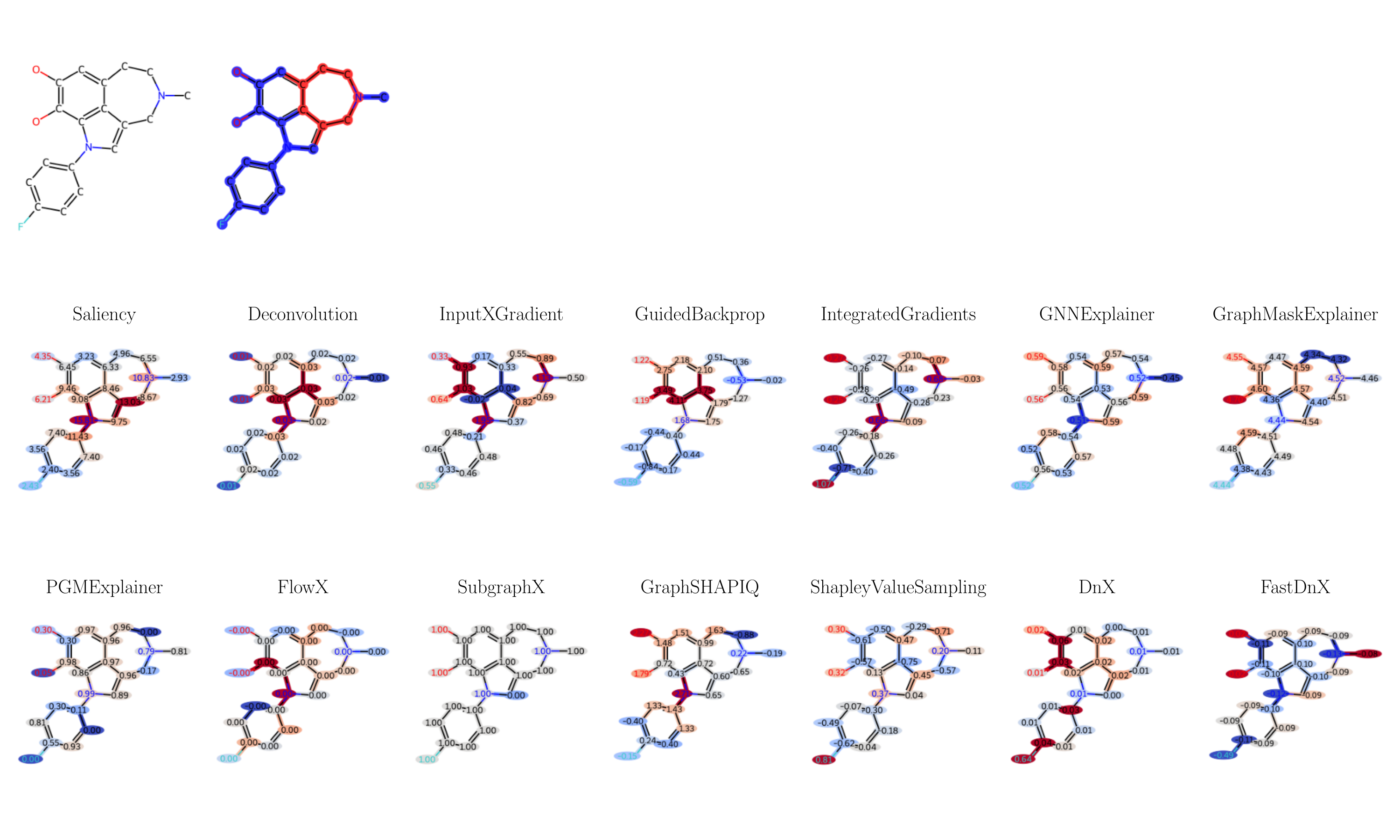}
    \caption{Node-level explanation examples on graphs from different classes in the \textbf{rings-max} task, using the GIN model and different explanation methods.}
    \label{fig:vis-max}
\end{figure}

Figures~\ref{fig:vis-b}–\ref{fig:vis-max} present examples of node explanations for the GIN classifier for each task using the evaluated explainers. The colors in the null explanations are scaled to highlight outlier scores based on the IQR method applied in our evaluation. Specifically, scores below Q1 – 1.5 × IQR or above Q3 + 1.5 × IQR are considered outliers, where Q1 and Q3 represent the 25th and 75th percentiles, respectively, and IQR = Q3 – Q1.

\clearpage

\section*{NeurIPS Paper Checklist}

\begin{enumerate}

\item {\bf Claims}
    \item[] Question: Do the main claims made in the abstract and introduction accurately reflect the paper's contributions and scope?
    \item[] Answer: \answerYes{} 
    \item[] Justification: Abstract and Introduction clearly states our contributions. Those are listed after "we introduce"

\item {\bf Limitations}
    \item[] Question: Does the paper discuss the limitations of the work performed by the authors?
    \item[] Answer: \answerYes{} 
    \item[] Justification: Discussion section provides limitations paragraph. 

\item {\bf Theory assumptions and proofs}
    \item[] Question: For each theoretical result, does the paper provide the full set of assumptions and a complete (and correct) proof?
    \item[] Answer: \answerYes{} 
    \item[] Justification: One proposition is presented with its proof in the Appendix.

    \item {\bf Experimental result reproducibility}
    \item[] Question: Does the paper fully disclose all the information needed to reproduce the main experimental results of the paper to the extent that it affects the main claims and/or conclusions of the paper (regardless of whether the code and data are provided or not)?
    \item[] Answer: \answerYes{} 
    \item[] Justification: In Appendix, we have a Reproducibility subsection.

\item {\bf Open access to data and code}
    \item[] Question: Does the paper provide open access to the data and code, with sufficient instructions to faithfully reproduce the main experimental results, as described in supplemental material?
    \item[] Answer: \answerYes{} 
    \item[] Justification: We provide links to the data and code, along with further details, in the Reproducibility subsection of the appendix.

\item {\bf Experimental setting/details}
    \item[] Question: Does the paper specify all the training and test details (e.g., data splits, hyperparameters, how they were chosen, type of optimizer) necessary to understand the results?
    \item[] Answer: \answerYes{} 
    \item[] Justification: They are provided in the Experimental Setup subsection in the main paper, and the Reproducibility subsection in the Appendix.

\item {\bf Experiment statistical significance}
    \item[] Question: Does the paper report error bars suitably and correctly defined or other appropriate information about the statistical significance of the experiments?
    \item[] Answer: \answerYes{} 
    \item[] Justification: Yes, we report error bars, also we report standard deviation of the results across seeds. 

\item {\bf Experiments compute resources}
    \item[] Question: For each experiment, does the paper provide sufficient information on the computer resources (type of compute workers, memory, time of execution) needed to reproduce the experiments?
    \item[] Answer: \answerYes{} 
    \item[] Justification: In the Reproducibility subsection in the appendix.
    
\item {\bf Code of ethics}
    \item[] Question: Does the research conducted in the paper conform, in every respect, with the NeurIPS Code of Ethics \url{https://neurips.cc/public/EthicsGuidelines}?
    \item[] Answer: \answerYes{} 
    \item[] Justification: We reviewed the code of ethics, and we followed it.

\item {\bf Broader impacts}
    \item[] Question: Does the paper discuss both potential positive societal impacts and negative societal impacts of the work performed?
    \item[] Answer: \answerYes{} 
    \item[] Justification: Those discussion is provided in the Appendix. 
    
\item {\bf Safeguards}
    \item[] Question: Does the paper describe safeguards that have been put in place for responsible release of data or models that have a high risk for misuse (e.g., pre-trained language models, image generators, or scraped datasets)?
    \item[] Answer: \answerYes{} 
    \item[] Justification: It's available in paragraph: safeguards in the Appendix. 

\item {\bf Licenses for existing assets}
    \item[] Question: Are the creators or original owners of assets (e.g., code, data, models), used in the paper, properly credited and are the license and terms of use explicitly mentioned and properly respected?
    \item[] Answer: \answerYes{} 
    \item[] Justification: Yes, we provide information about the licensing of assets in the Reproducibility subsection and the Safeguard paragraph in the Appendix as well as in links provided with dataset and code. 

\item {\bf New assets}
    \item[] Question: Are new assets introduced in the paper well documented and is the documentation provided alongside the assets?
    \item[] Answer: \answerYes{} 
    \item[] Justification: Yes, a novel dataset is created. It is hosted on HuggingFace with all needed documentation.

\item {\bf Crowdsourcing and research with human subjects}
    \item[] Question: For crowdsourcing experiments and research with human subjects, does the paper include the full text of instructions given to participants and screenshots, if applicable, as well as details about compensation (if any)? 
    \item[] Answer: \answerNA{} 
    \item[] Justification: No crowdsourcing and human subjects were employed

\item {\bf Institutional review board (IRB) approvals or equivalent for research with human subjects}
    \item[] Question: Does the paper describe potential risks incurred by study participants, whether such risks were disclosed to the subjects, and whether Institutional Review Board (IRB) approvals (or an equivalent approval/review based on the requirements of your country or institution) were obtained?
    \item[] Answer: \answerNA{} 
    \item[] Justification: There was no need for such approval. 

\item {\bf Declaration of LLM usage}
    \item[] Question: Does the paper describe the usage of LLMs if it is an important, original, or non-standard component of the core methods in this research? Note that if the LLM is used only for writing, editing, or formatting purposes and does \emph{not} impact the core methodology, scientific rigor, or originality of the research, declaration is not required.
    \item[] Answer: \answerYes{} 
    \item[] Justification: We provide information about LLMs usage. 

\end{enumerate}

\end{document}